\documentclass[10pt,journal,compsoc]{IEEEtran}
\usepackage{times}
\usepackage{epsfig}
\usepackage{graphicx}
\usepackage{subfigure}
\usepackage{amsmath}
\usepackage{amssymb}
\usepackage{booktabs}
\usepackage{cite}
\usepackage{multirow}
\usepackage{CJKutf8}
\usepackage{booktabs}
\usepackage{dsfont}
\usepackage{setspace}
\usepackage{xcolor}
\usepackage{url}
\usepackage[hidelinks]{hyperref}
\usepackage[ruled,linesnumbered]{algorithm2e}
\usepackage{bbm}
\allowdisplaybreaks
\usepackage{eqparbox}

\newcommand{\hc}[1]{{\color{black} #1}}

\begin{document}
%
\title{Skeleton Prototype Contrastive Learning with Multi-Level Graph Relation Modeling for Unsupervised Person Re-Identification}

%
%
%
%

\author{Haocong~Rao, Chunyan~Miao

\IEEEcompsocitemizethanks{
\IEEEcompsocthanksitem H. Rao and C. Miao are with Joint NTU-UBC Research Centre of Excellence in Active Living for the Elderly (LILY) and School of Computer Science and Engineering, Nanyang Technological University, Singapore 639798, Singapore. E-mail: haocongrao@gmail.com, haocong001@e.ntu.edu.sg, ascymiao@ntu.edu.sg.
\IEEEcompsocthanksitem Corresponding author: Chunyan Miao
\IEEEcompsocthanksitem \hc{Our codes are available at \href{https://github.com/Kali-Hac/SPC-MGR}{https://github.com/Kali-Hac/SPC-MGR}.}

}
} 
\IEEEtitleabstractindextext{%
\begin{abstract}
Person re-identification (re-ID) via 3D skeletons is an \hc{important emerging} topic with many merits. Existing solutions rarely explore valuable body-component relations in skeletal structure or motion, and they typically lack the ability to learn general representations with unlabeled skeleton data for person re-ID. This paper proposes a generic \textit{unsupervised} Skeleton Prototype Contrastive learning paradigm with Multi-level Graph Relation learning (SPC-MGR) to learn effective representations from \textit{unlabeled} skeletons to perform person re-ID. Specifically, we first construct \textit{unified multi-level skeleton graphs} to fully model body structure within skeletons. \hc{Then we propose a \textit{multi-head structural relation layer} to comprehensively capture relations of physically-connected body-component nodes in graphs. A \textit{full-level collaborative relation layer} is exploited to infer collaboration between motion-related body parts at various levels, so as to capture rich body features and recognizable walking patterns}.
Lastly, we propose a \textit{skeleton prototype contrastive learning scheme} that clusters feature-correlative instances of unlabeled graph representations and contrasts their inherent similarity with representative skeleton features (``\textit{skeleton prototypes}'') to learn discriminative skeleton representations for person re-ID.
Empirical evaluations show that SPC-MGR significantly outperforms several state-of-the-art skeleton-based methods, and it also achieves highly competitive person re-ID performance \hc{for} more general scenarios.

\end{abstract}

}

\maketitle

\IEEEdisplaynontitleabstractindextext

%
\IEEEpeerreviewmaketitle

\IEEEraisesectionheading{\section{Introduction}\label{sec:introduction}}

%
%
%
%
\IEEEPARstart{P}{erson} re-identification (re-ID) \hc{aims at identifying or matching} a target pedestrian across different views or scenes, which plays an essential role in safety-critical applications including intelligent video surveillance, security authentication and human tracking
\cite{nambiar2019gait,zheng2015towards,baltieri2011sarc3d,vezzani2013people,su2018multi,chen2018person,li2019unsupervised,qian2019leader,yu2020unsupervised}.
Conventional studies \cite{wang2003silhouette,wang2011human,wang2016person,zhao2017person,zhang2019densely,karianakis2018reinforced,ge2020selfpaced} typically utilize visual features such as human appearances, silhouettes and body \hc{textures} from RGB or depth images to discriminate different individuals. Nevertheless, this kind of methods are often vulnerable to appearance, lighting and clothing variation in practice. 
Compared with RGB-based and depth-based methods, 3D skeleton-based models \cite{barbosa2012re,andersson2015person,rao2020self,rao2021self,rao2021multi,rao2021sm} exploit 3D coordinates of numerous key joints to characterize human body and motion, which \hc{enjoys} smaller data size and better robustness to scale and view variation \cite{han2017space}. With these advantages, 3D skeleton data have drawn surging attention in the fields of person re-ID and gait recognition \cite{991036,liao2020model,rao2020self,rao2021multi,rao2021sm,rao2021self}. However, the way to model discriminative body and motion features with 3D skeleton data remains to be an open challenge.

\begin{figure}
    \centering
    \scalebox{0.42}{
    \includegraphics{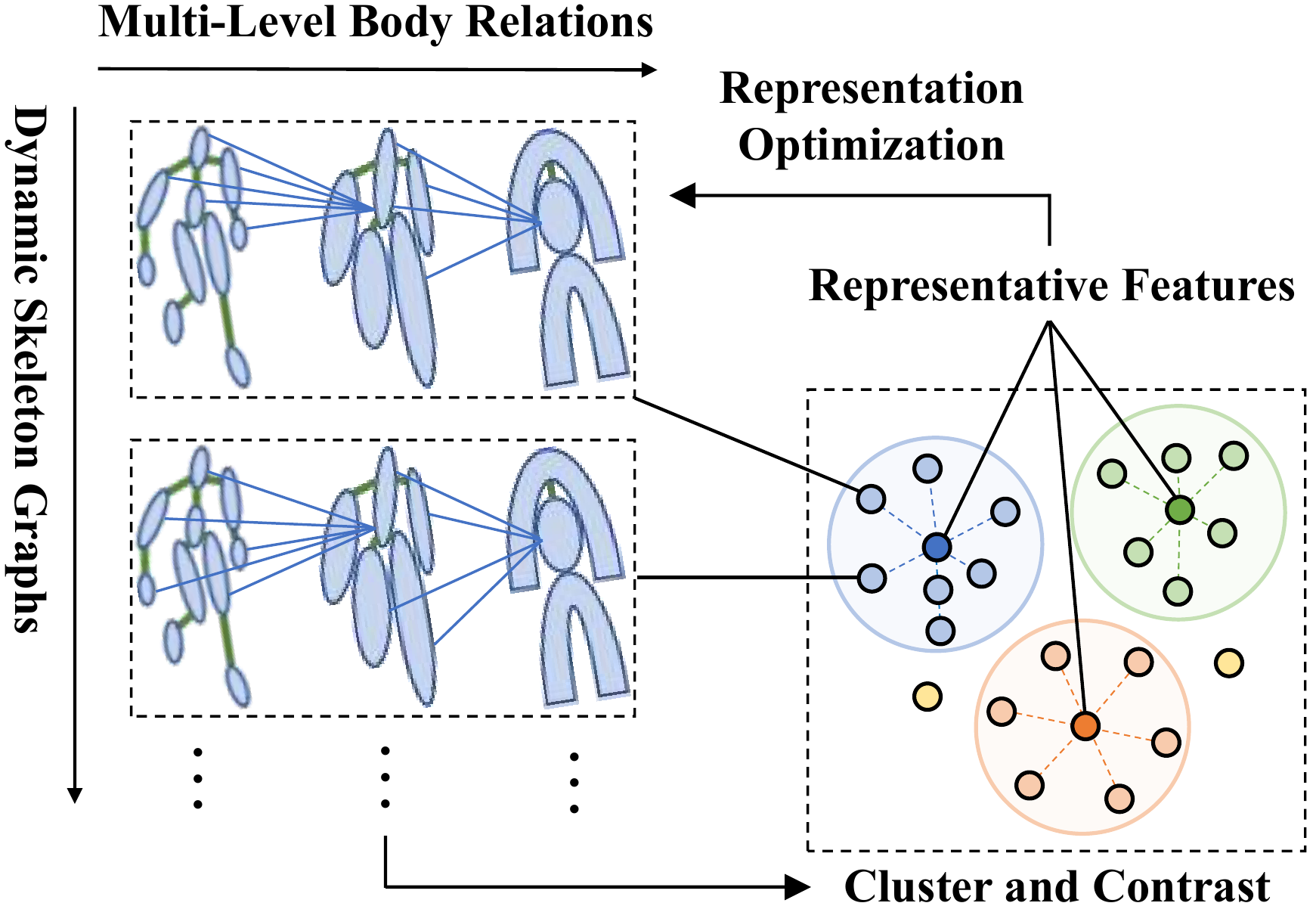}
    }
    \caption{Our approach constructs skeleton graphs to model multi-level body components and relations, and \hc{contrasts the clustered representative features} to learn effective skeleton representations for person re-ID.}
    \label{first_overview}
\end{figure}

To perform person re-ID via 3D skeletons, existing endeavors typically model skeleton features by two groups of methods. \hc{\textit{Skeleton descriptor based methods} \cite{barbosa2012re,andersson2015person,munaro2014one} manually} extract certain anthropometric and geometric attributes of body from skeleton data. However, these hand-crafted methods usually require domain knowledge such as human anatomy \cite{yoo2002extracting}, and cannot fully mine underlying features beyond human cognition. \hc{\textit{Deep neural network based methods} \cite{liao2020model,rao2020self,rao2021self} usually} leverage convolutional neural networks (CNN) or long short-term memory (LSTM) to learn skeleton representations with sequences of raw body-joint positions or pose descriptors ($e.g.,$ limb lengths). Nevertheless, these works rarely explore inherent relations between different body joints or components, \hc{which could} ignore some valuable structural information of human body. \hc{Taking} the human walking for example, neighbor body joints ``foot'' and ``knee'' have strong motion correlations, while they usually enjoy diverse degree of \textit{collaboration} with limb-level components ``leg'' and ``arm'' during movement, which can be exploited to catch unique and recognizable patterns \cite{murray1964walking}.
Other important flaws of this type of methods are label dependency and weak generalization ability. In practical terms, these methods usually require massive labeled data of pre-defined classes to either train the model from scratch \cite{liao2020model,rao2021multi} or fine-tune the pre-trained skeleton representations \cite{rao2020self,rao2021self,rao2021sm} to classify the known identities. As a result, 
they lack the flexibility to learn general and representative skeleton features that can re-identify different pedestrians under the unavailability of labels, which limits its application in many real-world scenarios.





\hc{To address the above challenges,} this work for the first time proposes a generic Skeleton Prototype Contrastive learning paradigm with Multi-level Graph Relation modeling (SPC-MGR) \hc{in Fig. \ref{first_overview}} that can comprehensively model body structure and relations at various levels and mine discriminative features from \textit{unlabeled} skeletons for person re-ID. Specifically, we first devise \textbf{\textit{multi-level graphs}} to represent each 3D skeleton in a \textit{unified} \textit{coarse-to-fine} manner, so as to fully model body structure within skeletons.
Then, to enable a comprehensive exploration of relations between different body components, we propose to model \textit{structural-collaborative body relations} within skeletons from multi-level graphs. In particular, since each body component is highly correlated with its physically-connected components and may possess different \textit{structural relations} ($e.g.,$ motion correlations), we propose a \textbf{\textit{multi-head structural relation layer}} (MSRL) to capture multiple relations between each body-component node and its neighbors within a graph, so as to aggregate key correlative features for effective node representations. Meanwhile,
motivated by the fact that dynamic cooperation of body components in motion could carry unique patterns ($e.g.$, gait) \cite{murray1964walking}, 
we propose a \textbf{\textit{full-level collaborative relation layer}} (FCRL) to adaptively infer \textit{collaborative relations} \hc{among motion-related components at both the \textit{same-level} and \textit{cross-level} in graphs.
Furthermore,} we exploit a multi-level graph feature fusion strategy to integrate features of different-level graphs via collaborative relations, which encourages the model to capture more graph structural semantics and discriminative skeleton features.
 Lastly, to mine effective features from \textit{unlabeled} skeleton graph representations (\hc{referred as \textit{skeleton instances}}), we propose a \textit{\textbf{skeleton prototype contrastive learning scheme} (SPC)}, which clusters correlative skeleton instances and contrasts their inherent similarity with the most representative skeleton features (\hc{referred as \textit{skeleton prototypes}}) 
 to learn general discriminative skeleton representations in an unsupervised manner. By maximizing the similarity of skeleton instances to their corresponding prototypes and their dissimilarity to other prototypes, SPC encourages the model to capture more discriminative skeleton features and class-related semantics ($e.g.,$ intra-class similarity) for person re-ID without using any label. 
 
 
An earlier and preliminary version of this work was reported in \cite{rao2021multi}. Compared with \cite{rao2021multi} that performs supervised skeleton representation for person re-ID, this paper focuses on the unsupervised re-ID problem under general settings. In this work, we explore unified and \hc{generalized} multi-level skeleton graphs to extend earlier graph representations to different skeleton data, and propose a novel skeleton prototype contrastive learning paradigm to mine discriminative skeleton features with the proposed unlabeled skeleton graphs for person re-ID. To our \hc{best} knowledge, this is also the first work that explores a generic unsupervised learning paradigm for skeleton-based person re-ID. \hc{We also} devise a new full-level collaborative relation layer to capture more comprehensive relations between different-level body components, and further explore a multi-level graph feature fusion strategy to enhance graph semantics and global pattern learning. To validate the above improvement, this work conducts comprehensive experiments on four public person re-ID datasets, and carries out more detailed analysis on different \hc{person re-ID tasks} ($e.g.,$ multi-view person re-ID) under more general scenarios. Furthermore, we propose two evaluation metrics, \textit{mean intrA-Class Tightness (mACT)} and \textit{mean inteR-Class Looseness (mRCL)}, to quantitatively evaluate the tightness of same-class representations and the looseness of different-class representations, so as to showcase the effectiveness of skeleton representations learned by \hc{the proposed approach}.

\hc{Our main contributions are} summarized as follows:
\begin{itemize}
\item We devise unified multi-level graphs to model 3D skeletons, and propose a novel Skeleton Prototype Contrastive learning paradigm with Multi-level Graph Relation modeling (SPC-MGR) to learn an effective representation from unlabeled skeleton data for unsupervised person re-ID.

\item We propose multi-head structural relation layer (MSRL) to capture relations of neighbor body components, and devise full-level collaborative relation layer (FCRL) to infer collaboration between different-level components, so as to learn more structural semantics and unique patterns.

\item We propose a skeleton prototype contrastive learning (SPC) scheme to encourage capturing representative discriminative skeleton features and high-level class-related semantics from unlabeled skeleton data for person re-ID.

\item We propose mean intra-class tightness (mACT) and mean inter-class looseness (mRCL) to evaluate skeleton representation learning, and show the effectiveness of skeleton representations learned by our unsupervised approach.


\item Extensive experiments show that the proposed SPC-MGR outperforms several state-of-the-art skeleton-based methods on four person re-ID benchmarks, and is also highly effective when applied to skeleton data estimated from large-scale RGB videos under more general re-ID settings.

\end{itemize}

The rest of our paper is organized as follows. Sec. \ref{related} provides an introduction for related works in the literature. Sec. \ref{approach} presents each technical module of the proposed approach. Sec. \ref{experiments} elucidates the details of experiments, and comprehensively compares our approach with existing solutions. Sec. \ref{discussion} conducts ablation study and extensive analysis of the proposed approach. Sec. \ref{conclusion} concludes this paper. Sec. \ref{ethical} provides a statement for potential ethical issues.

\section{Related Works}
\label{related}
\subsection{Skeleton-Based Person Re-Identification} 
\subsubsection{Hand-Crafted Methods}
Early skeleton-based works extract hand-crafted descriptors in terms of certain geometric, morphological or anthropometric attributes of human body. 
 Barbosa \textit{et al.} \cite{barbosa2012re} compute 7 Euclidean distances between the floor plane and joint or joint pairs to construct a distance matrix, which is learned by a quasi-exhaustive strategy to extract discriminative features for person re-ID. 
 Munaro \textit{et al.} \cite{munaro2014one} and Pala \textit{et al.} \cite{pala2019enhanced} further extend them to 13 ($D_{13}$) and 16 skeleton descriptors ($D_{16}$) respectively, and leverage support vector machine (SVM), $k$-nearest neighbor (KNN) or Adaboost classifiers for person re-ID. Since such solutions using 3D skeletons alone are hard to achieve satisfactory performance, they usually combine other modalities such as 3D point clouds \cite{munaro20143d} and 3D face descriptors \cite{pala2019enhanced} to improve person re-ID accuracy. 
\subsubsection{Supervised and Self-Supervised Methods}
Most recently, a few works exploit deep learning paradigms to learn gait representations from skeleton data for person re-ID in a supervised or self-supervised manner.
Liao \textit{et al.} \cite{liao2020model} propose PoseGait, which feeds 81 hand-crafted pose features of 3D skeletons into CNN for human recognition. Rao \textit{et al.} \cite{rao2020self} devise a self-supervised attention-based gait encoding (AGE) model with multi-layer LSTM to encode gait features from unlabeled skeleton sequences, and then fine-tune the learned features with the supervision of labels for person re-ID. In \cite{rao2021self}, they further propose a locality-awareness approach (SGELA) that combines various pretext tasks ($e.g.,$ reverse sequential reconstruction) and contrastive learning scheme to enhance self-supervised gait representation learning for the person re-ID task.
The latest self-supervised work is SM-SGE \cite{rao2021sm}, which utilizes a skeleton graph based reconstruction and inference mechanism to encode discriminative skeleton structure and motion features for \hc{the person re-ID task.}

Our work differs \hc{in following aspects.} We for the first time propose an \textit{unsupervised} learning paradigm to learn discriminative skeleton representations from unlabeled skeleton data to directly perform person re-ID. Our approach requires neither any form of feature engineering ($e.g.,$ hand-crafted skeleton descriptors \cite{barbosa2012re,munaro2014one,pala2019enhanced}) nor any skeleton label for training model \cite{liao2020model,rao2021multi} or fine-tuning representations \cite{rao2020self,rao2021self,rao2021sm}. Unlike most existing works that directly represent skeletons as sequences of body joints \cite{rao2020self,rao2021self} or pose descriptors \cite{liao2020model}, this work not only explores more comprehensive body-structure modeling by devising unified and \hc{generalized} multi-level skeleton graphs, but also systematically mines valuable body-component relations at various levels. Besides, the proposed approach needs neither self-supervised pretext tasks ($e.g.,$ reconstruction \cite{rao2021sm}) for pre-training or supervised classification networks for discriminative representation learning. Instead, it exploits the unsupervised skeleton prototype contrastive learning to mine representative skeleton features and class-related semantics from unlabeled 3D skeleton data. To the best of our knowledge, this is also the first attempt to leverage skeleton graph based relation modeling and contrastive learning to realize \hc{both skeleton representation learning and unsupervised person re-ID.}

\subsection{Contrastive Learning}
Contrastive learning has recently achieved great success in many self-supervised and unsupervised learning tasks \cite{hadsell2006dimensionality,wu2018unsupervised,oord2018representation,chen2020big,ge2020selfpaced,li2021prototypical,rao2021self}. Its general objective is to learn effective data representations by pulling closer positive pairs and pushing apart negative pairs in the feature space using contrastive losses, which are often designed based on certain auxiliary tasks ($e.g.,$ similarity metrics learning). For example, 
Wu \textit{et al.} \cite{wu2018unsupervised} devise an instance-level discrimination method in the form of exemplar task \cite{dosovitskiy2014discriminative} to perform image contrastive learning with noise-contrastive estimation loss (NCE) \cite{gutmann2010noise}. In \cite{oord2018representation}, contrastive predictive coding (CPC) based on a probabilistic contrastive loss (InfoNCE) is proposed to learn general representations for different domains. To optimize representation learning ($e.g.,$ consistency) in memory bank based contrastive methods \cite{zhuang2019local,tian2019contrastive,misra2020self}, some recent end-to-end works  \cite{ye2019unsupervised,ji2019invariant,chen2020a} utilize all samples of the current mini-batch to generate negative instance features, while the momentum-based approach \cite{he2020momentum} further explores the use of momentum-updated encoder and queue dictionary to improve consistency of both encoder and instance features.
The latest PCL \cite{li2021prototypical} integrates both contrastive learning and clustering into an expectation-maximization (EM) framework, which is highly efficient on unsupervised visual representation learning and inspires our work for 3D skeletons.

Our work differs from previous studies in following aspects. The proposed skeleton prototype contrastive learning scheme is devised to mine most representative and discriminative skeleton features ($i.e.,$ skeleton prototypes) from unlabeled multi-level skeleton graph representations of 3D skeleton sequences ($i.e.,$ skeleton instances). In this scheme, the skeleton instances and generated skeleton prototypes are exploited as the contrasting instances, which fundamentally differs from existing works \cite{chen2020a,he2020momentum,chen2020big,li2021prototypical} that use augmented samples of images as instances. 
The goal of skeleton prototype contrastive learning scheme is to maximize skeleton instances' similarity with their prototypes and dissimilarity to other prototypes using the proposed clustering-contrasting strategy, which does not require any extra memory or sampling mechanism \cite{ge2020selfpaced,li2021prototypical} and can directly learn effective skeleton representations for the person re-ID task without introducing any ground-truth label from source data domains.

\section{The Proposed Approach}
\label{approach}
\begin{figure*}[ht]
    \centering
    \scalebox{0.565}{
    \includegraphics{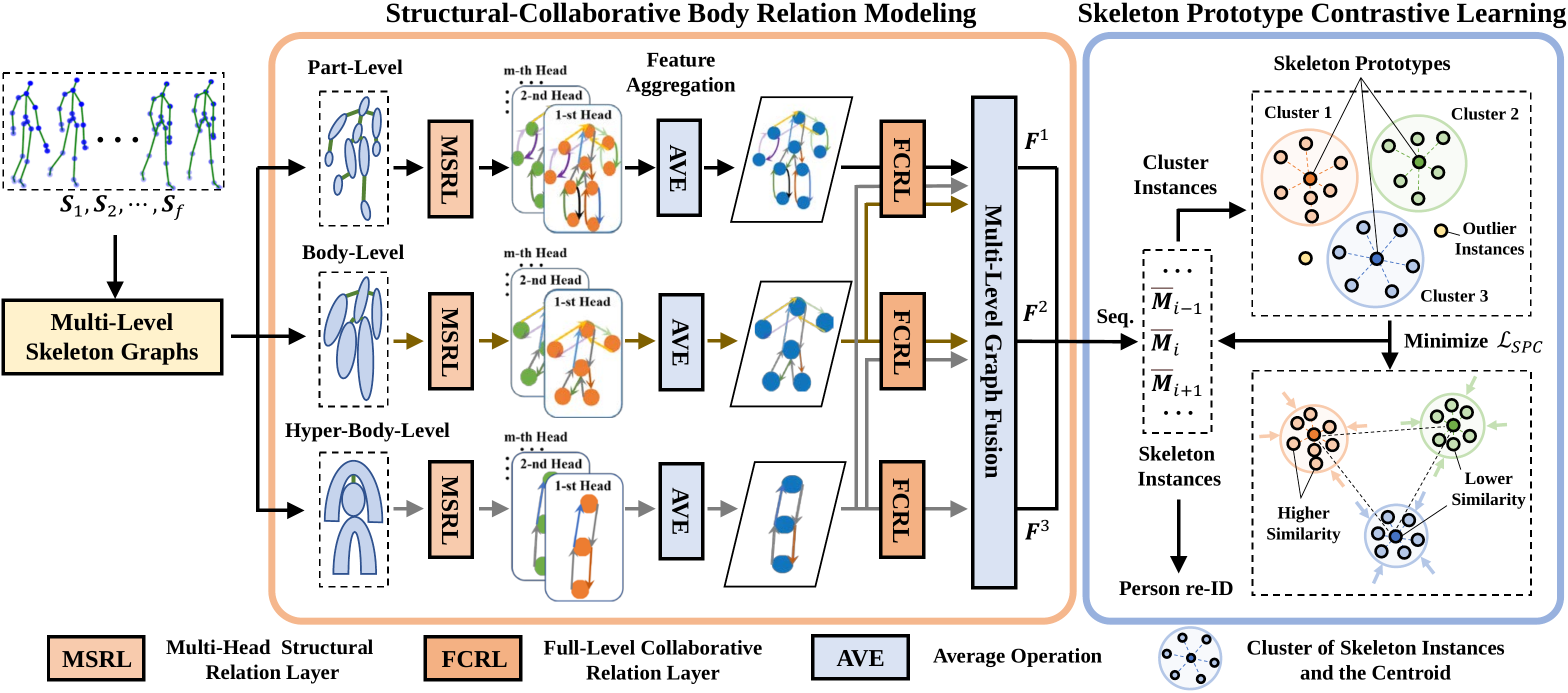}
    }
    \caption{Schematic diagram of SPC-MGR. \hc{Firstly,} each 3D skeleton of an input sequence $\boldsymbol{S}_{1},\boldsymbol{S}_{2},\cdots,\boldsymbol{S}_{f}$ is represented with part-level, body-level, and hyper-body-level graphs. \hc{Secondly,} we employ multi-head structural relation layers (MSRL) to capture structural relations of neighbor nodes in each graph, and averagely aggregate features learned by multiple heads to obtain node representations. Then, full-level collaborative relation layers (FCRL) infer the dynamic collaborative relations among the same-level and different-level body components, which are exploited to integrate key graph features into multi-level skeleton graph representations $\boldsymbol{F}^{1}$, $\boldsymbol{F}^{2}$, and $\boldsymbol{F}^{3}$. Next, we perform clustering on skeleton instances, which are sequence-level (``Seq.'') multi-level skeleton graph representations, to generate clusters and corresponding skeleton prototypes. Finally, during skeleton prototype contrastive learning, we enhance the similarity of instances belonging to the same prototype and maximize their dissimilarity to other prototypes by minimizing contrastive loss $\mathcal{L}_{\text {SPC}}$. The learned skeleton graph representations are exploited to perform person re-ID.
    }
    \label{model_overview}
\end{figure*}

Suppose that a 3D skeleton sequence $\boldsymbol{S}_{1:f}\!=\!(\boldsymbol{S}_1,\cdots,\boldsymbol{S}_{f})\in \mathbb{R}^{f \times J \times D}$, where $\boldsymbol{S}_{t}\in \mathbb{R}^{J \times D}$ is the $t^{th}$ skeleton with $J$ body joints and $D\!=\!3$ dimensions. Each skeleton sequence $\boldsymbol{S}_{1:f}$ corresponds to an ID label $y$, where $y\in \{1, \cdots, C\}$ and $C$ is the number of different persons. 
The training set $\Phi_{t}=\left\{\boldsymbol{S}^{t, i}_{1:f}\right\}_{i=1}^{N_{1}}$, probe set $\Phi_{p}=\left\{\boldsymbol{S}^{p, i}_{1:f}\right\}_{i=1}^{N_{2}}$, and gallery set $\Phi_{g}=\left\{\boldsymbol{S}^{g, i}_{1:f}\right\}_{i=1}^{N_{3}}$ contain $N_{1}$, $N_{2}$, and $N_{3}$ skeleton sequences of different persons under varying views or scenes. 
Our goal is to learn an embedding function $\psi(\cdot)$ that maps $\Phi_{p}$ and $\Phi_{g}$ to effective skeleton representations $\{\overline{\boldsymbol{M}}^{p,i}\}_{i=1}^{N_{2}}$ and $\{\boldsymbol{\overline{\boldsymbol{M}}}^{g,j}\}_{j=1}^{N_{3}}$ \textit{without using any label}, such that the representation $\overline{\boldsymbol{M}}^{p,i}$ in the probe set can match the representation $\overline{\boldsymbol{M}}^{g,j}$ of the same identity in the gallery set. 
The overview of the proposed approach is given in Fig. \ref{model_overview}, and we present the details of each technical component below.


\subsection{Unified Multi-Level Skeleton Graphs}
\label{graph_construct}
\begin{figure}
    \centering
    \scalebox{0.62}{
    \includegraphics{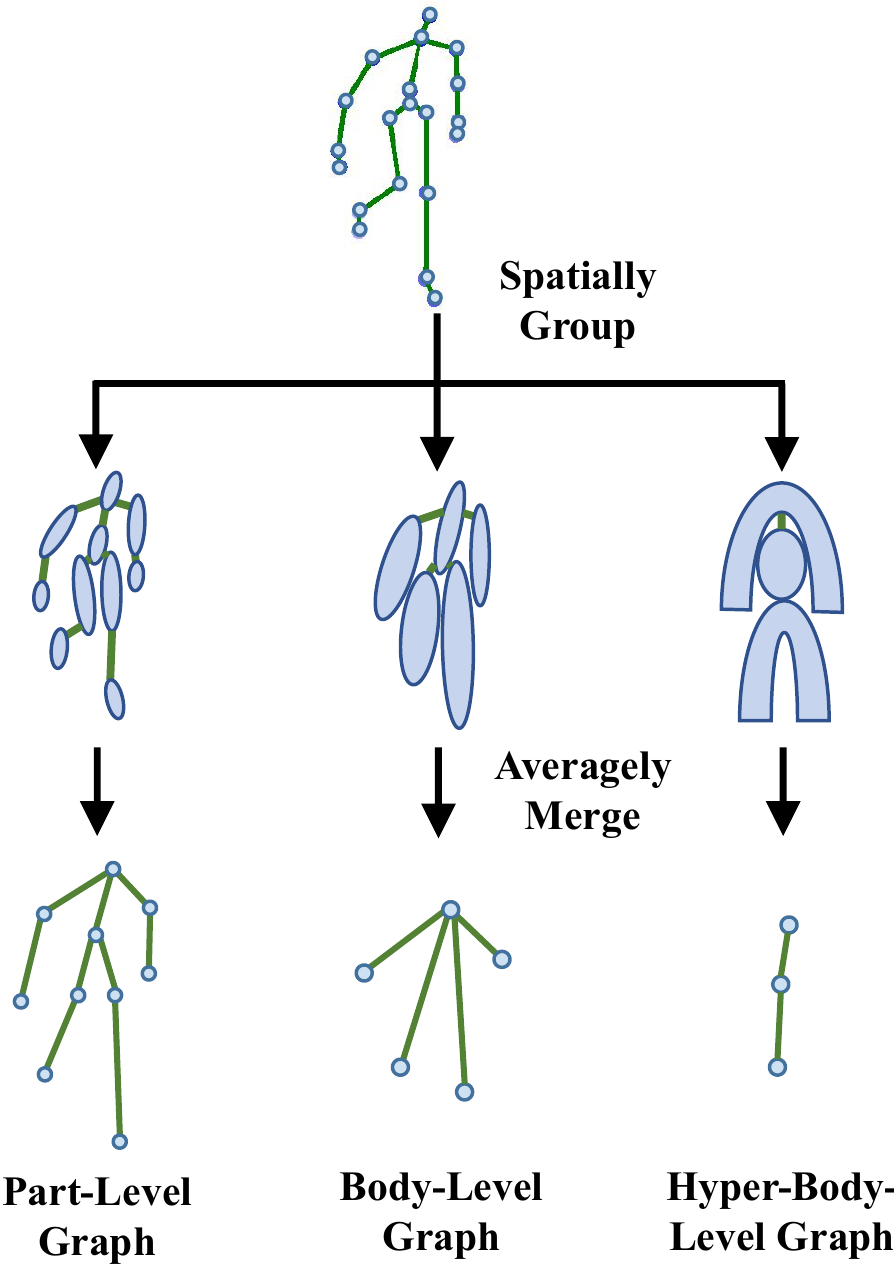}
    }
    \caption{Three graph levels for a skeleton. We spatially divide human body into 10, 5 and 3 partitions to construct part-level, body-level, and hyper-body-level graphs, and averagely merge internal body joints into nodes.}
    \label{MG}
\end{figure}

Inspired by the fact that human motion can be decomposed into movements of functional body-components ($e.g.,$ legs, arms) \cite{winter2009biomechanics}, we spatially group \hc{skeleton joints to be \textit{higher level} body components at their centroids.} Specifically, we first divide human skeletons into several partitions \textit{from coarse to fine}. Based on the nature of body structure, we specify the location of each body partition and \hc{its corresponding skeleton joints} of different sources ($e.g.,$ datasets).
Then, we adopt the weighted average of body joints in the same partition as the node of higher level body component and use its physical connections as edges, so as to build unified skeleton graphs for an input skeleton.
As shown in Fig. \ref{MG}, we construct three levels of skeleton graphs, namely \textit{part-level}, \textit{body-level} and \textit{hyper-body-level} graphs for each skeleton $\boldsymbol{S}$, which can be represented as $\mathcal{G}^1$, $\mathcal{G}^2$ and $\mathcal{G}^3$ respectively.
Each graph $\mathcal{G}^{l}(\mathcal{V}^{l}, \mathcal{E}^{l})$  ($l\in\{1,2,3\}$) consists of nodes $\mathcal{V}^{l}=\{\boldsymbol{v}^{l}_{1}, \boldsymbol{v}^{l}_{2}, \cdots,\boldsymbol{v}^{l}_{n_l}\}$, $\boldsymbol{v}^{l}_{i}\in\mathbb{R}^{D}$, $i\in\{1,\cdots,n_l\}$ and edges $\mathcal{E}^{l}=\{e^{l}_{i,j}\ | \boldsymbol{v}^{l}_{i}, \boldsymbol{v}^{l}_{j}\!\in\!\mathcal{V}^{l}\}$, $e^{l}_{i,j}\in\mathbb{R}$. Here $\mathcal{V}^{l}$ and $\mathcal{E}^{l}$ denote the set of nodes corresponding to different body components and the set of their internal connection relations, respectively. $n_l$ denotes the number of nodes in $\mathcal{G}^{l}$. 
More formally, we define a graph's adjacency matrix as $\mathbf{A}^{l} \in \mathbb{R}^{n_l \times n_l}$ to represent structural relations among $n_l$ nodes. We compute the \textit{normalized} structural relations between node $i$ and its neighbors, $i.e.$, $\sum_{j \in \mathcal{N}_{i}}\mathbf{A}^{l}_{i,j}=1$, where $\mathcal{N}_{i}$ denotes the neighbor nodes of node $i$. $\mathbf{A}^{l}$ is adaptively learned to capture flexible structural relations in the training stage.

\textbf{Remarks:} 
The proposed unified multi-level skeleton graphs can be generalized \hc{into} different skeleton datasets, which enables the pre-trained model to be directly transferred across different domains for generalized person re-ID (see Sec. \ref{generalized_reid}). Besides, they can be extended to skeleton data estimated from RGB videos to learn effective person re-ID representations (see Sec. \ref{estimated_skeletons}).

\subsection{Structural-Collaborative Body Relation Modeling}
\label{structural-collaborative}
The physical connections of body structure typically endow body components in a local partition with higher correlations, while components of different parts may act collaboratively in various global patterns during motion \cite{aggarwal1998nonrigid}. To exploit such internal relations to mine rich body-structure features and unique motion characteristics from skeletons, we propose the multi-level structural relation layer (MSRL) and full-level collaborative relation layer (FCRL) to model the structural and collaborative relations of body components from multi-level skeleton graphs as follows.

\subsubsection{Multi-Head Structural Relation Layer}
To capture latent body structural information and learn an effective representation for each body-component node in skeleton graphs, we propose to focus on features of structurally-connected neighbor nodes, which enjoy higher correlations (\hc{referred as \textbf{\textit{structural relations}}}) than distant pairs. For instance, adjacent nodes usually have closer spatial positions and similar motion tendency. Therefore, we devise a \textit{multi-head structural relation layer} (MSRL) to learn relations of neighbor nodes and aggregate the most correlative spatial features to represent each body-component node.

We first devise a basic \textit{structural relation head} based on the graph attention mechanism \cite{velickovic2018graph}, which can focus on more correlative neighbor nodes by assigning larger attention weights, to capture the internal relation $e^{l}_{i, j}$ between adjacent nodes $i$ and $j$ in the same graph \hc{as:}
\begin{equation}
\label{eq_1}
e^{l}_{i,j}=\text {LeakyReLU}\!\left({\mathbf{W}^{l}_{r}}^{\mathsf{T}}\left[\mathbf{W}^{l}_{v}{\boldsymbol{v}}^{l}_{i} \| \mathbf{W}^{l}_{v}{\boldsymbol{v}}^{l}_{j}\right]\right)
\end{equation}
where $\mathbf{W}^{l}_{v}\in \mathbb{R}^{D\times D_{h}}$ denotes the weight matrix to map the $l^{th}$ level node features $\boldsymbol{v}^{l}_{i}\in \mathbb{R}^{D}$ into a higher level feature space $ \mathbb{R}^{D_h}$, $\mathbf{W}^{l}_{r}\in \mathbb{R}^{2D_{h}}$ is a learnable weight matrix to perform relation learning in the $l^{th}$ level graph,  $\|$ indicates concatenating features of two nodes, and LeakyReLU$(\cdot)$ is a non-linear activation function. Then, to learn flexible structural relations to focus on more correlative nodes, we
normalize relations using the $\operatorname{softmax}$ function as follows:
\begin{equation}
\label{eq_2}
\mathbf{A}^{l}_{i,j}= \operatorname{softmax}_{j}\left(e^{l}_{i,j} \right) = \frac{\exp \left(e^{l}_{i,j} \right)}{\sum_{k \in \mathcal{N}_{i}} \exp \left(e^{l}_{i,k} \right)}
\end{equation}
where $\mathcal{N}_{i}$ denotes directly-connected neighbor nodes (including $i$) of node $i$ in graph. We use structural relations $\mathbf{A}^{l}_{i,j}$ to aggregate features of most relevant nodes to represent node $i$:
\begin{equation}
\label{eq_aggr}
\boldsymbol{\overline{v}}^{l}_{i}=\sigma\left(\sum_{j \in \mathcal{N}_{i}} \mathbf{A}^{l}_{i,j} \mathbf{W}^{l}_{v} \boldsymbol{v}^{l}_{j}\right)
\end{equation}
where $\sigma(\cdot)$ is a non-linear function and $\boldsymbol{\overline{v}}^{l}_{i}\in \mathbb{R}^{D_{h}}$ is feature representation of node $i$ computed by a structural relation head.

\begin{figure}[t]
    \centering
    \scalebox{0.61}{
    \includegraphics{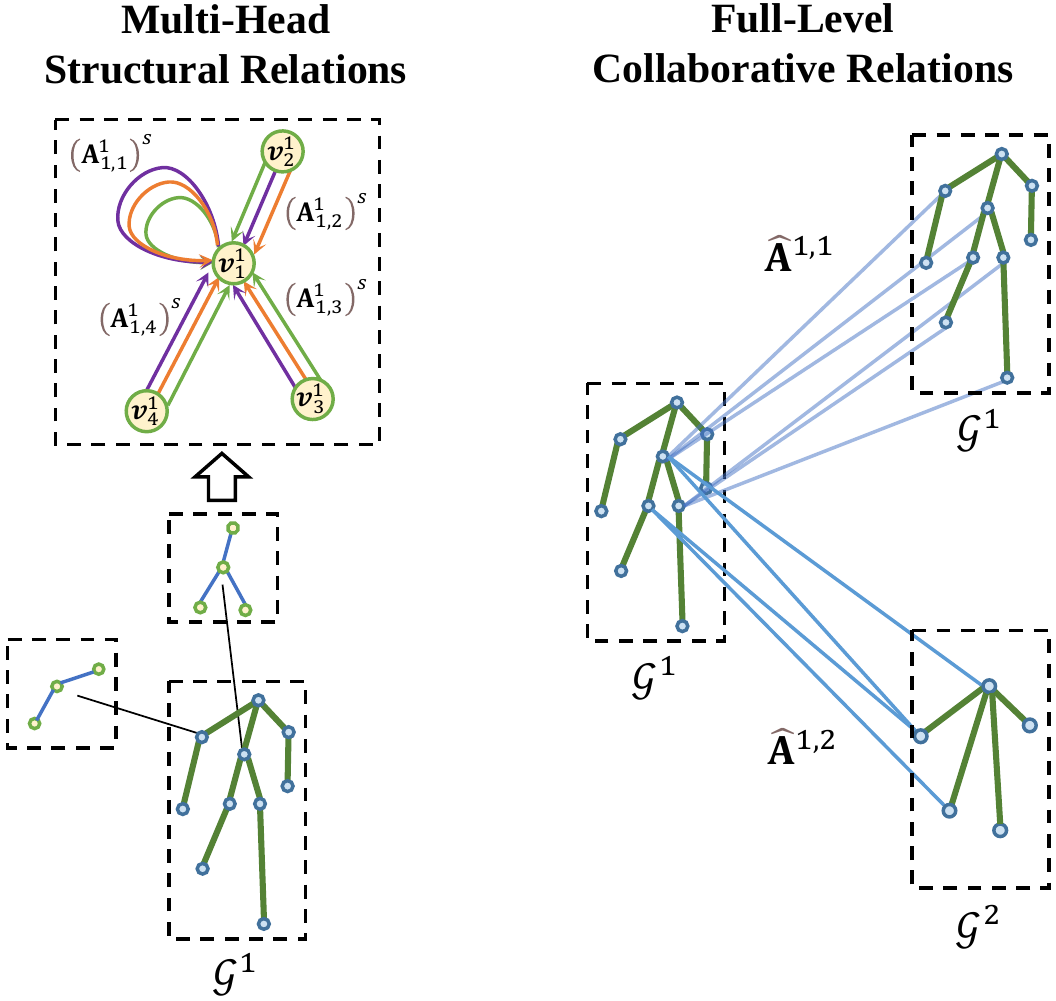}
    }
    \caption{Examples of multi-head structural relations in $\mathcal{G}^{1}$ (left) and full-level collaborative relations among graphs ($\mathcal{G}^{1}$, $\mathcal{G}^{1}$ and $\mathcal{G}^{1}$, $\mathcal{G}^{2}$) (right).}
    \label{relation_learn}
\end{figure}

To sufficiently capture potential structural relations ($e.g.,$ position similarity and movement correlations) between each node and its neighbor nodes, we employ \textit{multiple structural relation heads}, each of which independently executes the same computation of Eq. \ref{eq_aggr} to learn a potentially different structural relation\hc{, as shown in Fig. \ref{relation_learn}}. We \textit{averagely aggregate} features learned by $m$ different structural relation heads as the representation of node $i$ as follows:
\begin{equation}
\label{node_repre}
\boldsymbol{\widehat{v}}^{l}_{i}= \frac{1}{m} \sum^{m}_{s=1}\sigma\left( \sum_{j \in \mathcal{N}_{i}} (\mathbf{A}^{l}_{i,j})^{s} (\mathbf{W}^{l}_{v})^{s} \boldsymbol{v}^{l}_{j}\right)
\end{equation}where $\boldsymbol{\widehat{v}}^{l}_{i}\in\mathbb{R}^{D_{h}}$ denotes the multi-head feature representation of node $i$ in $\mathcal{G}^{l}$, $m$ is the number of structural relation heads, $(\mathbf{A}^{l}_{i,j})^{s}\in\mathbb{R}$ represents the structural relation between node $i$ and $j$ computed by the $s^{th}$ structural relation head, and $(\mathbf{W}^{l}_{v})^s$ denotes the corresponding weight matrix to perform feature mapping in the $s^{th}$ head. Here we use \textit{average} rather than concatenation operation to reduce feature dimension and allow for more structural relation heads. MSRL enables our model to capture the relations of correlative neighbor nodes (\hc{see Eq. \ref{eq_1} and \ref{eq_2}}) and integrates key spatial features into node representations of each graph (\hc{see Eq. \ref{eq_aggr} and \ref{node_repre}}). However, it only considers the local relations of the same-level components in graphs and is insufficient to capture global collaboration between different level body components, which motivates us to propose the full-level collaborative relation layer.

\subsubsection{Full-Level Collaborative Relation Layer}
Motivated by the natural property of human walking, \hc{$i.e.$, gait,} which could be represented by the dynamic cooperation among body joints or between different body components \cite{murray1964walking}, we expect our model to infer the degree of collaboration (referred as \textbf{\textit{collaborative relations}}) among body-component nodes in multi-level graphs, so as to capture more unique and recognizable walking patterns from the motion of skeletons.
For this purpose, we propose a \textit{full-level collaborative relation layer} (FCRL) to capture relations between a node and all motion-related nodes of the \textit{same level} and that between a node and its spatially corresponding \textit{higher level} body component or other potential components.
\hc{As shown in Fig. \ref{model_overview} and Fig. \ref{relation_learn},} we compute collaborative relation matrix
 $\mathbf{\widehat{A}}^{a,b} \in \mathbb{R}^{n_{a} \times n_{b}}$ ($a,b \in \{1, 2, 3\}, a\leq b$) between the $a^{th}$ level nodes $\mathcal{V}^{a}$ and the $b^{th}$ level nodes $\mathcal{V}^{b}$ as following: 
\begin{equation}
\label{collaborative_eq}
\mathbf{\widehat{A}}^{a, b}_{i, j}\!=\!\operatorname{softmax}_{j}\left({\boldsymbol{\widehat{v}}^{a}_{i}}^{\top} \boldsymbol{\widehat{v}}^{b}_{j}\right)\!=\!\frac{\exp \left({\boldsymbol{\widehat{v}}^{a}_{i}}^{\top} \boldsymbol{\widehat{v}}^{b}_{j}\right)}{\sum^{n_{b}}_{k=1} \exp \left({\boldsymbol{\widehat{v}}^{a}_{i}}^{\top} \boldsymbol{\widehat{v}}^{b}_{k}\right)}
\end{equation}
where $\mathbf{\widehat{A}}^{a, b}_{i, j}$ is the collaborative relation between node $i$ in $\mathcal{G}^a$ and node $j$ in $\mathcal{G}^{b}$. Here we use the inner product of multi-head node feature representations (see Eq. \ref{node_repre}) \hc{that retain key spatial information of nodes} to measure the degree of collaboration. 
Instead of merely considering body relations between adjacent level graphs \cite{rao2021multi}, FCRL can capture the global collaborative relations among both \textit{adjacent} and \textit{non-adjacent} graphs, and meanwhile provides full collaboration inference between a node and all potential motion-correlated nodes in the same graph.

\subsubsection{Multi-Level Graph Feature Fusion}
To enhance structural semantics of multiple graphs ($e.g.,$ global graph patterns) and adaptively integrate key correlative features in component collaboration, we exploit collaborative relations to fuse body-component node features across different spatial levels. We update the node representation ($\widehat{\boldsymbol{v}}_{i}^{a}$) of $a^{th}$ level graph by fusing collaborative node features ($\boldsymbol{\widehat{v}}^{b}_{j}$) learned from different graphs:
\begin{equation}
\label{eq_fuse}
\widehat{\boldsymbol{v}}_{i}^{a} \ \leftarrow \ \widehat{\boldsymbol{v}}_{i}^{a}+ \sum^{3}_{b=a}\left(\lambda^{a,b}_{C}\sum^{n_{b}}_{j=1}\mathbf{\widehat{A}}^{a, b}_{i, j}\ \mathbf{W}^{a,b}_{C} \  \boldsymbol{\widehat{v}}^{b}_{j}\right) 
\end{equation}
where $\mathbf{W}^{a,b}_{C}\in\mathbb{R}^{D_{h}\times D_{h}}$ is a learnable weight matrix to integrate features of collaborative node $\boldsymbol{\widehat{v}}^{b}_{j}$ of $b^{th}$ level into $a^{th}$ level \hc{node} $\widehat{\boldsymbol{v}}_{i}^{a}$. $n_{b}$ denotes the number of nodes in the $b^{th}$ level graph, and $\lambda^{a,b}_{C}$ represents the fusion coefficient between $a^{th}$ level and $b^{th}$ level graphs, which can be adjusted according to their inherent correlations ($e.g.,$ level similarity). We denote the fused $l^{th}$ level graph features of the $i^{th}$ skeleton as $\boldsymbol{F}^{l}_{i} \in \mathbb{R}^{n_{l}\times D_{h}}$ by concatenating all node representations. Inspired by \cite{rao2021sm}, we retain graph representations of each individual level and adopt their \textit{concatenation} to represent a skeleton as follows:
\begin{equation}
\label{ske_rep}
\boldsymbol{M}_{i}=[\boldsymbol{F}^{1}_{i};\boldsymbol{F}^{2}_{i}; \boldsymbol{F}^{3}_{i}]
\end{equation}
where $\boldsymbol{M}_{i}\in\mathbb{R}^{(n_{1}+n_{2}+n_{3})\times D_{h}}$ is the multi-level graph representation of the $i^{th}$ skeleton $\boldsymbol{S}_{i}$, and  $[;]$ indicates the concatenation of graph features. By combining all graph-level representations that integrate structural and collaborative body relation features (see Eq. \ref{eq_1}-Eq. \ref{eq_fuse}), we encourage the model to capture richer features of body structure and skeleton patterns at various levels. \hc{We show the effectiveness of the proposed structural-collaborative body relation learning by the relation visualization in Sec. \ref{body_relation_analysis} and Appendix.}


\subsection{Skeleton Prototype Contrastive Learning Scheme}
\label{SC_contrastive_learning}
As skeletons of the same individual typically share highly similar body attributes ($e.g.,$ anthropometric attributes) and unique walking patterns \cite{murray1964walking}, it is natural to consider mining the most \textit{typical} attributes or patterns to \hc{identify the same person from others.}
To achieve this goal and encourage the model to capture more high-level skeleton semantics ($e.g.,$ class-related patterns), we propose a Skeleton Prototype Contrastive learning (SPC) scheme to focus on the most \textit{representative} skeleton graph features (\hc{referred as \textbf{\textit{skeleton prototypes}}})  of pedestrians and exploit their inherent \textit{similarity} and \textit{dissimilarity} with other \textit{unlabeled} graph representations (referred as \textbf{\textit{skeleton instances}}) to learn general and discriminative representations of each individual for the person re-ID task.



\subsubsection{Skeleton Prototype Generation}
\label{protptype_generation}
Given multi-level graph representations ($\boldsymbol{M}_{1},\cdots,\boldsymbol{M}_{f}$) of an input skeleton sequence ($\boldsymbol{S}_{1:f}=\left(\boldsymbol{S}_{1},\cdots,\boldsymbol{S}_{f}\right)$), we first integrate graph features into a \textit{sequence-level} skeleton graph representation:
\begin{equation}
\label{seq_rep}
\overline{\boldsymbol{M}}=\frac{1}{f}\sum^{f}_{i=1}w_{i}\boldsymbol{M}_{i}
\end{equation}
where $\overline{\boldsymbol{M}}$ is the multi-level graph representation of skeleton sequence $\boldsymbol{S}_{1:f}$, which incorporates structural-collaborative features and temporal dynamics of $f$ consecutive multi-level skeleton graphs, and $w_{i}$ denotes the importance of $i^{th}$ skeleton graph representation. Here we assume that each skeleton equally contributes to representing graph features of a sequence, $i.e.$, $w_{i}=1$. For clarity, we use $\overline{\mathbb{M}}=\{\overline{\boldsymbol{M}}_{i}\}_{i=1}^{N_{1}}$ to represent multi-level graph representations of skeleton sequences in the training set $\Phi_{t}$, which are exploited as \textit{skeleton instances} in the proposed SPC scheme. 

Then, to gather skeleton instances $\overline{\mathbb{M}}$ that contain similar features to find the representative skeleton prototypes, we leverage the DBSCAN algorithm \cite{ester1996density}, which can discover clusters with arbitrary shapes or semantics, to perform clustering as: 
\begin{equation}
\label{dbscan}
\text{DBSCAN}(\overline{\mathbb{M}})\longrightarrow \overline{\mathbb{M}}^{1}, \overline{\mathbb{M}}^{2}, \cdots,\overline{\mathbb{M}}^{z},
\overline{\mathbb{M}}^{o}
\end{equation}
where $\overline{\mathbb{M}}= \overline{\mathbb{M}}^{1}\cup\overline{\mathbb{M}}^{2}\cup \cdots\cup\overline{\mathbb{M}}^{z}\cup\overline{\mathbb{M}}^{o}$, $z$ is the number of clusters ($i.e.,$ pseudo classes),
\hc{$\overline{\mathbb{M}}^{k}=\{\overline{\boldsymbol{M}}^{k}_{i}\}_{i=1}^{x_{k}}$, $k\in\{1,\cdots,z\}$, is the cluster} that contains $x_k$ instances belonging to the $k^{th}$ pseudo class, and $\overline{\mathbb{M}}^{o}=\{\overline{\boldsymbol{M}}^{o}_{i}\}_{i=1}^{x_{o}}$ denotes the set of outlier instances that do not belong to any cluster. We compute the centroid of each cluster, which \textit{averagely aggregates} features of skeleton instances in the cluster, to obtain corresponding skeleton prototype:
\begin{equation}
\label{prototype}
\boldsymbol{P}^{k}=\frac{1}{x_k}\sum^{x_k}_{i=1}\overline{\boldsymbol{M}}^{k}_{i}
\end{equation}
where \hc{$\overline{\boldsymbol{M}}^{k}_{i}\in\mathbb{R}^{(n_{1}+n_{2}+n_{3})\times D_{h}}$ is the $i^{th}$ skeleton instance} in the $k^{th}$ cluster, and $\boldsymbol{P}^{k}$ denotes the $k^{th}$ skeleton prototype. 



\subsubsection{Skeleton Prototype Contrastive Learning}
\label{contrastive_learning}
To focus on the typical and discriminative features of skeleton prototypes as well as facilitate learning high-level skeleton semantics from different prototypes, we propose to enhance the inherent similarity of a skeleton instance to corresponding skeleton prototype and maximize its dissimilarity to other skeleton prototypes with a skeleton prototype contrastive loss as:
\begin{equation}
\label{SPC}
    \mathcal{L}_{\text {SPC}}=\frac{1}{N}\sum_{k=1}^{z}\sum_{i=1}^{x_k}-\log \frac{\exp \left(\overline{\boldsymbol{M}}^{k}_{i} \cdot \boldsymbol{P}^{k} / \tau \right)}{\sum_{j=1}^{z} \exp \left(\overline{\boldsymbol{M}}^{k}_{i} \cdot \boldsymbol{P}^{j} / \tau \right)}
\end{equation}
where $N$ represents the number of all training instances, $z$ denotes the number of skeleton prototypes, $x_{k}$ is the number of skeleton instances belonging to the $k^{th}$ prototype $\boldsymbol{P}^{k}$, and $\tau$ represents the temperature for contrastive learning, where higher value of $\tau$ produces a softer probability distribution over prototypes and retains more similar information among clusters \cite{hinton2015distilling}. The skeleton prototype contrastive loss $\mathcal{L}_{\text {SPC}}$ can be viewed as a special form of InfoNCE loss \cite{oord2018representation}, which aims to maximize the dot product based similarity \hc{between the query for each skeleton instance and its positive key for the same prototype while maximizing the dissimilarity to negative keys for other prototypes.}

\subsubsection{Analysis of Skeleton Prototype Contrastive Learning}
The objective of skeleton prototype contrastive learning is to maximize the intra-prototype similarity and inter-prototype dissimilarity to mine discriminative skeleton features and identity-related semantics in an unsupervised manner. 
 Such learning process can be interpreted as to simultaneously enhance the tightness of skeleton instances belonging to the same prototype and the looseness among different prototypes. 
 Considering that a better representation space should have smaller intra-class distance (higher tightness) and larger inter-class distance (higher looseness), we 
can measure the tightness and looseness levels of the learned representations with regard to \textit{ground-truth} classes to verify whether the model has learned effective features for different classes from the unlabeled data. To this end, we propose two metrics named \textit{mean intrA-Class Tightness (mACT)} and \textit{mean inteR-Class Looseness (mRCL)} to quantitatively evaluate the \hc{merits} of the learn skeleton representations.

\begin{algorithm*}
\setstretch{1.3}
\caption{Main algorithm of SPC-MGR}\label{algorithm_1}
\KwIn{Unlabeled training skeleton sequences $\Phi_{t}=\left\{\boldsymbol{S}^{t, i}_{1:f}\right\}_{i=1}^{N_{1}}$,
maximum distance $\epsilon$ and minimum sample amount $a_{min}$ for the DBSCAN algorithm, initialized embedding function $\psi(\cdot)$ for skeleton representation learning, temperature $\tau$}
\KwOut{Embedding function $\psi(\cdot)$}
\While{not MaxEpoch}{ \tcp{Alternate clustering and contrastive learning}
$\overline{\boldsymbol{M}}_{i}=\psi(\boldsymbol{S}^{t,i}_{1:f})$ \quad \quad \quad \quad \quad \quad \quad \quad \quad  \tcp{Encode skeleton sequences to get instances $\left\{\overline{\boldsymbol{M}}_{i}\right\}_{i=1}^{N_{1}}$}
$\{\overline{\mathbb{M}}^{k}\}^{z}_{k=1},\overline{\mathbb{M}}^{o}=\text{DBSCAN}(\left\{\overline{\boldsymbol{M}}_{i}\right\}_{i=1}^{N_{1}}, \epsilon, a_{min})$  \quad \quad \tcp{Generate clusters and discard outliers}
 $\boldsymbol{P}^{k}=\text{Prototype}(\overline{\mathbb{M}}^{k})$\quad \quad \quad \quad \quad \quad \quad \quad \quad \quad \quad \quad \quad \tcp{Generate skeleton prototypes with Eq. \ref{prototype}}
$ \mathcal{L}_{\text     {SPC}}(\{\overline{\mathbb{M}}^{k}\}^{z}_{k=1},\{\boldsymbol{P}^{k}\}^{z}_{k=1},\tau)$ \quad \quad \quad \quad \quad \quad \quad \quad \ \  \tcp{Calculate SPC loss with Eq. \ref{SPC}}
update parameters of $\psi(\cdot)$ to minimize $ \mathcal{L}_{\text {SPC}}$ 
}
\end{algorithm*}

\textbf{Mean Intra-Class Tightness (mACT):} To obatin mACT, we first compute intra-class tightness (ACT) of the $k^{th}$ class with
\begin{equation}
\label{ACT_K}
\text{ACT}(k)=\frac{D_{\text{global-average}}}{D_{k\text{-intra-class}}}=\frac{s_{k} \sum_{i=1}^{C} \frac{1}{s_{i}} \sum_{j=1}^{s_{i}}\sum_{z=1}^{C} D\left(\boldsymbol{v}_{j}^{i}, \boldsymbol{c}_{z}\right)}{C^2 \sum_{j=1}^{s_k} D\left(\boldsymbol{v}_{j}^{k}, \boldsymbol{c}_{k}\right)}
\end{equation}
where $D_{k\text{-intra-class}}$ represents the \textit{average} intra-class distance between the representation, $\boldsymbol{v}^{k}_{j}$, of $j^{th}$ sample and the representation centroid, $\boldsymbol{c}_{k}$, belonging to the $k^{th}$ class, $D_{\text{global-average}}$ denotes the \textit{global average} distance between each representation and all centroids, $s_{i}$ represents the number of instances in the $i^{th}$ class, $C$ is the number of ground-truth classes, and $D(\boldsymbol{v}^{i}_j, \boldsymbol{c}_z)=1-\frac{\boldsymbol{v}^{i}_{j} \cdot \boldsymbol{c}_{z}}{\|\boldsymbol{v}^{i}_{j}\|_{2}\|\boldsymbol{c}_{z}\|_{2}}$ is the cosine distance between $\boldsymbol{v}^{i}_{j}$ and $\boldsymbol{c}_{z}$. Note that $1/{D_{k\text{-intra-class}}}$ can be used to represent the $k$-class tightness, but this result does NOT consider the factor of global feature distribution that leads to different representation distances. Therefore, we use $D_{\text{global-average}}/{D_{k\text{-intra-class}}}$ to indicate the degree of intra-class tightness. When $D_{k\text{-intra-class}}$ is smaller than  $D_{\text{global-average}}$, the representations in $k^{th}$ class are denser than the average, thus indicating a higher ACT of $k^{th}$ class. Then, we average the ACT values of all classes to represent the mean intra-class tightness (mACT) of representations by
$\text{mACT}=\frac{1}{C}\sum^{C}_{k=1} \text{ACT}(k)$.

\textbf{Mean Inter-Class Looseness (mRCL):} To measure the mRCL of representations, we compute the ratio of the average distance between different class centroids, $D_{\text{centroid-average}}$, to the global average distance, $D_{\text{global-average}}$ \hc{(see Eq. \ref{ACT_K})}, with 
\begin{align}
\label{mRCL}
\text{mRCL}=\frac{D_{\text{centroid-average}}}{D_{\text{global-average}}}=\frac{\sum^{C}_{i= 1}\sum^{C}_{j=1}D(\boldsymbol{c}_{i}, \boldsymbol{c}_{j})}{\sum^{C}_{i= 1}\frac{1}{s_i}\sum^{s_i}_{j=1}\sum^{C}_{z= 1}D(\boldsymbol{v}^{i}_{j}, \boldsymbol{c}_{z})}
\end{align}
where higher mRCL can be obtained with greater average distance $D_{\text{centroid-average}}$ between different representation centroids compared to the global average representation distance $D_{\text{global-average}}$. Full details of mACT and mRCL are provided in the Appendix. \hc{We conduct both quantitative and qualitative analysis of the proposed skeleton prototype contrastive learning scheme using the mACT and mRCL metrics, which shows that it can encourage the model to capture class-related semantics for more effective skeleton representations,
as detailed in Sec. \ref{qq_analysis} and Appendix.}

\subsection{The Entire Approach}
The computation flow of the proposed approach can be described as: $\boldsymbol{S}\rightarrow$
$\mathcal{G}$ (Sec. \ref{graph_construct})
$\rightarrow\boldsymbol{F}$ (Sec. \ref{structural-collaborative}) $\rightarrow\boldsymbol{M}$ (Eq. \ref{ske_rep})
$\rightarrow 
\overline{\boldsymbol{M}}$ (Sec. \ref{SC_contrastive_learning}) $\rightarrow \boldsymbol{P}$ (Eq. \ref{prototype}). For convenience, we use the embedding function $\psi(\cdot)$ to represent the multi-level skeleton graph representation encoding process, which can be formulated as $\psi(\boldsymbol{S})=\overline{\boldsymbol{M}}$. We perform skeleton prototype contrastive learning by minimizing $\mathcal{L}_{\text {SPC}}$, so as to optimize $\psi(\cdot)$ and learn effective skeleton representations in an unsupervised manner, as illustrated in Algorithm \ref{algorithm_1}.
To facilitate better skeleton representation learning with more reliable clusters, we optimize our model by alternating clustering and contrastive learning. More technical details are provided in the Appendix. For the person re-ID task, we exploit the learned embedding function $\psi(\cdot)$ to encode each skeleton sequence of the probe set $\Phi_{p}$ into corresponding multi-level graph representation, $\{\overline{\boldsymbol{M}}^{p,i}\}_{i=1}^{N_{2}}$, and match it with representations, $\{\overline{\boldsymbol{M}}^{g,j}\}_{j=1}^{N_{3}}$, of the same identity in the gallery set $\Phi_{g}$ using Euclidean distance.

\begin{table}[t]
\caption{Statistics of different datasets. Note: We construct different gallery and probe sets based on multiple testing splits (see Sec. \ref{exp_setup}). For CASIA-B dataset, we exploit 3D skeleton data estimated from RGB videos. }
\label{dataset_details}
\begin{center}
\scalebox{0.8}{
\setlength{\tabcolsep}{1.0mm}{
\begin{tabular}{l|c|c|c|c|c}
\specialrule{0.1em}{0.45pt}{0.45pt}
                                                                         & \textbf{KGBD} & \textbf{BIWI}                                                         & \textbf{KS20} & \textbf{IAS-Lab}                                                        & \textbf{CASIA-B}                                                               \\ \specialrule{0.1em}{0.45pt}{0.45pt}
\textbf{\# Train IDs}                                                    & 164           & 50                                                                    & 20            & 11                                                                  & 124                                                                            \\ \specialrule{0.1em}{0.45pt}{0.45pt}
\textbf{\begin{tabular}[c]{@{}l@{}}\# Train \\ \ \ \ Skeletons\end{tabular}}  & 188,742       & 205,764                                                               & 35,976        & 88,986                                                              & 706,480                                                                        \\ \specialrule{0.1em}{0.45pt}{0.45pt}
\textbf{\# Gallery IDs}                                                  & 164           & 28                                                                    & 20            & 11                                                                  & 62                                                                             \\ \specialrule{0.1em}{0.45pt}{0.45pt}
\textbf{\begin{tabular}[c]{@{}l@{}}\# Gallery \\ \ \ \ Skeletons\end{tabular}} & 188,700       & \begin{tabular}[c]{@{}c@{}}Walking: 4,932\\ Still: 3,186\end{tabular} & 3,252         & \begin{tabular}[c]{@{}c@{}}IAS-A: 6,978\\ IAS-B: 7,764\end{tabular} & \begin{tabular}[c]{@{}c@{}}Nm: 162,080\\ Cl: 54,400 \\ Bg: 53,880\end{tabular} \\ \specialrule{0.1em}{0.45pt}{0.45pt}
\textbf{\# Probe IDs}                                                    & 164           & 28                                                                    & 20            & 11                                                                  & 62                                                                             \\ \specialrule{0.1em}{0.45pt}{0.45pt}
\textbf{\begin{tabular}[c]{@{}l@{}}\# Probe \\ \ \ \ Skeletons\end{tabular}}   & 94,146        & \begin{tabular}[c]{@{}c@{}}Walking: 4,932\\ Still: 3,186\end{tabular} & 3,306         & \begin{tabular}[c]{@{}c@{}}IAS-A: 6,978\\ IAS-B: 7,764\end{tabular} & \begin{tabular}[c]{@{}c@{}}Nm: 162,080\\ Cl: 54,400 \\ Bg: 53,880\end{tabular} \\ \specialrule{0.1em}{0.2pt}{0.2pt}
\end{tabular}
}
}
\end{center}
\end{table}

\begin{figure}[t]
    \centering
     \quad 
         \scalebox{0.215}{\includegraphics{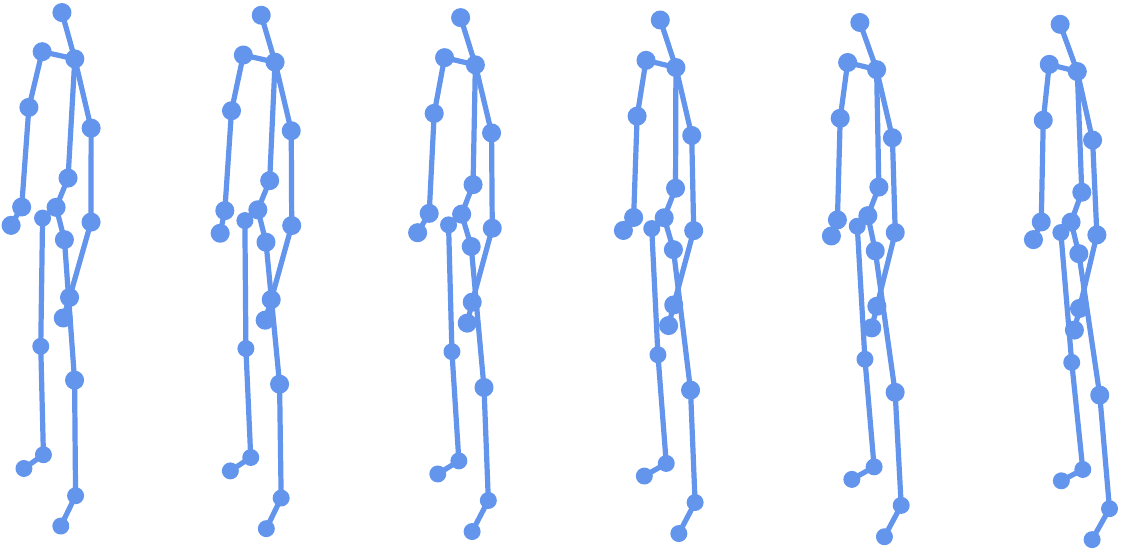}} 
           \quad \quad \quad \ \ \ \scalebox{0.245}{\includegraphics{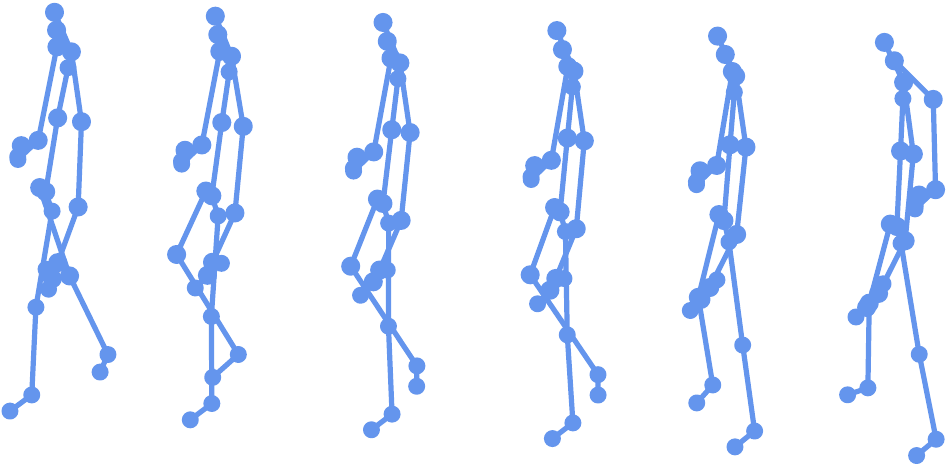}} 
           
      \scalebox{0.075}{\includegraphics{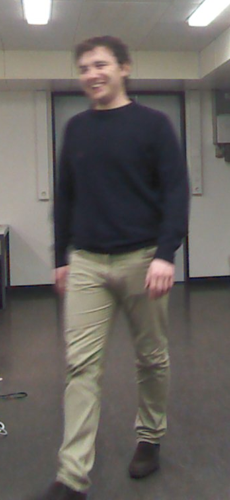}}
       \scalebox{0.075}{\includegraphics{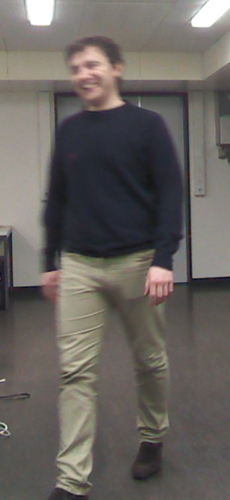}}
        \scalebox{0.075}{\includegraphics{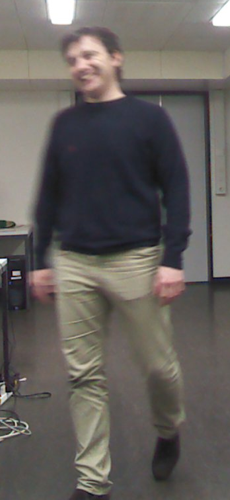}}
         \scalebox{0.075}{\includegraphics{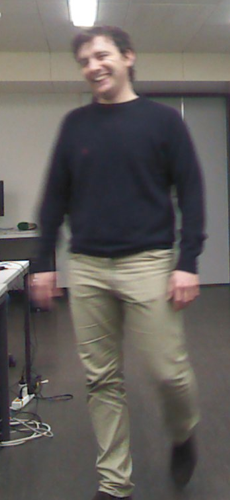}}
          \scalebox{0.075}{\includegraphics{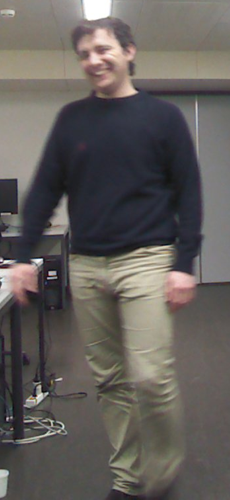}}
           \scalebox{0.075}{\includegraphics{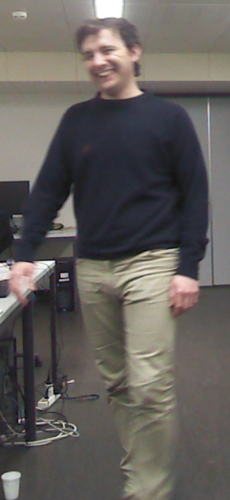}}  \quad \quad \  \ \
            \scalebox{0.205}{\includegraphics{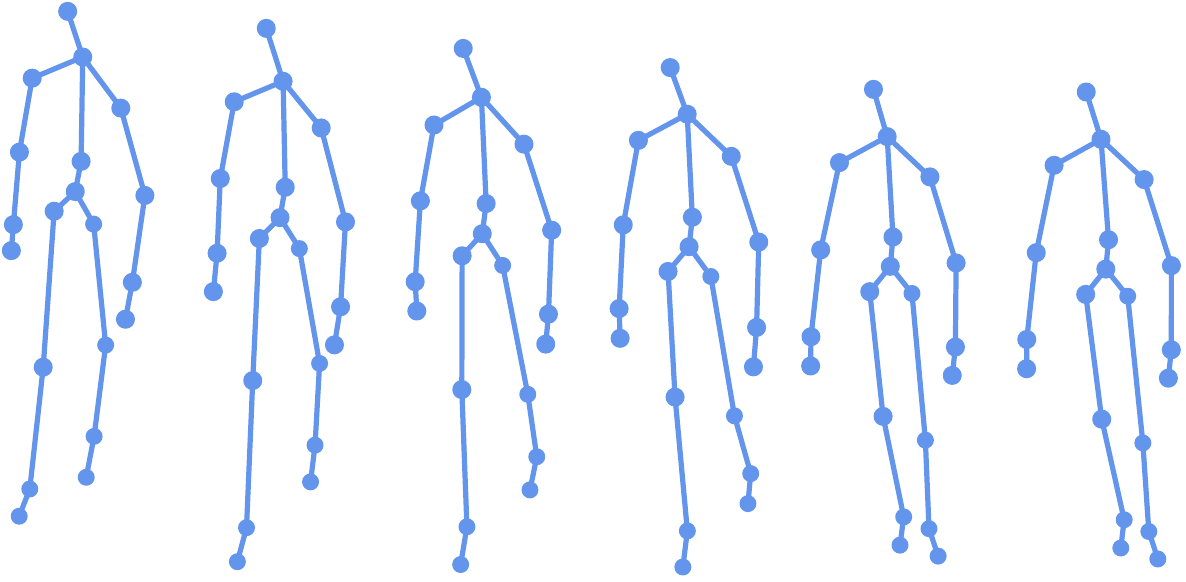}}
            
     \scalebox{0.10}{\includegraphics{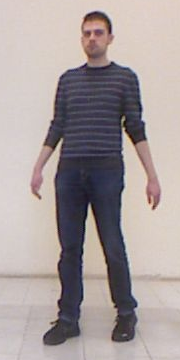}}
       \scalebox{0.10}{\includegraphics{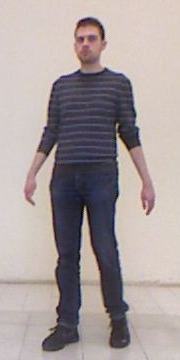}}
        \scalebox{0.10}{\includegraphics{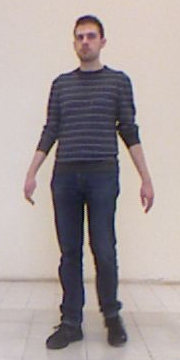}}
         \scalebox{0.10}{\includegraphics{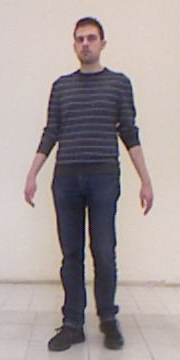}}
          \scalebox{0.10}{\includegraphics{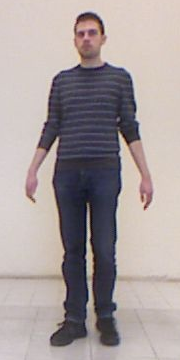}}
           \scalebox{0.10}{\includegraphics{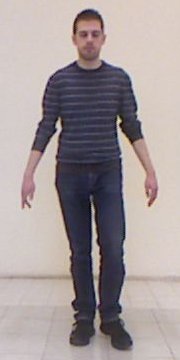}}    \quad \quad  \ \ \scalebox{0.205}{\includegraphics{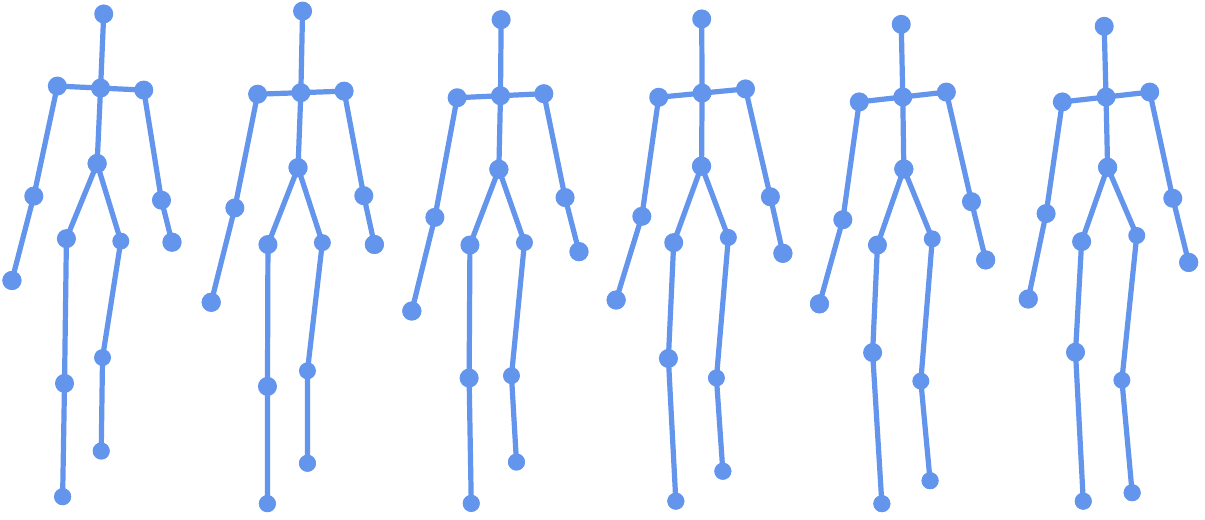}}      
           
 \ \scalebox{0.06}{\includegraphics{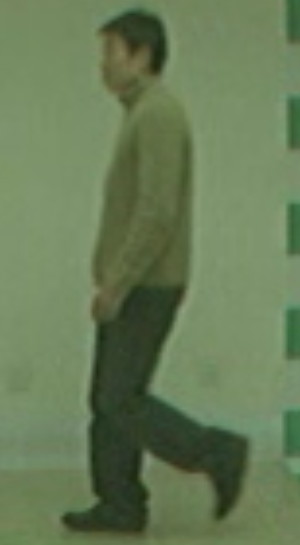}}
       \scalebox{0.06}{\includegraphics{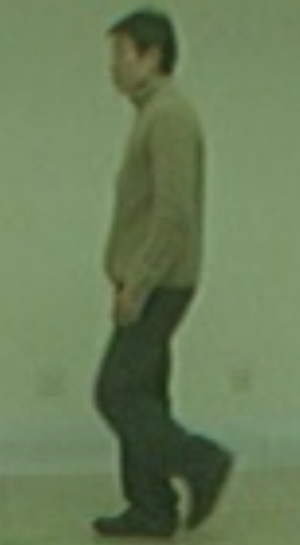}}
        \scalebox{0.06}{\includegraphics{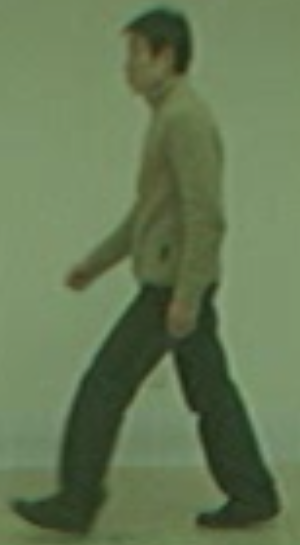}}
         \scalebox{0.06}{\includegraphics{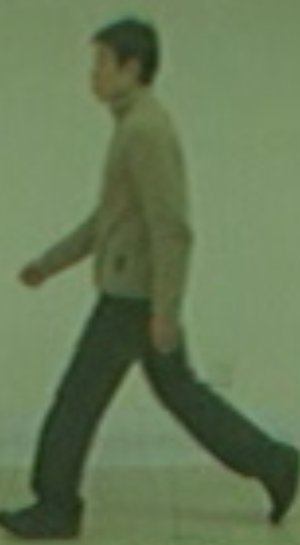}}
          \scalebox{0.06}{\includegraphics{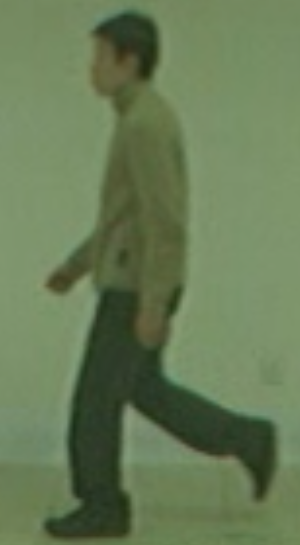}}
           \scalebox{0.06}{\includegraphics{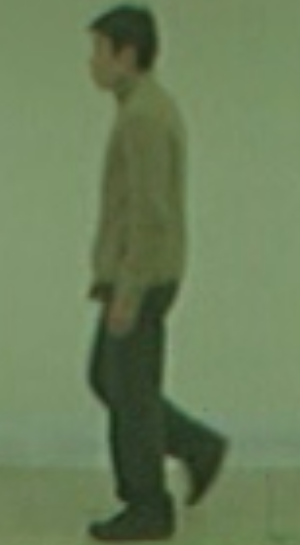}} \quad \quad \quad  \scalebox{0.09}{\includegraphics{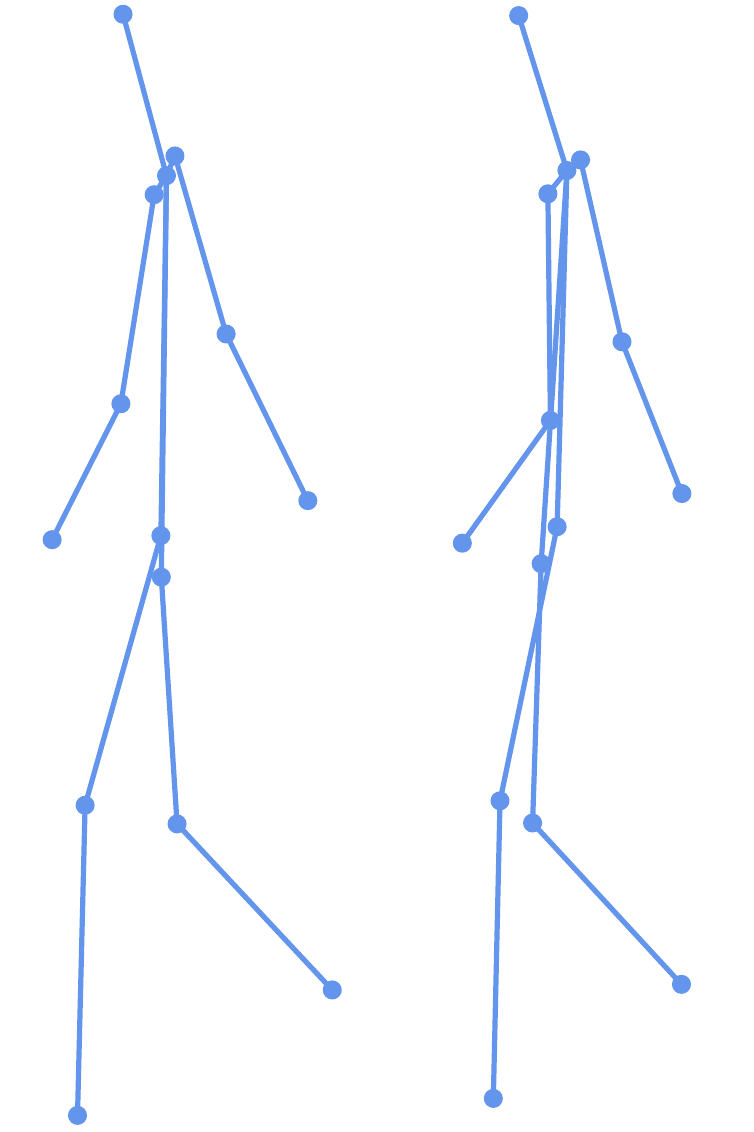}} \scalebox{0.09}{\includegraphics{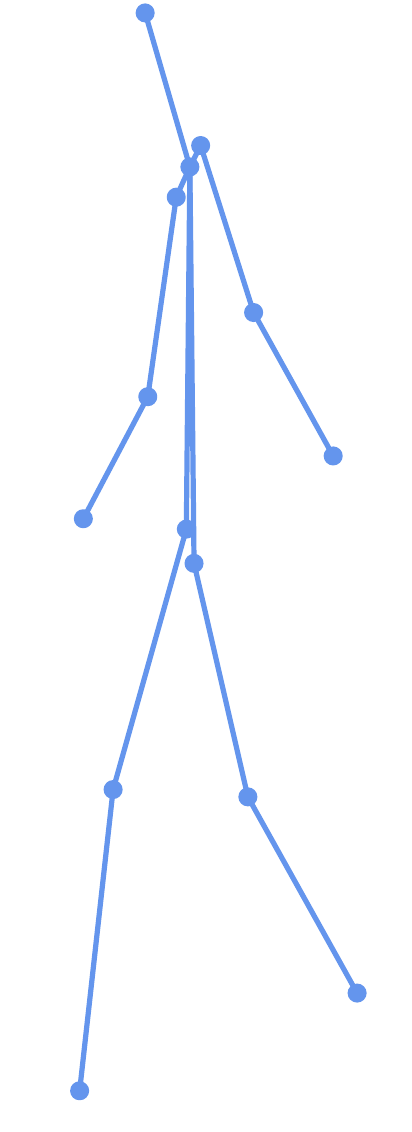}}
           \scalebox{0.09}{\includegraphics{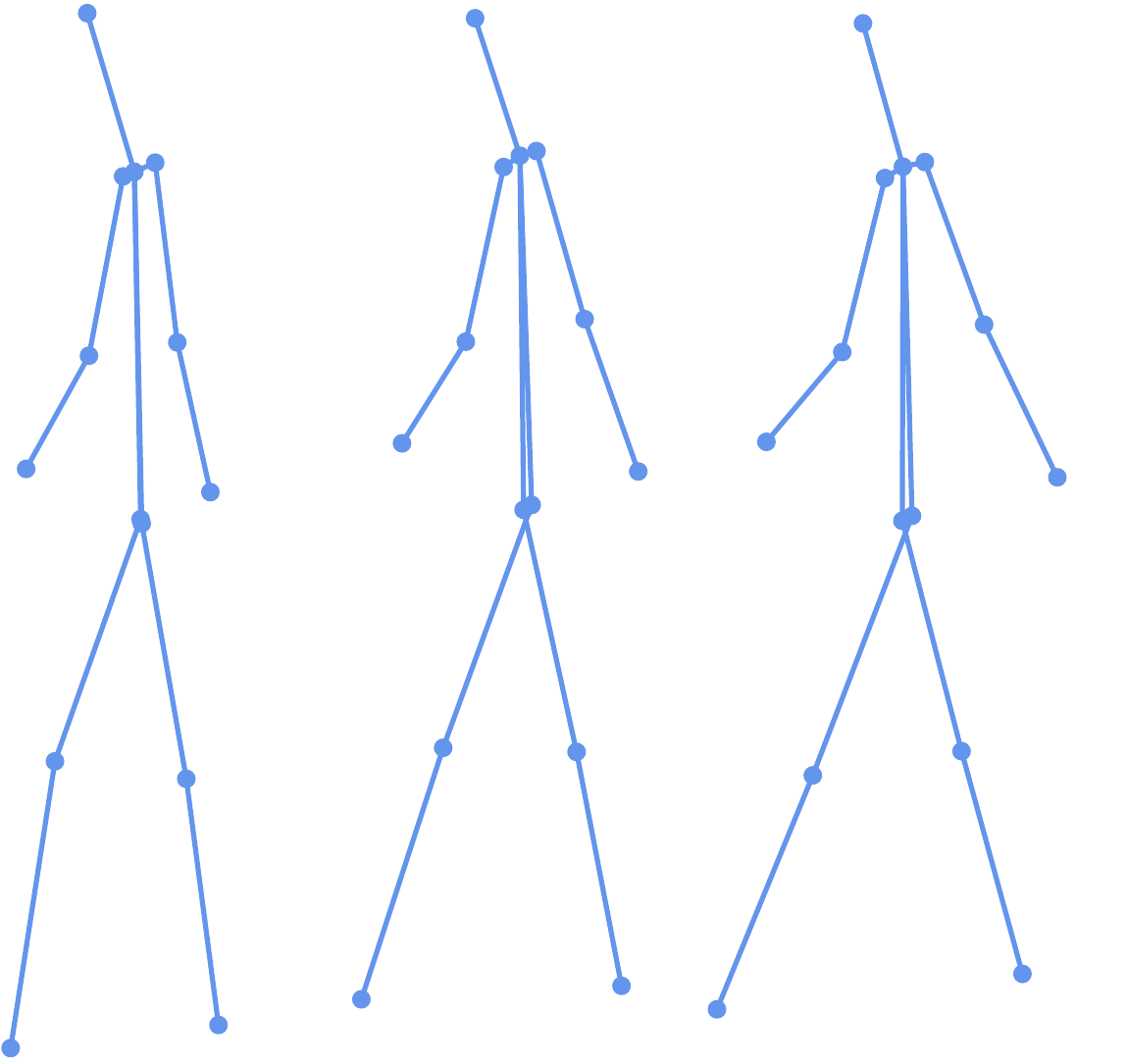}} \ 
    \caption{Examples of 3D skeletons in KGBD (left of first row), KS20 (right of first row), BIWI (second row), IAS-Lab (third row), and CASIA-B (last row). We show RGB samples of datasets that contain visual images. Note: Skeleton sequences of CASIA-B are estimated from RGB videos.  }
    \label{sample_skeleton}
\end{figure}

\begin{table*}[ht]
\centering
\caption{Performance comparison with existing hand-crafted, supervised, self-supervised, and unsupervised methods on KS20 (RVE setup), KGBD, and IAS-A testing sets. \hc{The amount of network parameters (million (M)) and computational complexity (giga floating-point operations (GFLOPs)) for the deep learning based methods are also reported.}
Bold numbers refer to the best \hc{cases} among self-supervised/unsupervised methods. }
\label{KS20_KGBD_IASA}
\scalebox{0.825}{
\setlength{\tabcolsep}{1.5mm}{
\begin{tabular}{llrrrrrrrrrrrrrc}
\specialrule{0.1em}{0.45pt}{0.45pt}
\textbf{}                                                                                         & \textbf{}               & \multicolumn{1}{l}{}                   & \multicolumn{1}{l}{}                & \multicolumn{4}{c}{\textbf{KS20}}                                                                                                                & \multicolumn{4}{c}{\textbf{KGBD}}                                                                                                                & \multicolumn{4}{c}{\textbf{IAS-A}}                                                                                            \\ \specialrule{0.1em}{0.45pt}{0.45pt}
\textbf{Types}                                                                                    & \textbf{Methods}        & \multicolumn{1}{l}{\textbf{\# Params}} & \multicolumn{1}{l}{\textbf{GFLOPs}} & \multicolumn{1}{l}{\textbf{top-1}} & \multicolumn{1}{l}{\textbf{top-5}} & \multicolumn{1}{l}{\textbf{top-10}} & \multicolumn{1}{l}{\textbf{mAP}} & \multicolumn{1}{c}{\textbf{top-1}} & \multicolumn{1}{c}{\textbf{top-5}} & \multicolumn{1}{c}{\textbf{top-10}} & \multicolumn{1}{c}{\textbf{mAP}} & \multicolumn{1}{c}{\textbf{top-1}} & \multicolumn{1}{c}{\textbf{top-5}} & \multicolumn{1}{c}{\textbf{top-10}} & \textbf{mAP}  \\ \specialrule{0.1em}{0.45pt}{0.45pt}
\multirow{2}{*}{\textbf{Hand-crafted}}                                                            & $D_{13}$ Descriptors \cite{munaro2014one}                     & —                                      & —                                   & 39.4                               & 71.7                               & 81.7                                & 18.9                             & 17.0                               & 34.4                               & 44.2                                & 1.9                              & 40.0                               & 58.7                               & 67.6                                & 24.5          \\
                                                                                                  & $D_{16}$ Descriptors \cite{pala2019enhanced}                    & —                                      & —                                   & 51.7                               & 77.1                               & 86.9                                & 24.0                             & 31.2                               & 50.9                               & 59.8                                & 4.0                              & 42.7                               & 62.9                               & 70.7                                & 25.2          \\ \specialrule{0.1em}{0.45pt}{0.45pt}
\multirow{2}{*}{\textbf{Supervised}}                                                              & PoseGait \cite{liao2020model}               & 8.93M                                 & 121.60                               & 49.4                               & 80.9                               & 90.2                                & 23.5                             & 50.6                               & 67.0                               & 72.6                                & 13.9                             & 28.4                               & 55.7                               & 69.2                                & 17.5          \\
                                                                                                  & MG-SCR \cite{rao2021multi}                 & 0.35M                                  & 6.60                                & 46.3                               & 75.4                               & 84.0                                & 10.4                             & 44.0                               & 58.7                               & 64.6                                & 6.9                              & 36.4                               & 59.6                               & 69.5                                & 14.1          \\ \specialrule{0.1em}{0.45pt}{0.45pt}
\multirow{4}{*}{\textbf{\begin{tabular}[c]{@{}l@{}}Self-supervised\\ /Unsupervised\end{tabular}}} & AGE \cite{rao2020self}                    & 7.15M                                  & 37.37                                & 43.2                              & 70.1                              & 80.0                               & 8.9                              & 2.9                               & 5.6                               & 7.5                                & 0.9                              & 31.1                              & 54.8                              & 67.4                               & 13.4         \\
                                                                                                  & SGELA \cite{rao2021self}                  & 8.47M                                  & 7.47                                 & 45.0                              & 65.0                              & 75.1                               & 21.2                            & 38.1                               & 53.5                              & 60.0                               & 4.5                             & 16.7                               & 30.2                              & 44.0                              & 13.2          \\
                                                                                                  & SM-SGE \cite{rao2021sm}                 & 5.58M                                  & 22.61                                & 45.9                               & 71.9                               & 81.1                                & 9.5                              & 38.2                               & 54.2                               & 60.7                                & 4.4                              & 34.0                               & 60.5                               & 71.6                                & 13.6          \\
                                                                                                  & \textbf{SPC-MGR (Ours)} & 0.01M                         & 0.12                        & \textbf{59.0}                      & \textbf{79.0}                      & \textbf{86.2}                       & \textbf{21.7}                    & \textbf{40.8}                      & \textbf{57.5}                      & \textbf{65.0}                       & \textbf{6.9}                     & \textbf{41.9}                      & \textbf{66.3}                      & \textbf{75.6}                       & \textbf{24.2} \\ \specialrule{0.1em}{0.2pt}{0.2pt}
\end{tabular}
}
}
\end{table*}

\section{Experiments}
\label{experiments}
\subsection{Experimental Setup}
\label{exp_setup}
We evaluate our approach on four skeleton-based person re-ID benchmarks: \textit{Kinect Gait Biometry Dataset (KGBD)} \cite{andersson2015person}, \textit{BIWI} \cite{munaro2014one}, \textit{KS20 VisLab Multi-View Kinect Skeleton Dataset} \cite{nambiar2017context}, \textit{IAS-Lab} \cite{munaro2014feature}, and a large-scale RGB video based multi-view gait dataset: \textit{CASIA-B} \cite{yu2006framework}.
They collect skeleton data from 164, 50, 20, 11, and 124 different individuals respectively, as detailed in Table \ref{dataset_details}.
 For BIWI and IAS-Lab datasets, we evaluate each testing set by setting it as the probe and the other one as the gallery. For KGBD, since no training and testing splits are given, we randomly leave one skeleton video of each person as the probe set, and equally divide the remaining videos into the training set and gallery set. For KS20, we design different splitting setups to evaluate the multi-view person re-ID performance of our approach. \hc{In} Random View Evaluation (RVE), we randomly select one sequence from each viewpoint as the probe sequence and use \hc{one half} of the remaining skeleton sequences for training and \hc{the other half} as the gallery. \hc{For} Cross-View Evaluation (CVE), we divide probe samples by each viewpoint, including left lateral at $0^{\circ}$, left diagonal at $30^{\circ}$, frontal at $90^{\circ}$, right diagonal at $130^{\circ}$, and right lateral at $180^{\circ}$, to construct five individual viewpoint-based probe sets. We test each of them by matching identities from other four viewpoints to evaluate cross-view person re-ID performance. \hc{More details about the experimental setups are provided in the Appendix.
 Experiments with each evaluation setup are repeated for multiple times and the average performance is reported.}



\subsection{Implementation Details} 
The numbers of nodes in the part-level, body-level, and hyper-torso-level graphs are $n_{1}=10$, $n_{2}=5$, and $n_{3}=3$ respectively. The sequence length $f$ on four skeleton-based datasets (KS20, KGBD, BIWI, IAS-Lab) is empirically set to $6$, which achieves best performance in average among different settings. For the largest dataset CASIA-B with roughly estimated skeleton data from RGB frames, we set sequence length $f=40$ for training/testing.
The node feature dimension is $D_{h}=8$ and the number of structural relation heads is $m=16$ for KGBD and $m=8$ for other datasets. We use $\lambda^{a,b}_{C}=1$ ($a, b \in \{1,2,3\}$) to averagely fuse multi-level graph features. For the DBSCAN algorithm, we empirically use maximum distance $\epsilon=0.6$ (KGBD, BIWI), $\epsilon=0.8$  (KS20, IAS-Lab), $\epsilon=0.75$ (CASIA-B), and adopt minimum amount of samples $a_{min}=4$ for KGBD and $a_{min}=2$ for other datasets. We set the temperature $\tau$ to $0.06$ (KGBD), $0.075$ (CASIA-B), $0.07$ (BIWI), $0.08$ (KS20, IAS-Lab) for skeleton prototype contrastive learning.
We employ Adam optimizer with learning rate $0.00035$ for all datasets. The batch size is $256$ for KGBD and $128$ for other datasets. 
For all methods compared in our experiments, we select optimal model parameters for training, \hc{and use their pre-defined skeleton descriptors} or pre-trained skeleton representations for person re-ID.
Full implementation details are provided in the Appendix and our codes are available at \href{https://github.com/Kali-Hac/SPC-MGR}{https://github.com/Kali-Hac/SPC-MGR}. 
\begin{table*}[ht]
\centering
\caption{Performance comparison with existing hand-crafted, supervised, self-supervised, and unsupervised methods on IAS-B, BIWI-Still (BIWI-S), and BIWI-Walking (BIWI-W) testing sets. Bold numbers refer to the best \hc{cases} among self-supervised/unsupervised methods.}
\label{IASB_BIWIS_BIWIW}
\scalebox{0.825}{
\setlength{\tabcolsep}{2.6mm}{
\begin{tabular}{llrrrrrrrrrrrr}
\specialrule{0.1em}{0.45pt}{0.45pt}
\textbf{}                                                                                    & \textbf{} & \multicolumn{4}{c}{\textbf{IAS-B}}                                & \multicolumn{4}{c}{\textbf{BIWI-S}}                           & \multicolumn{4}{c}{\textbf{BIWI-W}}                         \\ \specialrule{0.1em}{0.45pt}{0.45pt}
                                                                                 \textbf{Types}                 & \textbf{Methods}        & \textbf{top-1} & \textbf{top-5} & \textbf{top-10} & \textbf{mAP}  & \textbf{top-1} & \textbf{top-5} & \textbf{top-10} & \textbf{mAP}  & \textbf{top-1} & \textbf{top-5} & \textbf{top-10} & \textbf{mAP}  \\ \specialrule{0.1em}{0.45pt}{0.45pt}
\multirow{2}{*}{\textbf{Hand-crafed}}                                                             & $D_{13}$ Descriptors \cite{munaro2014one}              & 43.7           & 68.6           & 76.7            & 23.7          & 28.3           & 53.1           & 65.9            & 13.1          & 14.2           & 20.6           & 23.7            & 17.2          \\
                                                                                                  & $D_{16}$ Descriptors \cite{pala2019enhanced}              & 44.5           & 69.1           & 80.2            & 24.5          & 32.6           & 55.7           & 68.3            & 16.7          & 17.0           & 25.3           & 29.6            & 18.8          \\ \specialrule{0.1em}{0.45pt}{0.45pt}
\multirow{2}{*}{\textbf{Supervised}}                                                              & PoseGait \cite{liao2020model}         & 28.9           & 51.6           & 62.9            & 20.8          & 14.0           & 40.7           & 56.7            & 9.9           & 8.8            & 23.0           & 31.2            & 11.1          \\
                                                                                                  & MG-SCR \cite{rao2021multi}          & 32.4           & 56.5           & 69.4            & 12.9          & 20.1          & 46.9           & 64.1            & 7.6          & 10.8           & 20.3           & 29.4            & 11.9          \\ \specialrule{0.1em}{0.45pt}{0.45pt}
\multirow{4}{*}{\textbf{\begin{tabular}[c]{@{}l@{}}Self-supervised\\ /Unsupervised\end{tabular}}} & AGE \cite{rao2020self}              & 31.1          & 52.3          & 64.2           & 12.8         & 25.1          & 43.1          & 61.6           & 8.9          & 11.7          & 21.4          & 27.3           & 12.6         \\
                                                                                                  & SGELA \cite{rao2021self}            & 22.2          & 40.8          & 50.2           & 14.0         & 25.8          & 51.8          & 64.4           & 15.1         & 11.7          & 14.0          & 14.7           & 19.0          \\
                                                                                                  & SM-SGE \cite{rao2021sm}           & 38.9          & 64.1          & 75.8           & 13.3          & 31.3          & 56.3           & 69.1           & 10.1         & 13.2          & 25.8          & 33.5           & 15.2         \\
                                                                                                  & \textbf{SPC-MGR (Ours)}             & \textbf{43.3}  & \textbf{68.4}  & \textbf{79.4}   & \textbf{24.1} & \textbf{34.1}  & \textbf{57.3}  & \textbf{69.8}   & \textbf{16.0} & \textbf{18.9}  & \textbf{31.5}  & \textbf{40.5}   & \textbf{19.4} \\ \specialrule{0.1em}{0.2pt}{0.2pt}
\end{tabular}
}
}
\end{table*}

\begin{table*}[ht]
\centering
\caption{Performance comparison with state-of-the-art self-supervised and unsupervised methods \hc{with} cross-view evaluation (CVE) setup of KS20 datasets. $0^{\circ}, 30^{\circ}, 90^{\circ}, 130^{\circ}$, and $180^{\circ}$ denote probe or gallery sets \hc{in} different views.}
\label{KS20_CVE}
\scalebox{0.825}{
\setlength{\tabcolsep}{0.62mm}{
\begin{tabular}{clcccccccccccccccccccc}
\specialrule{0.1em}{0.45pt}{0.45pt}
\multicolumn{1}{l}{}               & \textbf{Gallery}       & \multicolumn{4}{c}{$\boldsymbol{0^{\circ}}$}                                    & \multicolumn{4}{c}{$\boldsymbol{30^{\circ}}$}                                   & \multicolumn{4}{c}{$\boldsymbol{90^{\circ}}$}                                   & \multicolumn{4}{c}{$\boldsymbol{150^{\circ}}$}                                  & \multicolumn{4}{c}{$\boldsymbol{180^{\circ}}$}                                  \\ \specialrule{0.1em}{0.45pt}{0.45pt}
\multicolumn{1}{l}{\textbf{Probe}} & \textbf{}              & \textbf{top-1} & \textbf{top-5} & \textbf{top-10} & \textbf{mAP}  & \textbf{top-1} & \textbf{top-5} & \textbf{top-10} & \textbf{mAP}  & \textbf{top-1} & \textbf{top-5} & \textbf{top-10} & \textbf{mAP}  & \textbf{top-1} & \textbf{top-5} & \textbf{top-10} & \textbf{mAP}  & \textbf{top-1} & \textbf{top-5} & \textbf{top-10} & \textbf{mAP}  \\ \specialrule{0.1em}{0.45pt}{0.45pt}
\multirow{4}{*}{$\boldsymbol{0^{\circ}}$}        & AGE \cite{rao2020self}                    & 46.7           & 74.2           & 83.5            & 22.5          & 11.0           & 35.7           & 47.5            & 10.0          & 8.1            & 29.9           & 47.5            & 9.2           & 7.5            & 26.7           & 43.5            & 8.4           & 7.0            & 23.0           & 37.4            & 8.2           \\
                                   & SGELA \cite{rao2021self}                 & 76.2           & 89.6           & 92.8            & 37.1          & 15.1           & 27.3           & 35.1            & 19.9          & 10.1           & 27.5           & 40.9            & 18.2          & 10.7           & 21.5           & 29.3            & 18.0          & 15.4           & 25.8           & 38.0            & 12.6 \\
                                   & SM-SGE \cite{rao2021sm}                & 58.4           & 84.7           & 92.2            & 27.7          & 17.2           & 50.0           & 63.3            & 10.8          & 7.2            & 21.9           & 39.1            & 10.5          & 4.4            & 19.4           & 34.7            & 9.3           & 10.0           & 23.8           & 33.1            & 9.4           \\
                                   & \textbf{SPC-MGR (Ours)} & \textbf{78.9}  & \textbf{94.1}  & \textbf{97.3}   & \textbf{52.9} & \textbf{26.2}  & \textbf{53.1}  & \textbf{71.5}   & \textbf{22.9} & \textbf{39.1}  & \textbf{59.8}  & \textbf{71.5}   & \textbf{31.4} & \textbf{30.5}  & \textbf{57.4}  & \textbf{72.7}   & \textbf{26.6} & \textbf{27.0}  & \textbf{52.0}  & \textbf{66.4}   & \textbf{19.9}          \\ \specialrule{0.1em}{0.45pt}{0.45pt}
\multirow{4}{*}{$\boldsymbol{30^{\circ}}$}       & AGE                    & 10.1           & 42.8           & 57.8            & 8.8           & 52.3           & 82.7           & 91.5            & 25.0          & 15.0           & 35.6           & 58.5            & 8.8           & 10.1           & 24.2           & 41.8            & 8.1           & 7.8            & 24.2           & 34.3            & 8.3           \\
                                   & SGELA                  & 13.1           & 19.6           & 22.6            & 19.4          & 70.9           & 88.2           & 91.8            & 40.5          & 11.8           & 24.5           & 36.3            & 16.5          & 6.9            & 22.6           & 31.7            & 15.4          & 9.2            & 15.4           & 22.9            & 13.9          \\
                                   & SM-SGE                 & 18.1           & 48.4           & 65.0            & 11.5          & 60.2           & 82.0           & 89.8            & 28.2          & 12.5           & 27.2           & 35.3            & 10.7          & 7.5            & 23.4           & 33.8            & 10.6          & 8.8            & 27.2           & 39.1            & 10.5          \\
                                   & \textbf{SPC-MGR (Ours)} & \textbf{39.1}  & \textbf{60.2}  & \textbf{69.1}   & \textbf{26.2} & \textbf{75.4}  & \textbf{95.7}  & \textbf{96.5}   & \textbf{56.7} & \textbf{40.2}  & \textbf{62.5}  & \textbf{72.3}   & \textbf{32.4} & \textbf{28.9}  & \textbf{55.1}  & \textbf{66.0}   & \textbf{24.9} & \textbf{18.4}  & \textbf{48.1}  & \textbf{66.4}   & \textbf{16.1} \\ \specialrule{0.1em}{0.45pt}{0.45pt}
\multirow{4}{*}{$\boldsymbol{90^{\circ}}$}       & AGE                    & 7.5            & 27.3           & 43.2            & 8.7           & 9.0            & 28.5           & 44.1            & 9.3           & 57.4           & 81.4           & 90.7            & 19.2          & 13.8           & 41.1           & 57.1            & 9.0           & 7.8            & 30.0           & 46.0            & 8.3           \\
                                   & SGELA                  & 9.6            & 19.8           & 29.7            & 16.4          & 10.8           & 15.6           & 20.4            & 17.5          & 48.4           & 75.7           & 86.5            & 31.6          & 17.1           & 35.7           & 43.0            & 22.0          & 13.5           & 23.4           & 31.8            & 21.3          \\
                                   & SM-SGE                 & 19.1           & 33.1           & 48.1            & 12.4          & 23.1           & 40.6           & 57.4            & 11.5          & 72.2           & 89.1           & 92.8            & 24.9          & 20.9           & 48.4           & 69.4            & 12.8          & 19.4           & 36.9           & 51.6            & 11.3          \\
                                   & \textbf{SPC-MGR (Ours)} & \textbf{37.5}  & \textbf{67.2}  & \textbf{75.0}   & \textbf{26.0} & \textbf{41.8}  & \textbf{65.2}  & \textbf{74.2}   & \textbf{32.2} & \textbf{86.7}  & \textbf{98.1}  & \textbf{99.2}   & \textbf{63.1} & \textbf{59.0}  & \textbf{82.4}  & \textbf{86.3}   & \textbf{40.7} & \textbf{34.8}  & \textbf{62.1}  & \textbf{77.0}   & \textbf{24.8} \\ \specialrule{0.1em}{0.45pt}{0.45pt}
\multirow{4}{*}{$\boldsymbol{150^{\circ}}$}      & AGE                    & 6.7            & 21.3           & 34.7            & 8.2           & 7.9            & 23.4           & 38.9            & 8.9           & 15.2           & 35.9           & 54.4            & 9.2           & 45.3           & 70.5           & 82.1            & 18.7          & 11.3           & 37.1           & 50.2            & 8.9           \\
                                   & SGELA                  & 5.8            & 18.8           & 28.0            & 14.2          & 11.6           & 15.5           & 20.7            & 16.8          & 17.6           & 47.1           & 53.2            & 24.5          & 59.6           & 81.5           & 89.1            & 36.8          & 17.0           & 29.8           & 32.5            & \textbf{23.0} \\
                                   & SM-SGE                 & 8.4            & 24.4           & 37.8            & 10.4          & 12.9           & 26.6           & 36.3            & 10.9          & 24.1           & 53.4           & 66.3            & 12.9          & 64.4           & 85.9           & 95.0            & 25.5          & 17.8           & 40.9           & 59.1            & 12.1          \\
                                   & \textbf{SPC-MGR (Ours)} & \textbf{28.5}  & \textbf{59.8}  & \textbf{71.9}   & \textbf{23.3} & \textbf{25.4}  & \textbf{49.6}  & \textbf{65.2}   & \textbf{22.3} & \textbf{57.4}  & \textbf{77.0}  & \textbf{85.2}   & \textbf{40.8} & \textbf{77.3}  & \textbf{96.5}  & \textbf{97.7}   & \textbf{58.6} & \textbf{35.9}  & \textbf{62.5}  & \textbf{79.3}   & \textbf{23.0} \\ \specialrule{0.1em}{0.45pt}{0.45pt}
\multirow{4}{*}{$\boldsymbol{180^{\circ}}$}      & AGE                    & 7.9            & 17.7           & 32.6            & 8.1           & 5.2            & 22.4           & 33.4            & 8.3           & 10.5           & 25.6           & 34.0            & 8.2           & 11.6           & 33.1           & 52.9            & 8.8           & 47.1           & 72.4           & 82.6            & 22.6          \\
                                   & SGELA                  & 14.0           & 29.1           & 39.2            & 21.3          & 11.9           & 20.6           & 25.9            & 17.3          & 18.6           & 37.8           & 49.7            & 19.4          & 22.7           & 45.9           & 55.2            & 20.7          & \textbf{74.5}           & \textbf{92.7}  & 95.1            & 38.3          \\
                                   & SM-SGE                 & 5.6            & 20.0           & 30.6            & 8.5           & 6.6            & 22.7           & 31.6            & 8.6           & 13.8           & 34.1           & 45.6            & 9.4           & 10.3           & 37.5           & 56.6            & 10.4          & 51.9           & 79.7           & 87.8            & 25.6          \\
                                   & \textbf{SPC-MGR (Ours)} & \textbf{28.5}  & \textbf{53.5}  & \textbf{62.5}   & \textbf{21.7} & \textbf{17.2}  & \textbf{37.1}  & \textbf{50.0}   & \textbf{18.8} & \textbf{31.6}  & \textbf{53.5}  & \textbf{66.8}   & \textbf{30.0} & \textbf{31.3}  & \textbf{56.3}  & \textbf{77.3}   & \textbf{26.4} & 65.2  & 89.5           & \textbf{96.5}   & \textbf{45.9} \\ \specialrule{0.1em}{0.2pt}{0.2pt}
\end{tabular}
}
}
\end{table*}

\subsection{Evaluation Metrics} 
We compute Cumulative Matching Characteristics (CMC) curve, which typically adopts top-1, top-5, and top-10 accuracy as the quantitative metrics, to show ratios that a probe sample matches the same identity in different-sized ranked lists of gallery samples. We also report Mean Average Precision (mAP) \cite{zheng2015scalable} to evaluate the overall performance of our approach. 
Unlike existing skeleton-based re-ID works that take only top-1 accuracy and nAUC measurements \cite{gray2008viewpoint} under the assumption of fixed classes, this work evaluates all existing methods with more comprehensive metrics (top-1, top-5, top-10 accuracy and mAP) under the frequently-used person re-ID evaluation protocol, $i.e.,$ match IDs between probe and gallery, which can be extended to more general scenarios with varying classes such as generalized person re-ID tasks.

\subsection{Performance Comparison}
\label{performance_comparison}
 In this section, we compare our approach with state-of-the-art self-supervised and unsupervised skeleton-based person re-ID methods \cite{rao2020self,rao2021self,rao2021sm} on KS20, KGBD, IAS-Lab, and BIWI datasets \hc{with} different probe settings in Table \ref{KS20_KGBD_IASA} and Table \ref{IASB_BIWIS_BIWIW}.
As a reference for the overall performance, we include the latest supervised skeleton-based person re-ID methods \cite{liao2020model,rao2021multi} and representative hand-crafted person re-ID methods \cite{munaro20143d, pala2019enhanced}. \hc{For deep learning based methods, we also report their model sizes, $i.e.$, amount of network parameters, and computational complexity in Table \ref{KS20_KGBD_IASA}. }


\subsubsection{Comparison with Self-Supervised and Unsupervised Methods}
As presented in Table \ref{KS20_KGBD_IASA} and Table \ref{IASB_BIWIS_BIWIW}, the proposed SPC-MGR enjoys distinct advantages over existing self-supervised and unsupervised methods on all datasets. Compared with AGE model \cite{rao2020self} that learns skeleton features based on body-joint sequence representations, our approach consistently achieves higher person re-ID performance by a large margin of \hc{$7.2$ to $37.9\%$} top-1 accuracy and \hc{$6.0$ to $12.8\%$} mAP on different datasets, which demonstrates that the proposed multi-level skeleton graph representations with structural-collaborative body relation learning are more effective on modeling discriminative skeleton features for the person re-ID task. Our approach significantly outperforms the state-of-the-art skeleton contrastive learning method SGELA \cite{rao2021self} by up to $25.2\%$ top-1 accuracy and $11.0\%$ mAP on KS20, KGBD, IAS-A, and IAS-B testing sets. On two testing sets of BIWI, both SGELA and the proposed approach obtain comparable mAP, while our SPC-MGR can  achieve superior overall performance with higher top-1 ($7.2$-$8.3\%$), top-5 ($5.5$-$17.5\%$), and top-10 accuracy ($5.4$-$25.8\%$). \hc{Finally,} our approach also performs better than the latest graph-based skeleton representation learning method SM-SGE \cite{rao2021sm} 
by a distinct margin of $2.6$-$13.1\%$ top-1 accuracy and $2.5$-$12.2\%$ mAP on all datasets. In contrast to direct inter-sequence contrastive learning \cite{rao2021self} or manually devising pretext tasks for skeleton representation learning \cite{rao2020self,rao2021sm}, our approach can automatically mine most representative skeleton features by contrasting sequence-level representations (instances) and cluster-level representations (prototypes), which enables our model to learn better skeleton representations for person re-ID.
\hc{Moreover, our model requires only 0.01M parameters and evidently lower computational complexity for skeleton representation learning compared with existing self-supervised and unsupervised methods, as shown in Table \ref{KS20_KGBD_IASA}, which demonstrates its superior efficiency for person re-ID tasks.}

\begin{table*}[t]
\caption{Ablation study of our model with different components: Multi-level skeleton graphs (MG), multi-head structural relation layer (MSRL), full-level collaborative relation layer (FCRL), and skeleton prototype contrastive learning (SPC). ``SG'' denotes employing the single-level graph (part-level graph) and ``+'' indicates using the corresponding model component. ``SG + MSRL'' is evaluated under random model initialization without SPC.}
\label{ablation}
\scalebox{0.825}{
\setlength{\tabcolsep}{3.0mm}{
\begin{tabular}{llcccccccccccc}
\specialrule{0.1em}{0.45pt}{0.45pt}
\multirow{2}{*}{\textbf{Id}} & \multirow{2}{*}{\textbf{Configurations}} & \multicolumn{2}{c}{\textbf{KS20}} & \multicolumn{2}{c}{\textbf{KGBD}} & \multicolumn{2}{c}{\textbf{IAS-A}} & \multicolumn{2}{c}{\textbf{IAS-B}} & \multicolumn{2}{c}{\textbf{BIWI-W}} & \multicolumn{2}{c}{\textbf{BIWI-S}} \\
                             &                                          & \textbf{top-1}   & \textbf{mAP}   & \textbf{top-1}   & \textbf{mAP}   & \textbf{top-1}    & \textbf{mAP}   & \textbf{top-1}    & \textbf{mAP}   & \textbf{top-1}    & \textbf{mAP}    & \textbf{top-1}    & \textbf{mAP}    \\ \specialrule{0.1em}{0.45pt}{0.45pt}
1                            & Baseline                                 & 17.0             & 9.5            & 20.5             & 4.4            & 29.4              & 13.8           & 30.2              & 13.3           & 10.9              & 14.1            & 24.8              & 9.3             \\
2                            & SG + MSRL                                & 18.6             & 10.2           & 21.4             & 3.7            & 30.3              & 14.3           & 31.8              & 13.3           & 11.2              & 13.8            & 26.0              & 11.0            \\
3                            & SG + MSRL + SPC                          & 28.4             & 15.5           & 26.2             & 5.7            & 37.9              & 21.5           & 38.5              & 20.8           & 15.4              & 16.4            & 27.3              & 12.2            \\
4                            & MG + MSRL + SPC                          & 45.1             & 21.2           & 34.5             & 6.3            & 40.0              & 22.5           & 41.9              & 23.2           & 18.1              & 16.9            & 31.5              & 13.4            \\
5                            & MG + MSRL + FCRL + SPC                   & 59.0             & 21.7           & 40.8            & 6.9            & 41.9              & 24.2           & 43.3              & 24.1           & 18.9              & 19.4            & 34.1              & 16.0            \\ \specialrule{0.1em}{0.2pt}{0.2pt}
\end{tabular}
}
}
\end{table*}

\begin{table*}[t]
\centering
\caption{Performance of our approach on different datasets when exploiting different levels of graphs. ``$\checkmark$'' indicates using corresponding graph level. }
\label{graph_level}
\scalebox{0.825}{
\setlength{\tabcolsep}{2.45mm}{
\begin{tabular}{ccccccccccccccc}
\specialrule{0.1em}{0.45pt}{0.45pt}
\multirow{2}{*}{\textbf{Hyper-body-level}} & \multirow{2}{*}{\textbf{Body-level}} & \multirow{2}{*}{\textbf{Part-level}} & \multicolumn{2}{c}{\textbf{KS20}}                                     & \multicolumn{2}{c}{\textbf{KGBD}}                                     & \multicolumn{2}{c}{\textbf{IAS-A}}                                    & \multicolumn{2}{c}{\textbf{IAS-B}}                                    & \multicolumn{2}{c}{\textbf{BIWI-W}}                                   & \multicolumn{2}{c}{\textbf{BIWI-S}}                                   \\
                                           &                                      &                                      & \multicolumn{1}{l}{\textbf{top-1}} & \multicolumn{1}{l}{\textbf{mAP}} & \multicolumn{1}{l}{\textbf{top-1}} & \multicolumn{1}{l}{\textbf{mAP}} & \multicolumn{1}{l}{\textbf{top-1}} & \multicolumn{1}{l}{\textbf{mAP}} & \multicolumn{1}{l}{\textbf{top-1}} & \multicolumn{1}{l}{\textbf{mAP}} & \multicolumn{1}{l}{\textbf{top-1}} & \multicolumn{1}{l}{\textbf{mAP}} & \multicolumn{1}{l}{\textbf{top-1}} & \multicolumn{1}{l}{\textbf{mAP}} \\ \specialrule{0.1em}{0.45pt}{0.45pt}
\checkmark                                         &                                      &                                      & 29.9                               & 15.6                             & 15.2                               & 4.2                              & 30.6                               & 20.4                             & 30.9                               & 21.1                             & 14.5                               & 3.7                              & 20.9                               & 9.4                              \\
                                           & \checkmark                                     &                                      & 44.3                               & 19.9                             & 20.8                               & 4.0                              & 34.7                               & 20.9                             & 38.5                               & 22.9                             & 16.5                               & 15.4                             & 27.9                               & 12.4                             \\
                                           &                                      & \checkmark                                     & 48.2                               & 20.9                             & 34.2                               & 5.4                              & 41.5                               & 21.7                             & 39.1                               & 20.3                             & 17.8                               & 16.9                             & 29.9                               & 13.7                             \\
\checkmark                                           & \checkmark                                     &                                      & 46.1                               & 20.5                             & 26.7                               & 5.7                              & 39.3                               & 21.3                             & 37.8                               & 22.0                             & 16.6                               & 16.9                             & 33.4                               & 14.1                             \\
\checkmark                                           &                                      & \checkmark                                     & 51.4                               & 21.1                             & 34.6                               & 6.0                              & 41.3                               & 22.8                             & 38.9                               & 23.2                             & 18.3                               & 16.5                             & 34.0                               & 13.5                             \\
                                           & \checkmark                                     & \checkmark                                     & 53.3                               & 21.3                             & 35.6                               & 6.3                              & 41.7                               & 21.9                             & 41.5                               & 23.1                             & 18.9                               & 18.5                             & 32.5                               & 15.1                             \\
\checkmark                                           & \checkmark                                     & \checkmark                                     & 59.0                               & 21.7                             & 40.8                               & 6.9                              & 41.9                               & 24.2                             & 43.3                               & 24.1                             & 19.3                               & 18.9                             & 34.1                               & 15.3                             \\ \specialrule{0.1em}{0.2pt}{0.2pt}
\end{tabular}
}
}
\end{table*}

\begin{table*}[t]
\centering
\caption{Performance of our approach on different datasets when employing different amounts of structural relation heads ($m=2, 4, 8, 16$). }
\label{m}
\scalebox{0.825}{
\setlength{\tabcolsep}{4.8mm}{
\begin{tabular}{ccccccccccccc}
\specialrule{0.1em}{0.45pt}{0.45pt}
\multirow{2}{*}{$m$} & \multicolumn{2}{c}{\textbf{KS20}} & \multicolumn{2}{c}{\textbf{KGBD}} & \multicolumn{2}{c}{\textbf{IAS-A}} & \multicolumn{2}{c}{\textbf{IAS-B}} & \multicolumn{2}{c}{\textbf{BIWI-W}} & \multicolumn{2}{c}{\textbf{BIWI-S}} \\
                            & \textbf{top-1}   & \textbf{mAP}   & \textbf{top-1}   & \textbf{mAP}   & \textbf{top-1}    & \textbf{mAP}   & \textbf{top-1}    & \textbf{mAP}   & \textbf{top-1}    & \textbf{mAP}    & \textbf{top-1}    & \textbf{mAP}    \\ \specialrule{0.1em}{0.45pt}{0.45pt}
2                           & 52.7             & 20.1           & 30.3             & 6.2            & 38.9              & 22.2           & 40.7              & 23.1           & 17.3              & 17.5            & 25.8              & 11.5            \\
4                           & 53.9             & 21.4           & 33.7             & 6.5            & 39.3              & 19.4           & 40.8              & 23.4           & 17.8              & 17.8            & 30.5              & 13.3            \\
8                           & 59.0             & 21.7           & 38.9             & 6.6            & 41.9              & 24.2           & 43.3              & 24.1           & 18.9              & 19.4            & 34.1              & 16.0            \\
16                          & 56.5             & 21.6           & 40.8             & 6.9            & 41.0              & 23.4           & 42.1              & 23.8           & 18.8              & 19.3            & 33.2              & 14.4            \\ \specialrule{0.1em}{0.2pt}{0.2pt}
\end{tabular}
}
}
\end{table*}

\begin{table*}[t]
\centering
\caption{Performance of our approach on different datasets when setting different coefficients ($\lambda_{C}=0.1, 0.25, 0.5, 1.0$) for multi-level graph feature fusion. }
\label{lambda}
\scalebox{0.825}{
\setlength{\tabcolsep}{4.8mm}{
\begin{tabular}{ccccccccccccc}
\specialrule{0.1em}{0.45pt}{0.45pt}
\multirow{2}{*}{$\lambda_{C}$} & \multicolumn{2}{c}{\textbf{KS20}} & \multicolumn{2}{c}{\textbf{KGBD}} & \multicolumn{2}{c}{\textbf{IAS-A}} & \multicolumn{2}{c}{\textbf{IAS-B}} & \multicolumn{2}{c}{\textbf{BIWI-W}} & \multicolumn{2}{c}{\textbf{BIWI-S}} \\
                                 & \textbf{top-1}   & \textbf{mAP}   & \textbf{top-1}   & \textbf{mAP}   & \textbf{top-1}    & \textbf{mAP}   & \textbf{top-1}    & \textbf{mAP}   & \textbf{top-1}    & \textbf{mAP}    & \textbf{top-1}    & \textbf{mAP}    \\ \specialrule{0.1em}{0.45pt}{0.45pt}
0.1                              & 53.3             & 20.4           & 34.1             & 6.3            & 40.8              & 22.3           & 40.6              & 21.7           & 18.4              & 18.2            & 32.8              & 13.5            \\
0.25                             & 54.2             & 20.2           & 34.0             & 5.7            & 40.9              & 22.2           & 43.1              & 23.9           & 18.4              & 17.1            & 33.3              & 13.5            \\
0.5                              & 55.5             & 21.2           & 35.3             & 6.3            & 41.4              & 22.8           & 43.6              & 24.0           & 18.5              & 18.5            & 33.4              & 13.4            \\
1.0                              & 59.0             & 21.7           & 40.8             & 6.9            & 41.9              & 24.2           & 43.3              & 24.1           & 18.9              & 19.4            & 34.1              & 16.0            \\ \specialrule{0.1em}{0.2pt}{0.2pt}
\end{tabular}
}
}
\end{table*}

We also compare the performance of our approach with state-of-the-art skeleton-based counterparts \hc{with} the cross-view evaluation (CVE) setup of KS20. 
As shown in Table \ref{KS20_CVE}, our approach remarkably outperforms the latest self-supervised and multi-view skeleton-based methods \hc{SGELA \cite{rao2021self} and SM-SGE \cite{rao2021sm}} by an average margin of $6.5$-$32.1\%$ top-1 accuracy, $12.8$-$41.0\%$ top-5 accuracy, $17.5$-$40.0\%$ top-10 accuracy, and $5.2$-$22.8\%$ mAP on 24 out of 25 testing combinations of \hc{probe views and gallery views,} which demonstrates that our model can learn more discriminative skeleton representations with better robustness against viewpoint variations for cross-view person re-ID.



\subsubsection{Comparison with Hand-Crafted and Supervised Methods} 
Compared with $D_{13}$ \cite{munaro2014one} and $D_{16}$ \cite{pala2019enhanced} that extract hand-crafted geometric and anthropometric skeleton descriptors, our model achieves a significant improvement of person re-ID performance by $1.5$-$23.8\%$ top-1 accuracy on KGBD, BIWI-S, and BIWI-W. Despite gaining similar performance on IAS-A and IAS-B, these methods are inferior to our approach by at least $7.3\%$ top-1 accuracy \hc{on more challenging datasets such as KS20  and KGBD that contains more viewpoints and individuals.} 
\hc{Furthermore,} with \textit{unlabeled} 3D skeletons as the only input, the proposed approach can obtain comparable or even superior performance to two state-of-the-art supervised methods PoseGait \cite{liao2020model} and MG-SCR \cite{rao2021multi} \hc{on five out of six testing sets (KS20, IAS-A, IAS-B, BIWI-S, BIWI-W).} 
Interestingly, with massive labels as the supervision, these methods still fail to obtain satisfactory person re-ID accuracy and even perform worse than hand-crafted methods on datasets with frequent view, shape, and appearance changes \hc{in KS20, IAS, and BIWI with the probe-gallery person re-ID setup}. Considering that our approach does not require any manual annotation and can achieve highly competitive and more balanced performance \hc{with a significantly smaller size of network parameters, as shown in Table \ref{KS20_KGBD_IASA},} it can be a more general solution to skeleton-based person re-ID and related tasks. We will show broader applications of our approach to more general person re-ID scenarios in Sec. \ref{broader_applications}.


\begin{table*}[t]
\centering
\caption{Performance of our approach on different datasets when setting different temperatures ($\tau=0.06, 0.07, 0.08, 0.1, 0.25, 0.5$) for the SPC scheme. }
\label{t}
\scalebox{0.825}{
\setlength{\tabcolsep}{4.7mm}{
\begin{tabular}{ccccccccccccc}
\specialrule{0.1em}{0.45pt}{0.45pt}
\multirow{2}{*}{$\tau$} & \multicolumn{2}{c}{\textbf{KS20}} & \multicolumn{2}{c}{\textbf{KGBD}} & \multicolumn{2}{c}{\textbf{IAS-A}} & \multicolumn{2}{c}{\textbf{IAS-B}} & \multicolumn{2}{c}{\textbf{BIWI-W}} & \multicolumn{2}{c}{\textbf{BIWI-S}} \\
                                     & \textbf{top-1}   & \textbf{mAP}   & \textbf{top-1}   & \textbf{mAP}   & \textbf{top-1}    & \textbf{mAP}   & \textbf{top-1}    & \textbf{mAP}   & \textbf{top-1}    & \textbf{mAP}    & \textbf{top-1}    & \textbf{mAP}    \\ \specialrule{0.1em}{0.45pt}{0.45pt}
0.06                                 & 58.3             & 21.2           & 40.8             & 6.9            & 41.2              & 22.5           & 43.2              & 24.0           & 18.5              & 18.7            & 34.5              & 15.6            \\
0.07                                 & 58.6             & 21.3           & 38.6             & 6.6            & 41.5              & 22.7           & 43.0              & 23.8           & 18.9              & 19.4            & 34.1              & 16.0            \\
0.08                                 & 59.0             & 21.7           & 36.6             & 6.6            & 41.9              & 24.2           & 43.3              & 24.1           & 18.8              & 18.9            & 33.9              & 14.5            \\
0.1                                  & 57.7             & 21.6           & 34.0             & 6.5            & 41.4              & 23.1           & 43.8              & 24.2           & 18.6              & 17.6            & 32.8              & 12.9            \\
0.25                                 & 57.0             & 21.5           & 27.7             & 5.7            & 41.0              & 21.4           & 43.2              & 23.5           & 18.3              & 16.2            & 33.8              & 15.1            \\
0.5                                  & 56.5             & 21.1           & 19.8             & 5.0            & 40.8              & 24.0           & 41.0              & 24.5           & 19.0              & 18.5            & 33.0              & 15.0            \\ \specialrule{0.1em}{0.2pt}{0.2pt}
\end{tabular}
}
}
\end{table*}

\begin{table*}[t]
\centering
\caption{Performance of our approach on different datasets when setting different minimum sample amount ($a_{min}=1, 2, 3, 4$) for the DBSCAN algorithm. }
\label{a_min}
\scalebox{0.825}{
\setlength{\tabcolsep}{4.8mm}{
\begin{tabular}{ccccccccccccc}
\specialrule{0.1em}{0.45pt}{0.45pt}
\multirow{2}{*}{$a_{min}$} & \multicolumn{2}{c}{\textbf{KS20}} & \multicolumn{2}{c}{\textbf{KGBD}} & \multicolumn{2}{c}{\textbf{IAS-A}} & \multicolumn{2}{c}{\textbf{IAS-B}} & \multicolumn{2}{c}{\textbf{BIWI-W}} & \multicolumn{2}{c}{\textbf{BIWI-S}} \\
                                 & \textbf{top-1}   & \textbf{mAP}   & \textbf{top-1}   & \textbf{mAP}   & \textbf{top-1}    & \textbf{mAP}   & \textbf{top-1}    & \textbf{mAP}   & \textbf{top-1}    & \textbf{mAP}    & \textbf{top-1}    & \textbf{mAP}    \\ \specialrule{0.1em}{0.45pt}{0.45pt}
1                                & 53.9             & 21.5           & 30.3             & 6.3            & 41.2             & 23.1           & 43.0              & 23.6           & 18.4              & 18.5            & 34.0              & 14.3            \\
2                                & 59.0             & 21.7           & 34.4             & 6.5            & 41.9             & 24.2           & 43.3              & 24.1           & 18.9              & 19.4            & 34.1              & 16.0            \\
3                                & 58.4             & 21.2           & 36.6             & 6.5            & 41.1              & 22.6           & 43.2              & 23.9           & 19.1              & 19.3            & 32.0              & 13.8            \\
4                                & 58.6             & 21.5           & 40.8             & 6.9            & 40.0              & 22.1           & 42.8              & 24.2           & 18.8              & 18.1            & 30.9              & 13.3            \\ \specialrule{0.1em}{0.2pt}{0.2pt}
\end{tabular}
}
}
\end{table*}

\begin{table*}[t]
\centering
\caption{Performance of our approach on different datasets when setting different maximum distances ($\epsilon=0.4, 0.6, 0.8, 1.0$) for the DBSCAN algorithm. }
\label{eps}
\scalebox{0.825}{
\setlength{\tabcolsep}{4.8mm}{
\begin{tabular}{ccccccccccccc}
\specialrule{0.1em}{0.45pt}{0.45pt}
\multirow{2}{*}{$\epsilon$} & \multicolumn{2}{c}{\textbf{KS20}}                                     & \multicolumn{2}{c}{\textbf{KGBD}}                                     & \multicolumn{2}{c}{\textbf{IAS-A}}                                    & \multicolumn{2}{c}{\textbf{IAS-B}}                                    & \multicolumn{2}{c}{\textbf{BIWI-W}}                                   & \multicolumn{2}{c}{\textbf{BIWI-S}}                                   \\
                              & \multicolumn{1}{l}{\textbf{top-1}} & \multicolumn{1}{l}{\textbf{mAP}} & \multicolumn{1}{l}{\textbf{top-1}} & \multicolumn{1}{l}{\textbf{mAP}} & \multicolumn{1}{l}{\textbf{top-1}} & \multicolumn{1}{l}{\textbf{mAP}} & \multicolumn{1}{l}{\textbf{top-1}} & \multicolumn{1}{l}{\textbf{mAP}} & \multicolumn{1}{l}{\textbf{top-1}} & \multicolumn{1}{l}{\textbf{mAP}} & \multicolumn{1}{l}{\textbf{top-1}} & \multicolumn{1}{l}{\textbf{mAP}} \\ \specialrule{0.1em}{0.45pt}{0.45pt}
0.4                           & 53.3                               & 21.7                             & 33.9                               & 5.8                              & 39.8                               & 23.1                             & 43.2                               & 24.3                             & 18.6                               & 17.6                             & 24.6                               & 11.2                             \\
0.6                           & 55.7                               & 21.5                             & 40.8                               & 6.9                              & 41.3                               & 22.6                             & 42.7                               & 24.0                             & 18.9                               & 19.4                             & 34.1                               & 16.0                             \\
0.8                           & 59.0                               & 21.7                             & 22.1                               & 5.4                              & 41.9                               & 24.2                             & 43.3                               & 24.1                             & 17.3                               & 18.5                             & 33.2                               & 14.4                             \\
1.0                           & 54.3                               & 21.0                             & 17.7                               & 4.0                              & 40.7                               & 22.2                             & 41.0                               & 23.8                             & 12.3                               & 15.1                             & 28.9                               & 12.0                             \\ \specialrule{0.1em}{0.2pt}{0.2pt}
\end{tabular}
}
}
\end{table*}

\section{Further Analysis}
\label{discussion}

\subsection{Ablation Study}
In this section, we conduct ablation study to demonstrate the necessity of each component in the proposed approach. We use raw skeleton sequences as the baseline and report average performance of each configuration. As reported in Table \ref{ablation}, we can draw the following conclusions. The model utilizing single-level skeleton graph with MSRL (Id = 2, 3) shows higher performance than the baseline (Id = 1) that directly uses raw body-joint sequences by $0.3$-$8.5\%$ top-1 accuracy and $0.7$-$7.7\%$ mAP on all datasets, regardless of using SPC. Such results demonstrate the effectiveness of graph representations, as it can model richer body structural information and mine valuable body-component relations to obtain a more discriminative skeleton representation. Compared with the model without contrastive learning (Id = 2), employing SPC (Id = 3) obtains consistent re-ID performance improvement by up to $9.8\%$ top-1 accuracy and $7.5\%$ mAP on different datasets. This justifies that the proposed SPC is a highly effective contrastive learning paradigm, which enables the model to mine more typical and unique skeleton features of different identities from the unlabeled graph representations for person re-ID. Exploiting the proposed multi-level graphs (Id = 4) performs better than solely using single-level graph (Id = 3) with a remarkable margin of $2.1$-$16.7\%$ top-1 accuracy and $0.4$-$5.7\%$ mAP, which demonstrates that modeling body structure and relations at various levels with the proposed graph representations (MG) can encourage the model \hc{to learn more useful skeleton features for person re-ID.} We will compare the effects of different level graphs in the next section as well. Adding FCRL further improves the overall performance in terms of \hc{both top-1 accuracy by $0.8$-$13.9\%$ and mAP by $0.5$-$2.6\%$} on different datasets. \hc{Such results verify our claim that combining structural and collaborative body relation learning can facilitate capturing richer features of body structure and skeleton patterns for the person re-ID task.}



 \begin{figure*}[t]
    \centering
    \subfigure[Collaborative Relations (KS20)]{\scalebox{0.5}{\label{CR_KS20}\quad\includegraphics[]{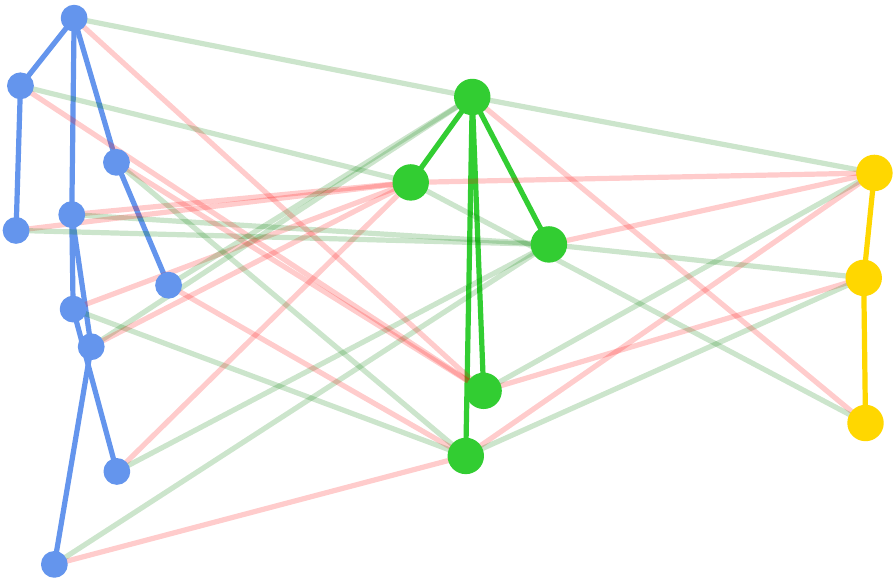}}}
    \quad  \quad \quad
     \subfigure[Collaborative Relations (BIWI)]{\scalebox{0.5}{\label{CR_BIWI}\quad\includegraphics[]{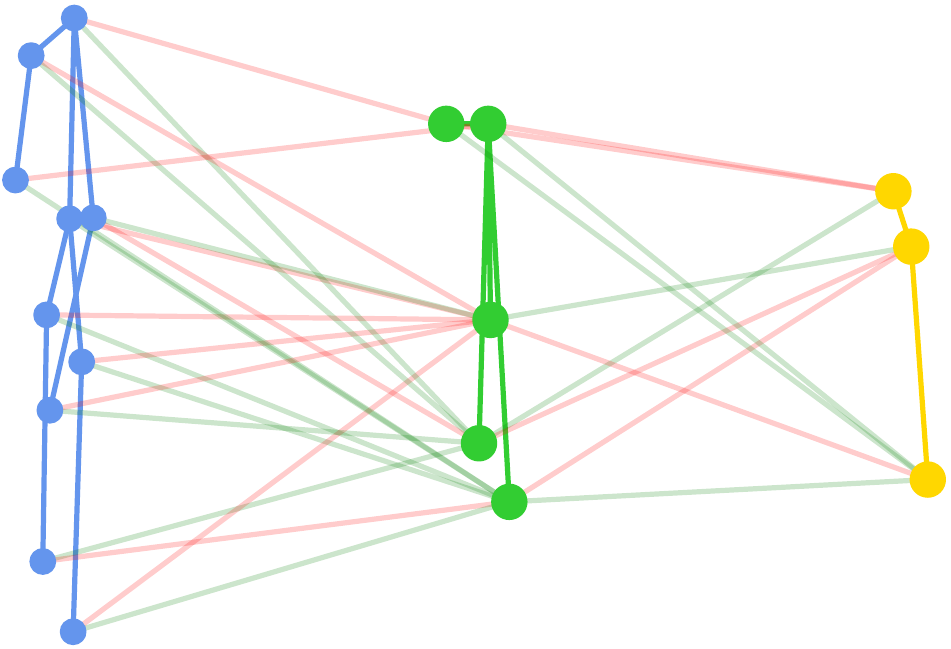}}}
     \quad  \quad \quad
      \subfigure[Collaborative Relations (IAS)]{\scalebox{0.5}{\label{CR_IAS}\quad\includegraphics[]{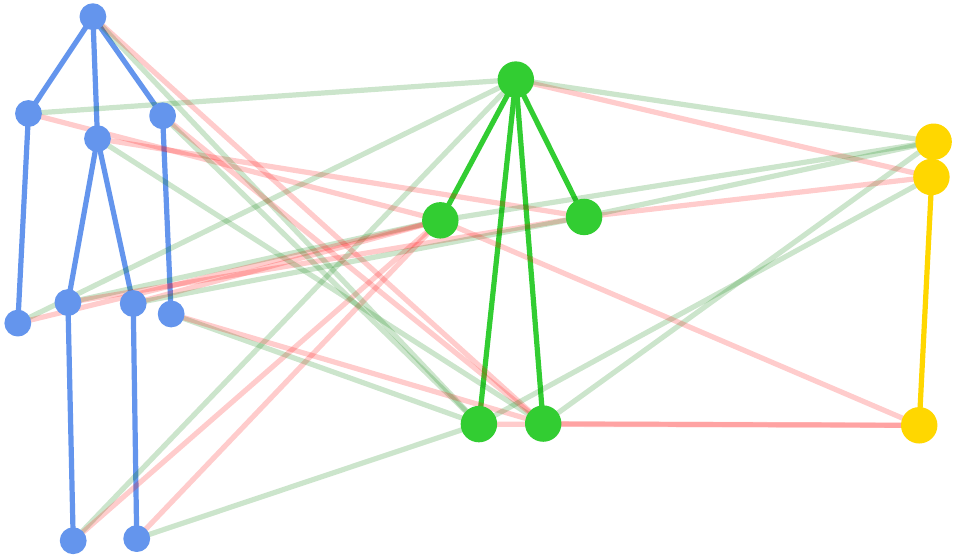}}}
  \subfigure[CR Matrix (KS20) ]{\scalebox{0.16}{\label{CR_matrice_KS20_1}\includegraphics[]{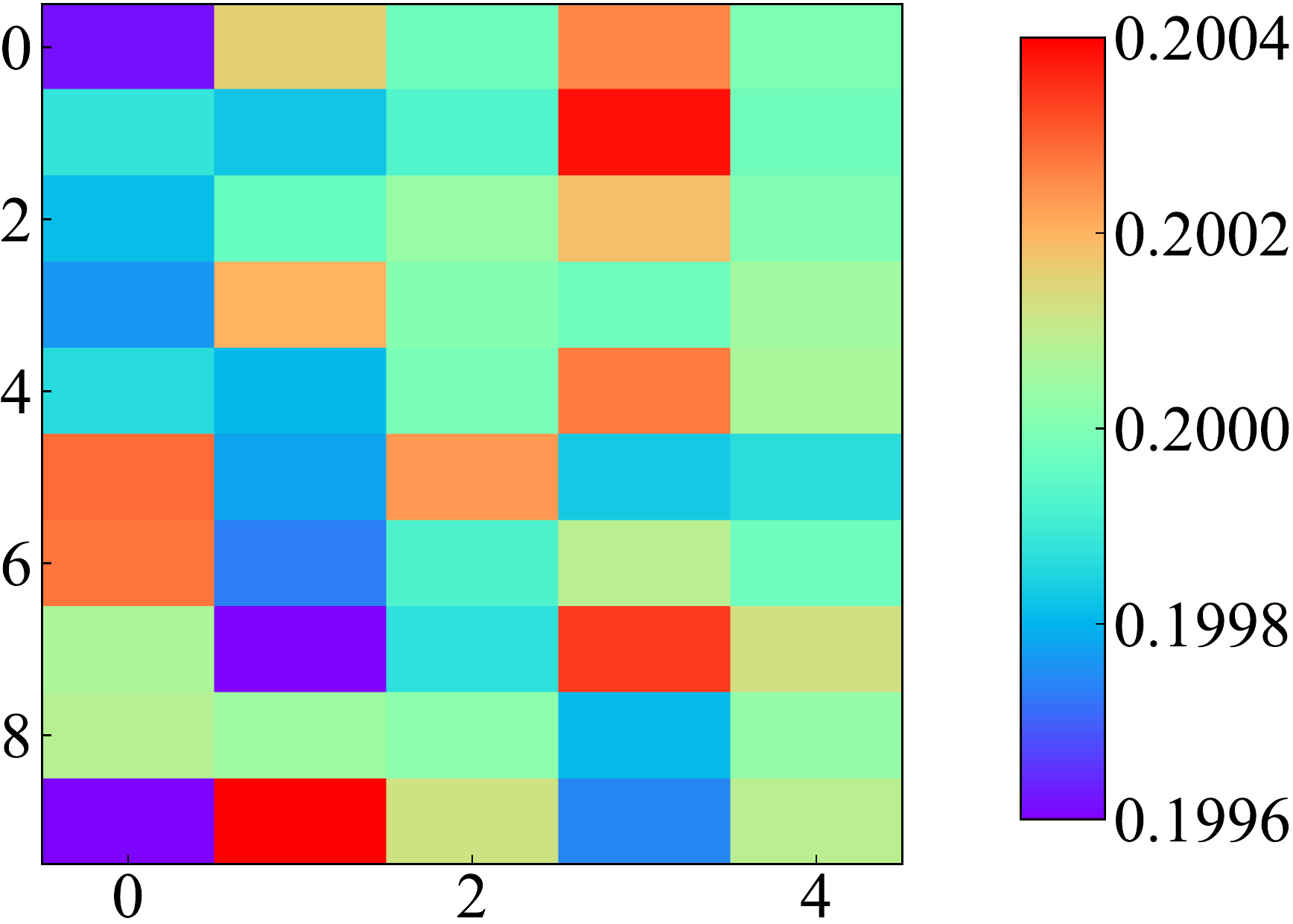}}}
 \
    \subfigure[CR Matrix (KS20)]{\scalebox{0.16}{\label{CR_matrice_KS20_2}\includegraphics[]{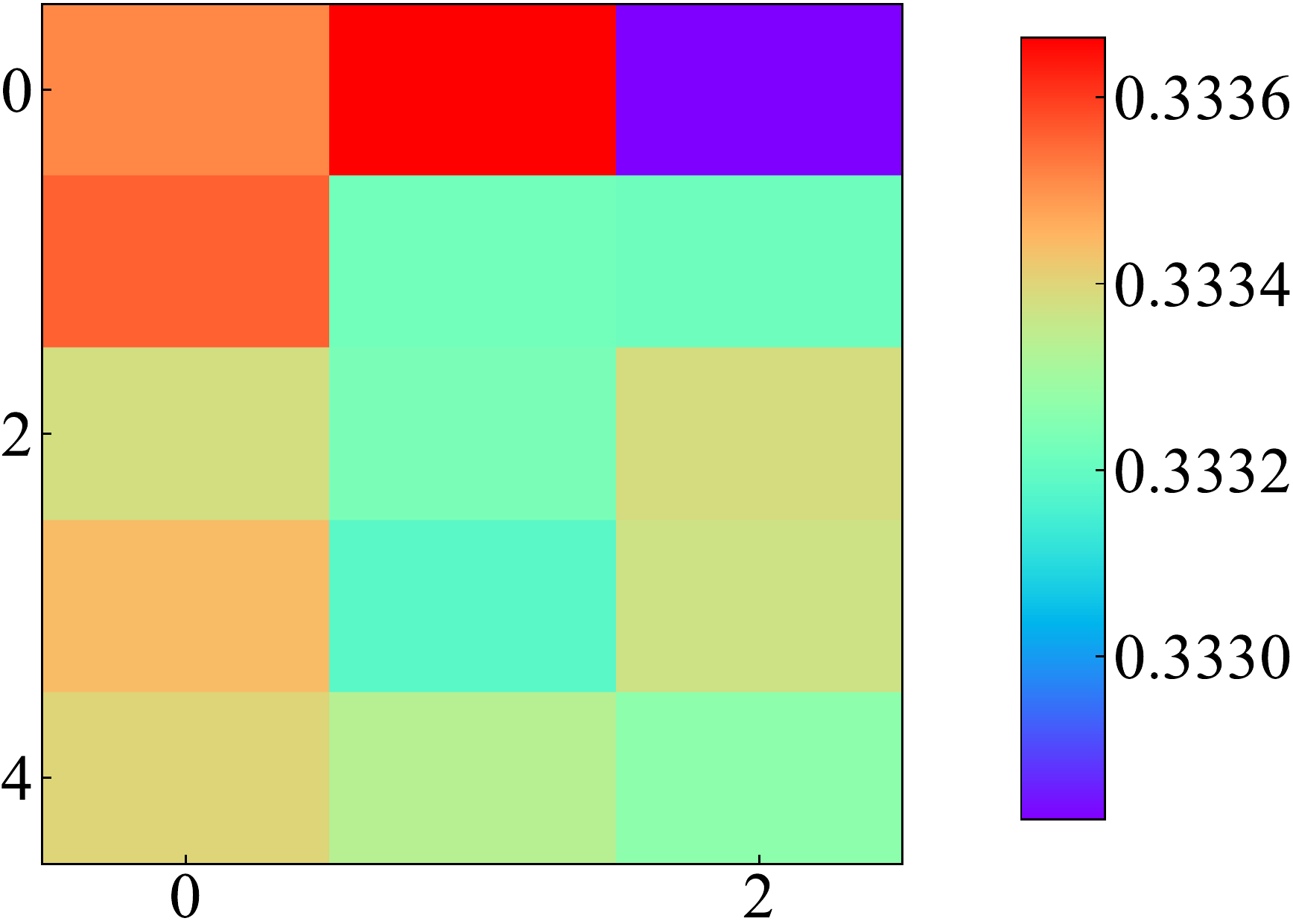}}}
 \ \subfigure[CR Matrix (BIWI) ]{\scalebox{0.16}{\label{CR_matrice_BIWI_1}\includegraphics[]{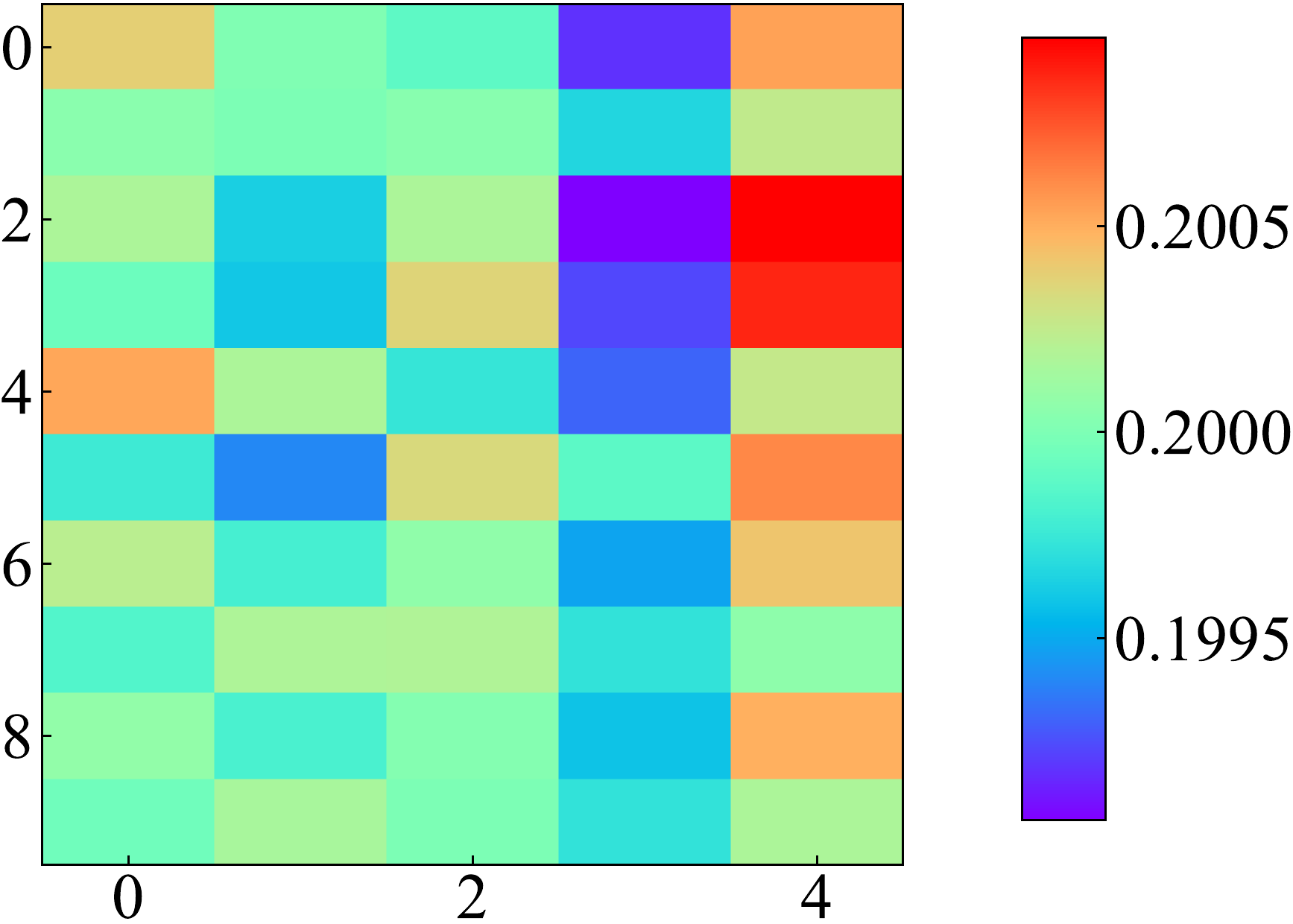}}}
 \
    \subfigure[CR Matrix (BIWI)]{\scalebox{0.16}{\label{CR_matrice_BIWI_2}\includegraphics[]{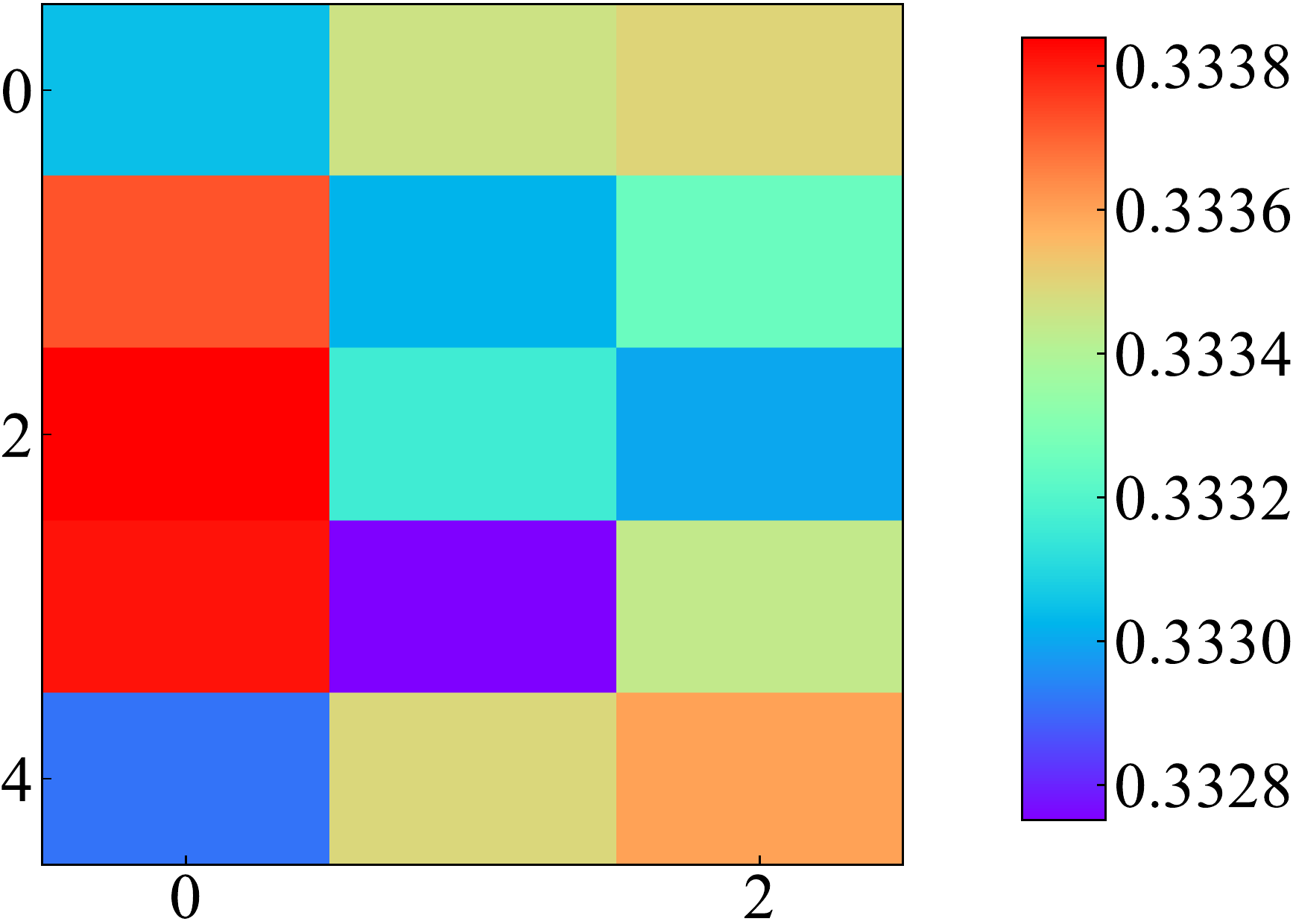}}}
  \   \subfigure[CR Matrix (IAS) ]{\scalebox{0.16}{\label{CR_matrice_IAS_1}\includegraphics[]{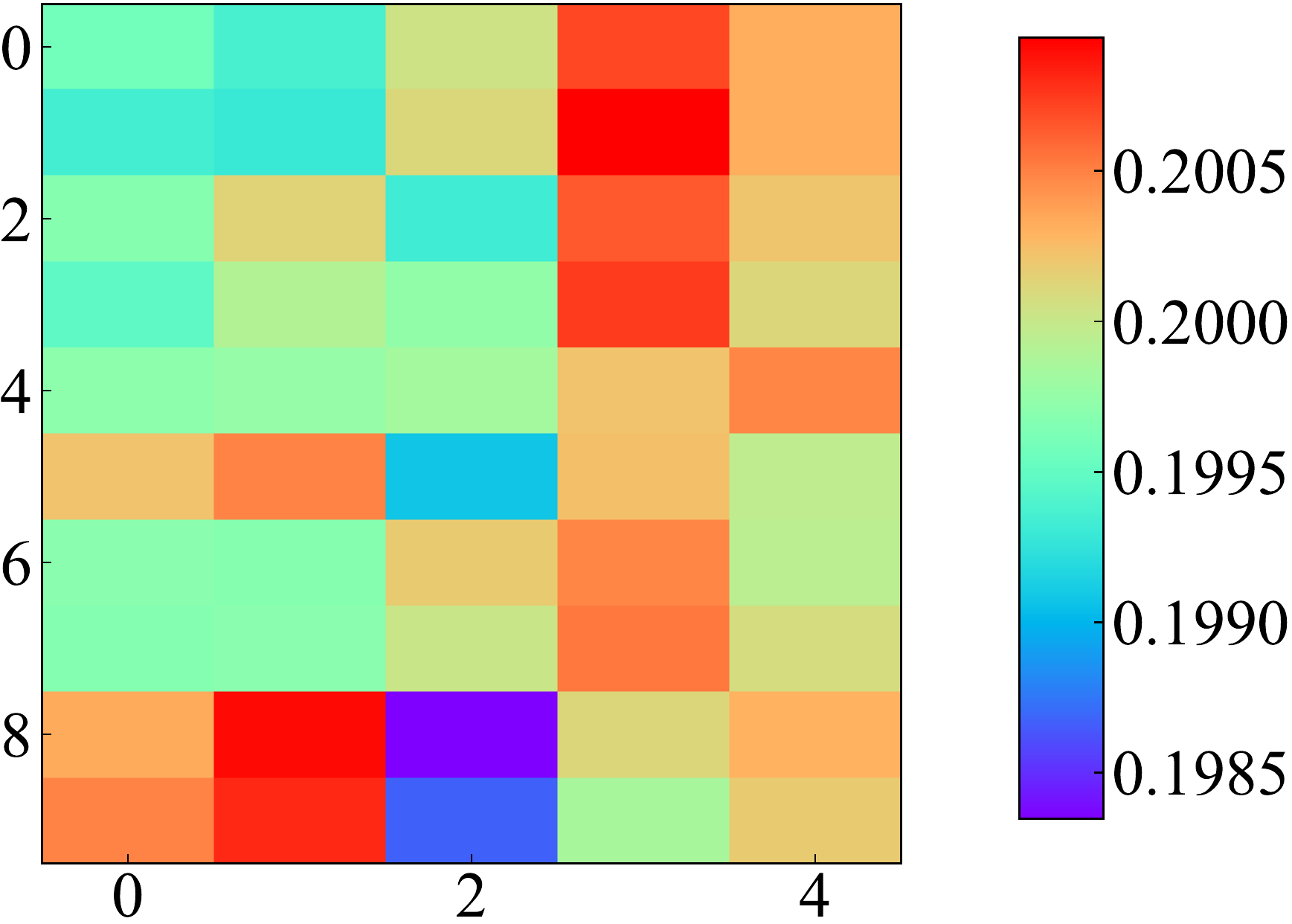}}}
 \
    \subfigure[CR Matrix (IAS)]{\scalebox{0.16}{\label{CR_matrice_IAS_2}\includegraphics[]{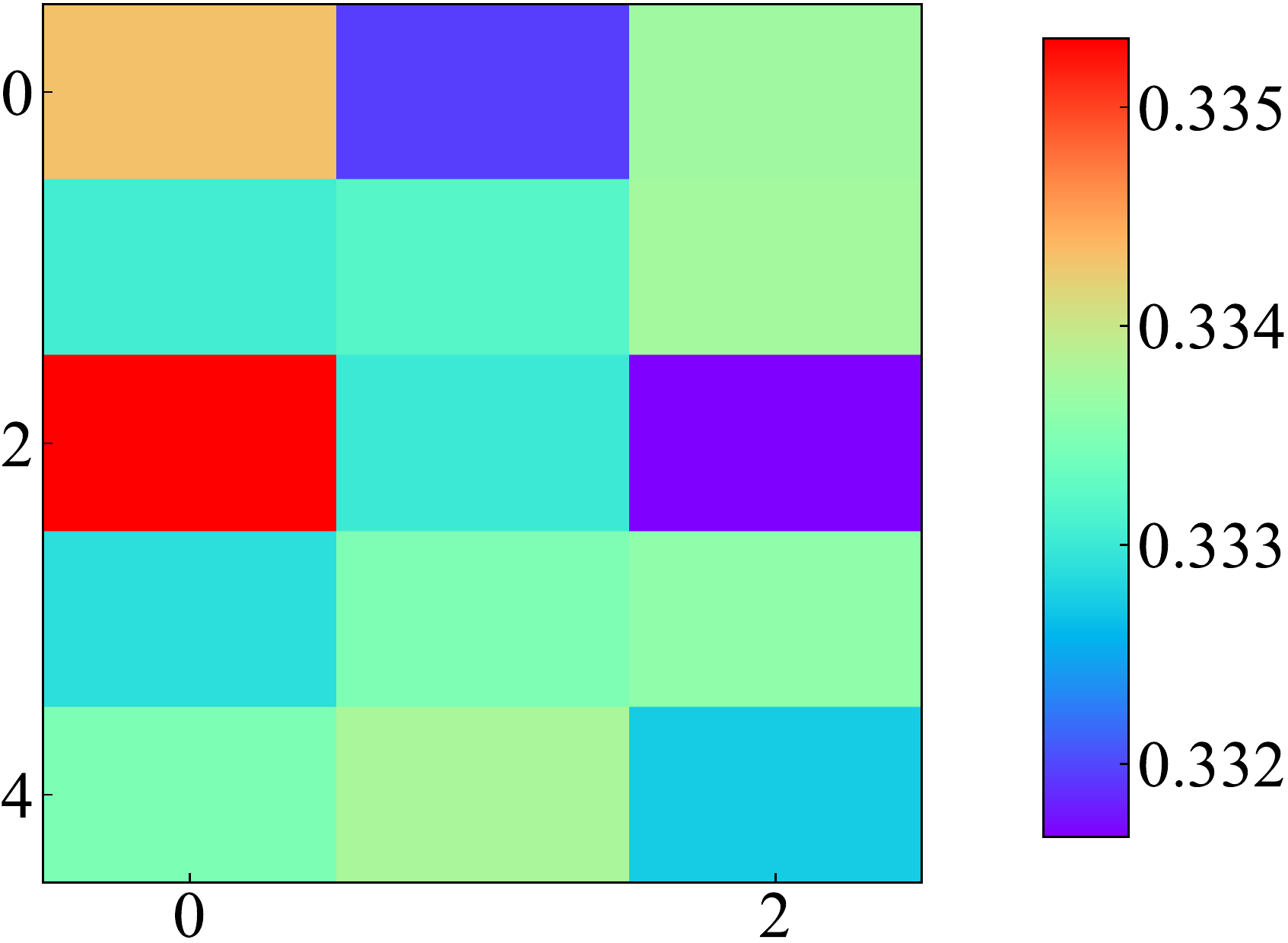}}}
    \caption{(a)-(c): Visualization of body-component collaborative relations (CR) across adjacent levels on KS20, BIWI and IAS. (d), (f), (h): CR value matrices between $\mathcal{G}_{1}$ and $\mathcal{G}_{2}$.  (e), (g), (i): CR value matrices between $\mathcal{G}_{2}$ and $\mathcal{G}_{3}$. The ordinate and abscissa denote indices of body-component nodes in low-level graph and high-level graphs. We visualize top-2 relation values in each matrix row by drawing red and green lines in (a)-(c).}
    \label{CR_relation_all}
\end{figure*}

\subsection{Discussions}

\subsubsection{Different Skeleton Graph Levels}
To analyze the effects of different graph levels, \hc{Table \ref{graph_level} evaluates} the performance of our approach when combining different graphs for person re-ID under the same optimal model setting. 
Exploiting low-level graph representations can achieve evidently better person re-ID performance than using high-level graphs by $1.3$-$19.0\%$ top-1 accuracy on most datasets, which suggests that the low-level graphs ($e.g.,$ part-level) that contain more body-component nodes and body structural/relational information can benefit learning more discriminative skeleton representations for person re-ID.
Combining different levels of graphs can improve the person re-ID performance compared with solely using single-level graphs (first three rows in Table \ref{graph_level}) in most cases, while the model exploiting all level graphs is the best performer among different datasets. This further justifies that the proposed multi-level graphs that model body structure and motion at various levels enable our approach to capture more recognizable structural features and pattern information for the person re-ID task.


 
 \subsubsection{Different Structural Relation Head Settings} 
 As shown in Table \ref{m}, we show the effects of different amounts of structural relation heads on our approach.
 It is observed that introducing more learnable structural relation heads \hc{with larger $m$ values} can improve the performance of our approach on different datasets, as it can encourage capturing more spatially or motion related features of adjacent body components.   
 However, too many structural relation heads\hc{, $m=16$, for example,} may cause the model to learn redundant relation information and \hc{degrade model performance} on relatively limited training data such as KS20.
 
  \subsubsection{Different Graph Fusion Coefficients}
  The hyper-parameter $\lambda_{C}^{a,b}$ controls the extent of fusing collaborative node features between $a^{th}$ level and $b^{th}$ level graphs. Here we use a unified $\lambda_{C}$
  \hc{to equally fuse features between different graph levels}, so as to better analyze the overall effects of collaborative fusion. As reported in Table \ref{lambda}, improving $\lambda_{C}$ enables our model to achieve better Re-ID performance in terms of both top-1 accuracy and mAP on all datasets. Such results verify the necessity of adequate collaborative graph fusion to learn more effective multi-level skeleton graph representations for person re-ID. Based on this observation, we set $\lambda_{C}^{a,b}=1$ to fuse skeleton graph features. 
 
\subsubsection{Temperature Settings for SPC Scheme}
 As shown in Table \ref{t}, the model using lower temperature $\tau$ can obtain slightly better person re-ID performance. Since higher temperatures tend to retain more similar features between different skeleton prototypes, it might reduce the capacity of contrastive learning on capturing inter-prototype differences and their underlying discriminative features. 
 This result is also consistent with our previous analysis that the proposed SPC scheme under an appropriate temperature can encourage learning higher mRCL, $i.e.,$ inter-class differences, and more effective skeleton representations for the person re-ID task. We also found that a high temperature could lead to an evident degradation of model performance on the KGBD dataset that contains more individuals. \hc{In practice,} we select the best model temperature for datasets of different sizes to achieve more balanced and better person re-ID performance.

 \subsubsection{Different Settings for DBSCAN} 
 We adjust the sample density, $i.e.$ minimum sample amount $a_{min}$ within the maximum distance $\epsilon$, in the DBSCAN algorithm to encourage more balanced and stable clustering. In Table \ref{a_min} and Table \ref{eps}, \hc{it is seen that} relatively lower $a_{min}$ and higher $\epsilon$, which enhances the connectedness of skeleton instances in larger clusters and generates less prototypes, can boost the model performance in most cases. However, on larger datasets such as KGBD, setting a high $\epsilon$ is shown to reduce the overall performance, while adopting lower $\epsilon$ and higher $a_{min}$, $i.e.,$ \hc{improving} the amount of prototypes, can facilitate person re-ID performance. It suggests that more diverse skeleton prototypes may encourage mining richer discriminative features in the case of more identities.

\begin{figure*}[t]
    \centering
        \subfigure[Training mACT]{\scalebox{0.305}{\includegraphics[]{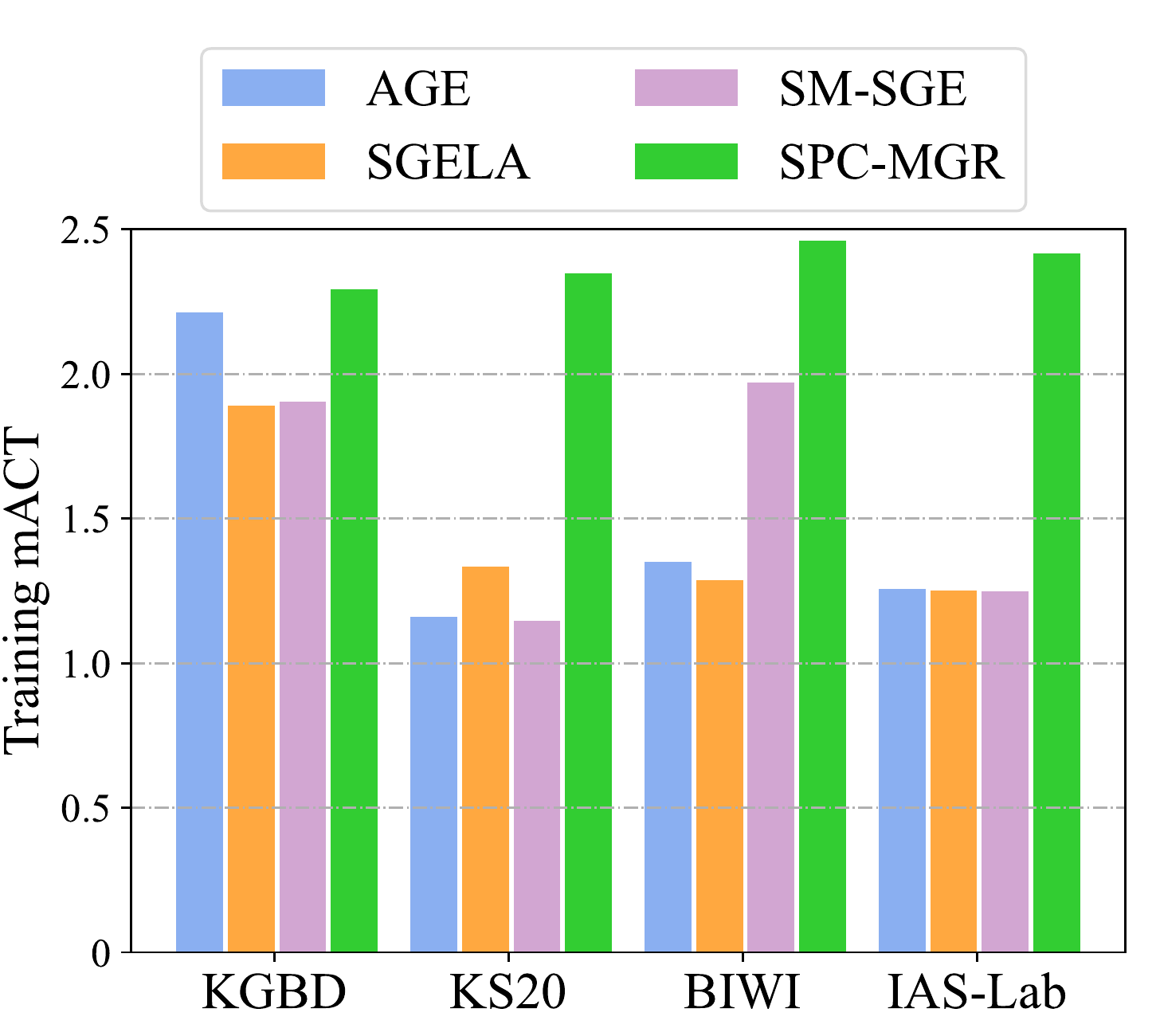}}} 
     \ 
    \subfigure[Testing mACT]{\scalebox{0.305}{\label{test_mACT}\includegraphics[]{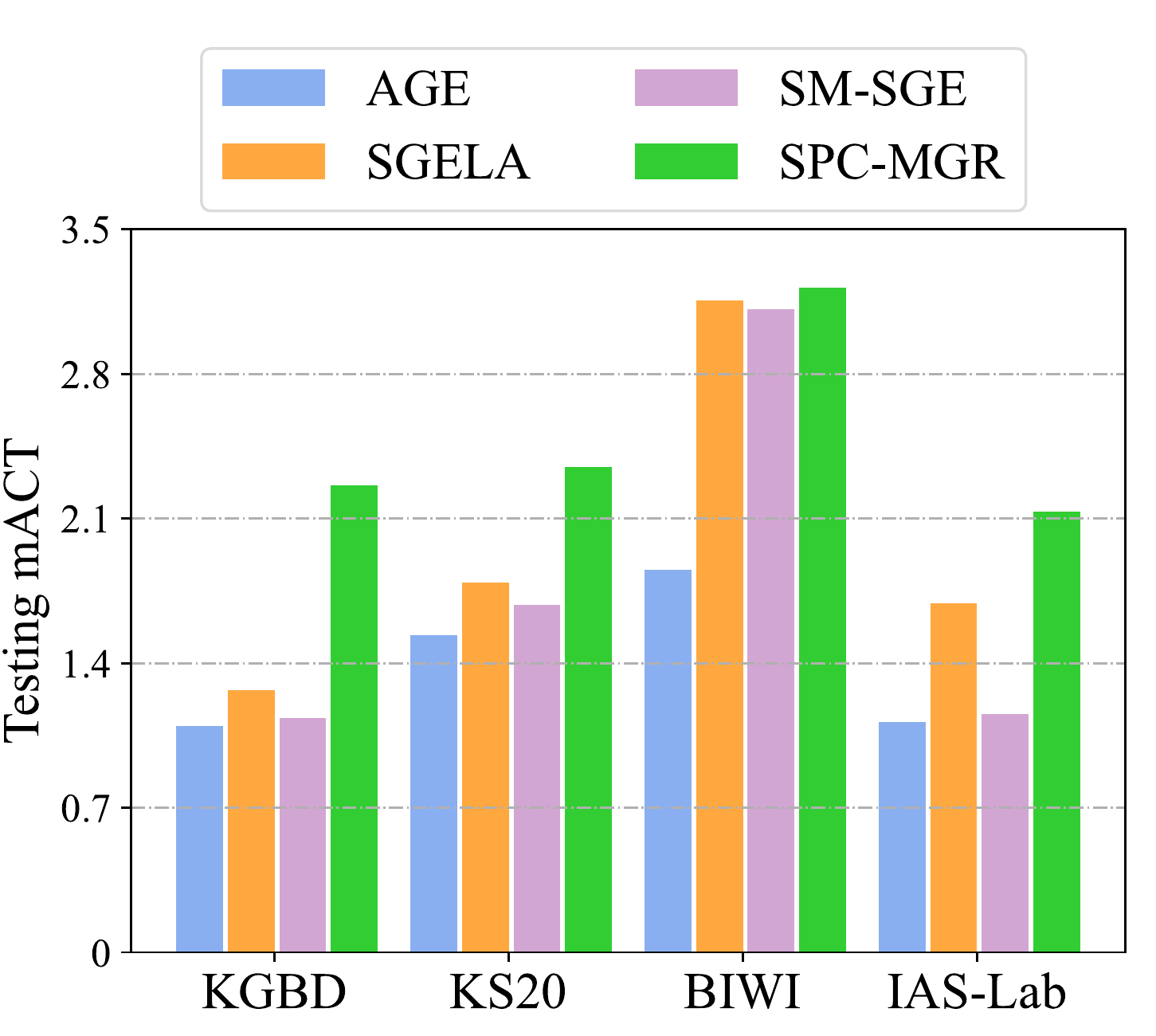}}} 
         \ 
    \subfigure[Training mRCL]{\scalebox{0.305}{\includegraphics[]{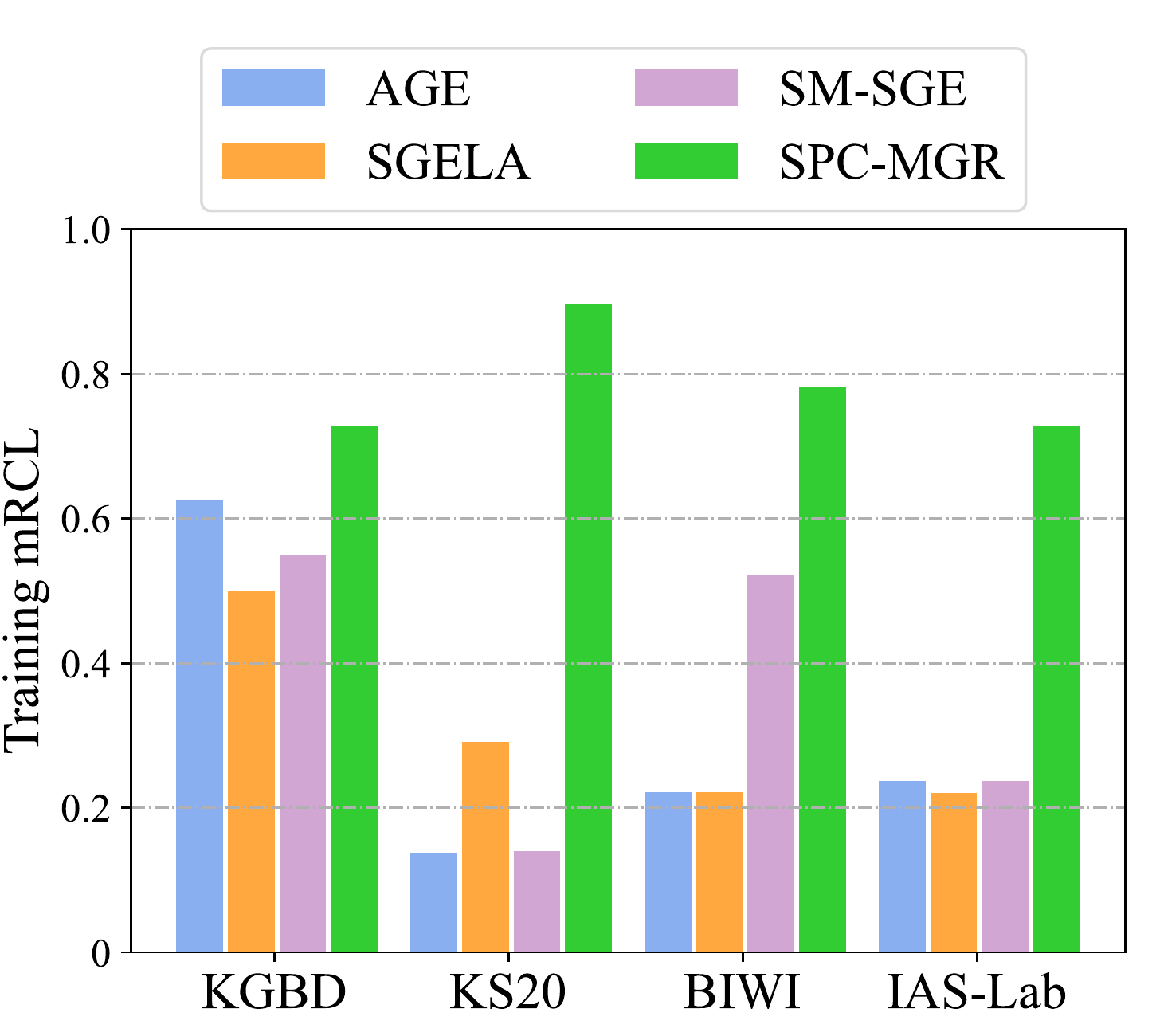}}} 
         \ 
    \subfigure[Testing mRCL]{\scalebox{0.305}{\label{test_mRCL}\includegraphics[]{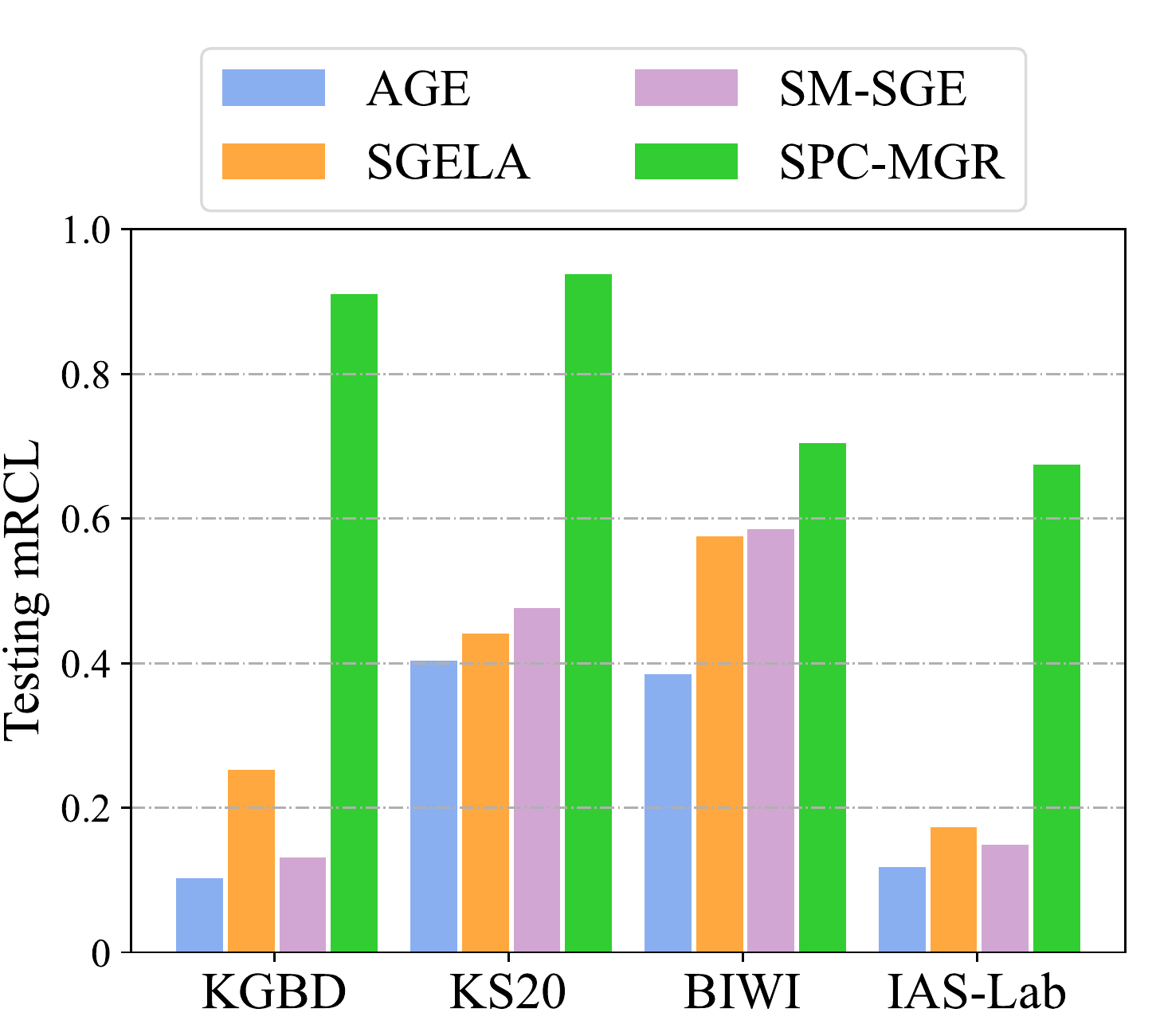}}} 
    \caption{Mean intra-class tightness (mACT) and mean inter-class looseness (mRCL) comparison with existing self-supervised and unsupervised methods (AGE, SGELA, SM-SGE) on different datasets. Note: We report the average mACT and mRCL of all testing sets of BIWI and IAS-Lab.}
    \label{mACT_mICL_comp}
\end{figure*}

\begin{figure*}[t]
    \centering
      \subfigure[AGE (BIWI)]{\scalebox{0.23}{\label{AGE_1_10_BIWI}\includegraphics[]{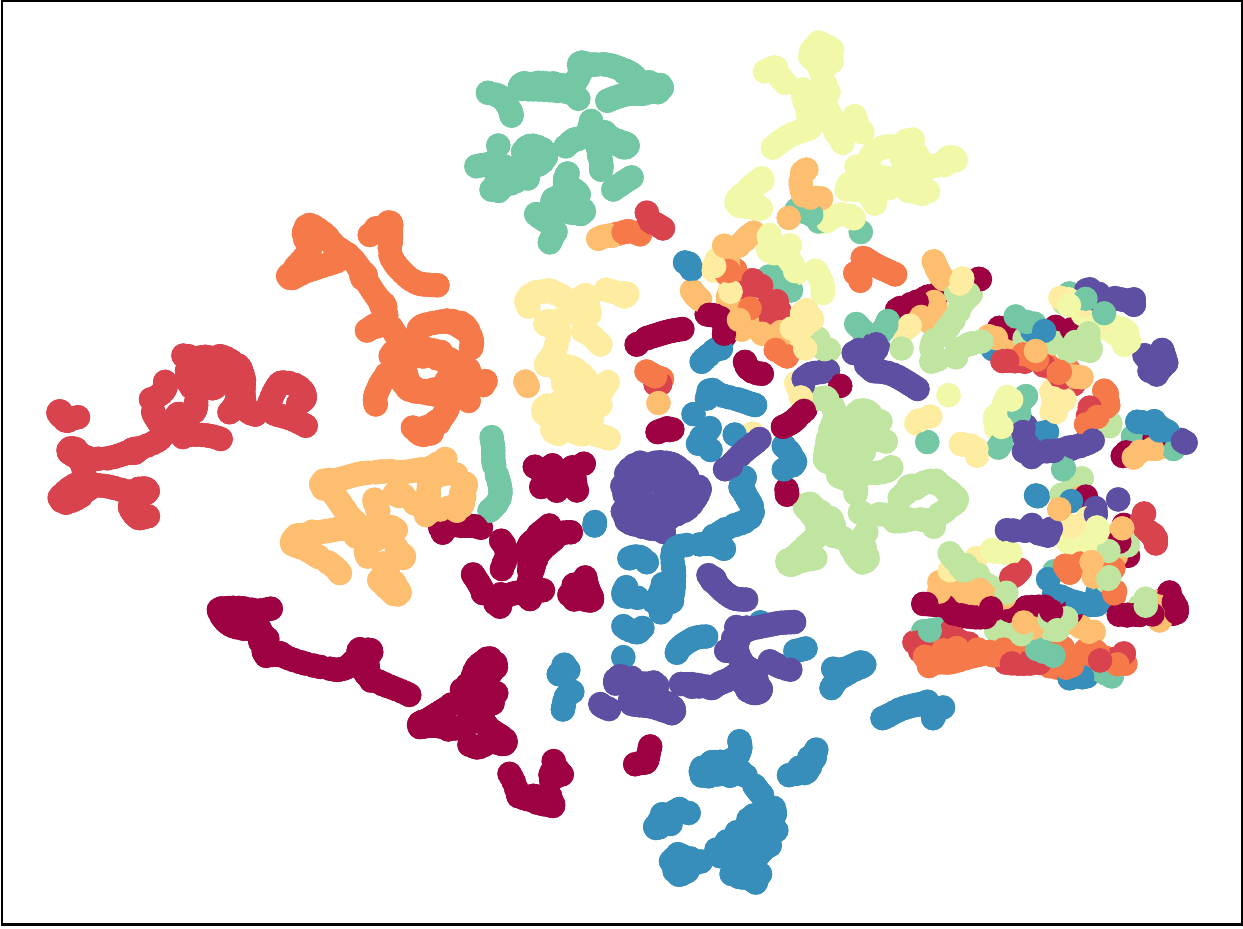}}} 
       \subfigure[SM-SGE (BIWI)]{\scalebox{0.23}{\label{SMSGE_1_10_BIWI}\includegraphics[]{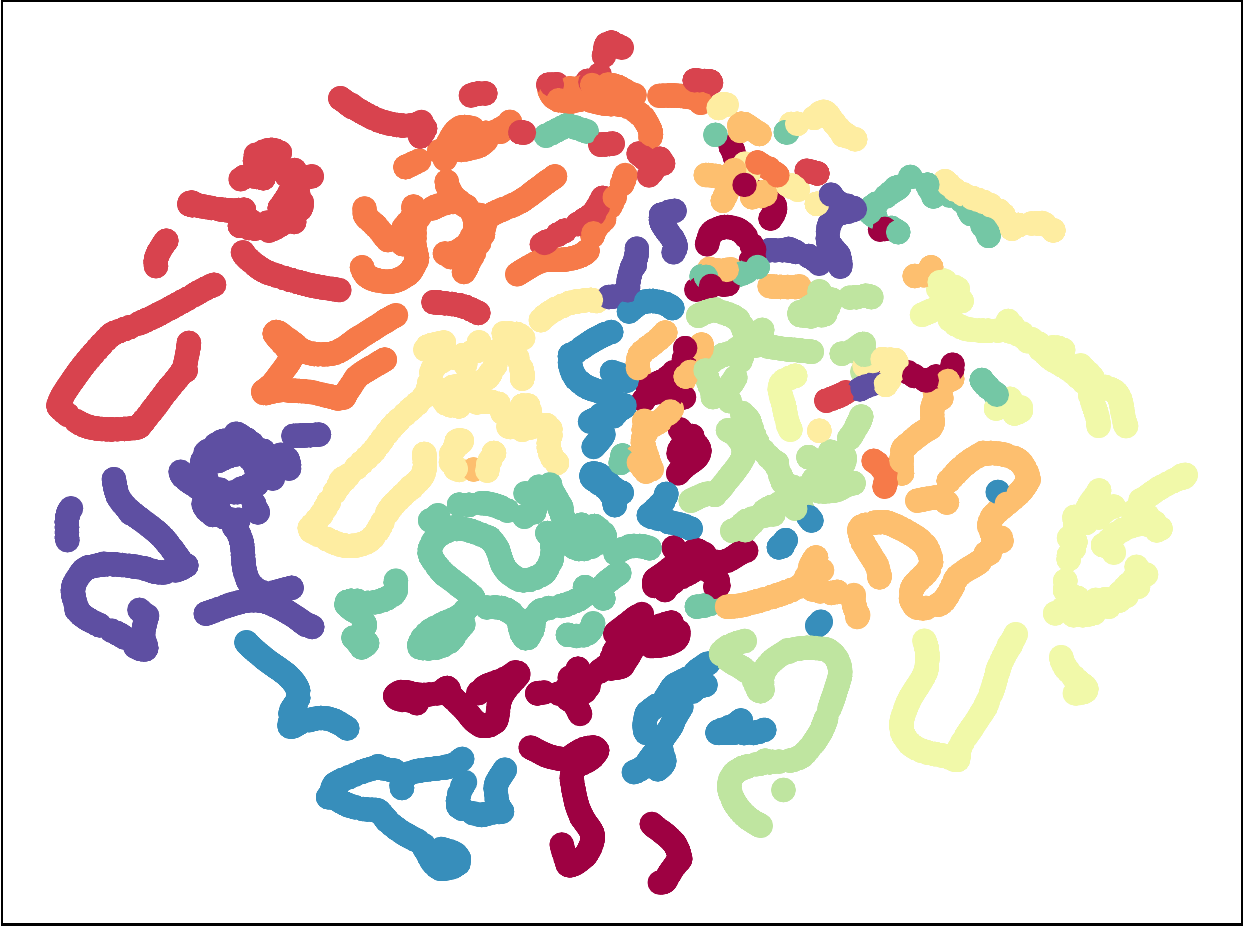}}} 
       \subfigure[SPC-MGR (BIWI)]{\scalebox{0.23}{\label{SPC_1_10_BIWI}\includegraphics[]{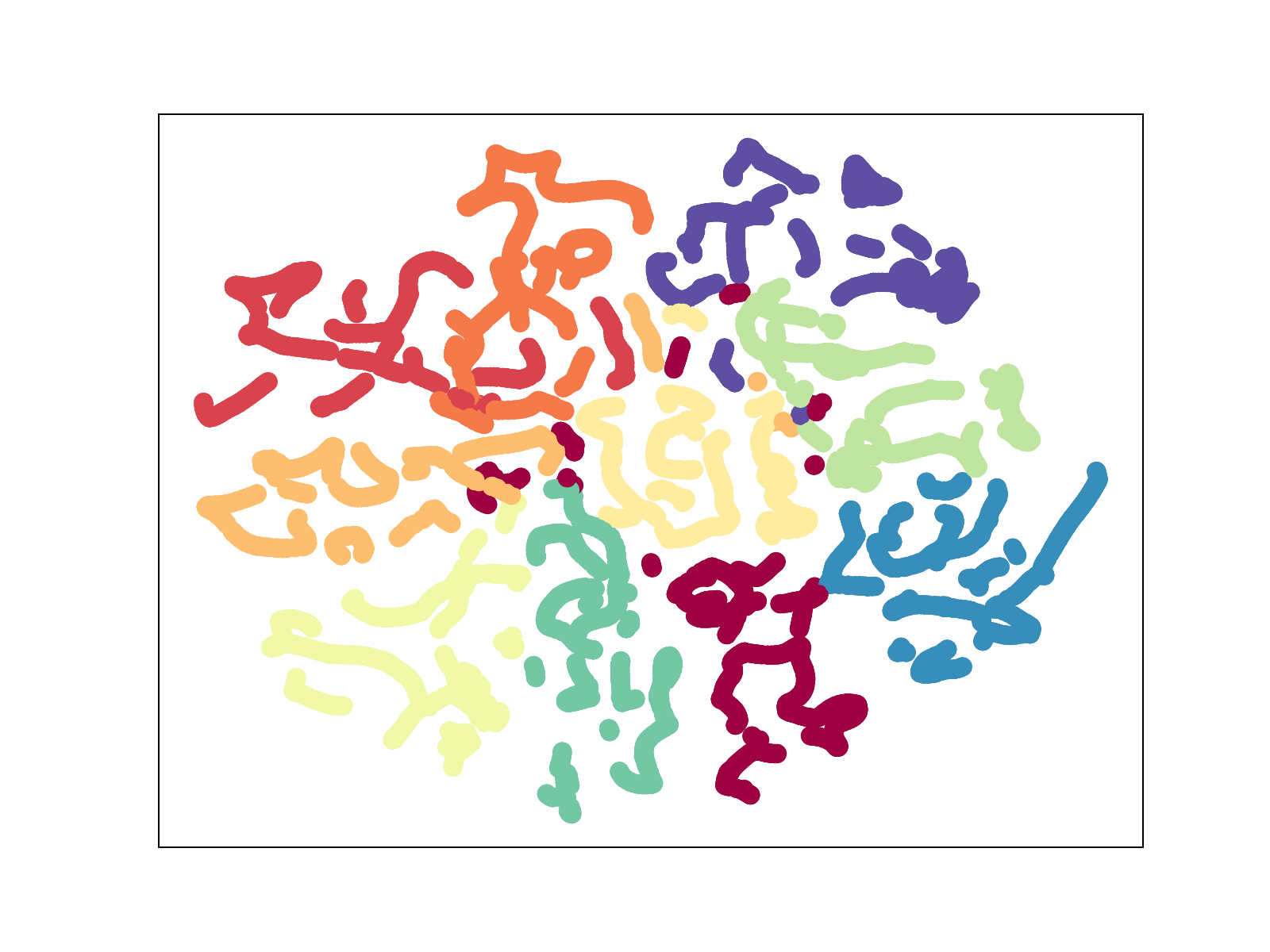}}} \ \
        \subfigure[AGE (KS20)]{\scalebox{0.23}{\label{AGE_1_10_KS20}\includegraphics[]{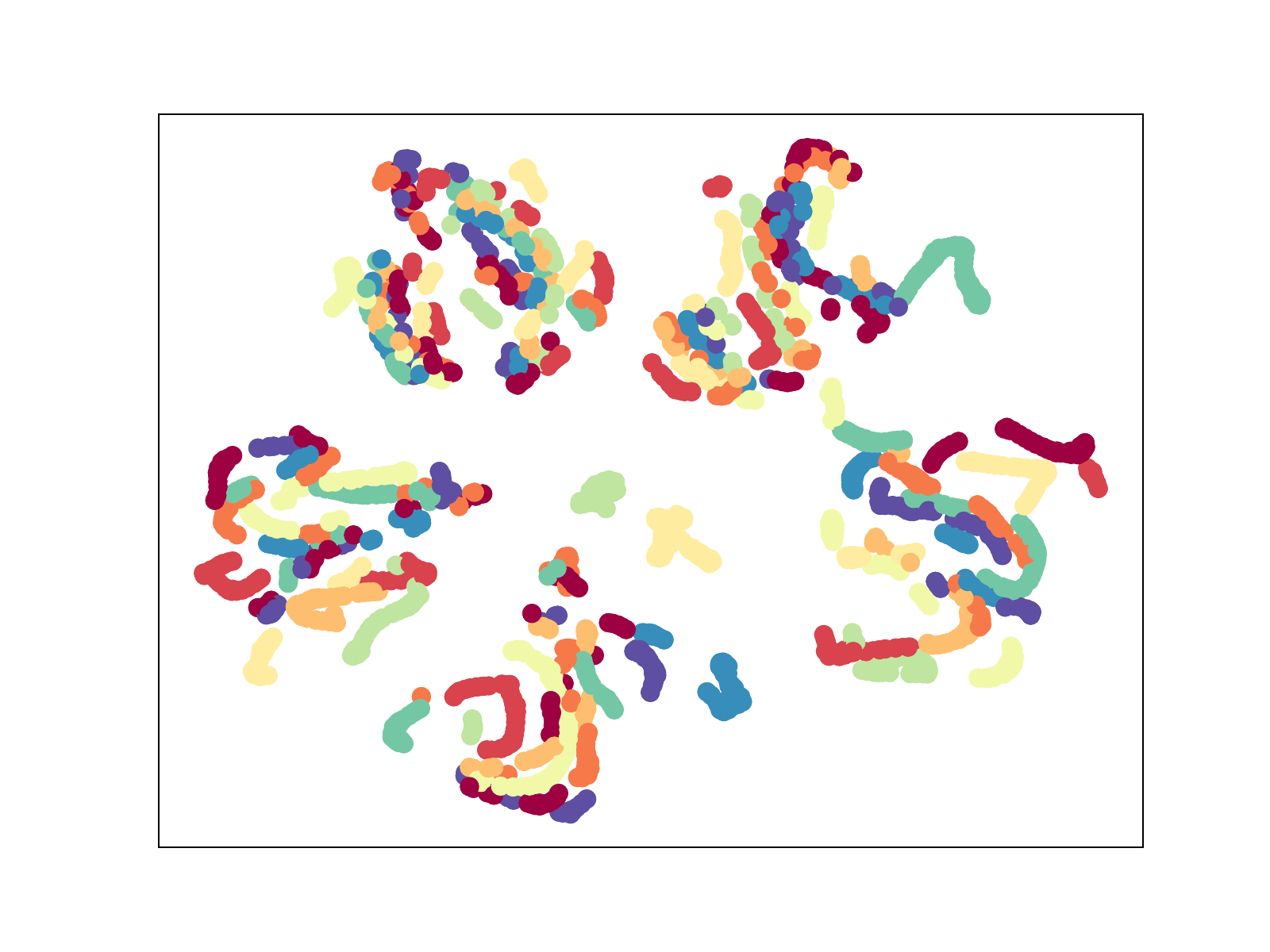}}}
       \subfigure[SM-SGE (KS20)]{\scalebox{0.23}{\label{SMSGE_1_10_KS20}\includegraphics[]{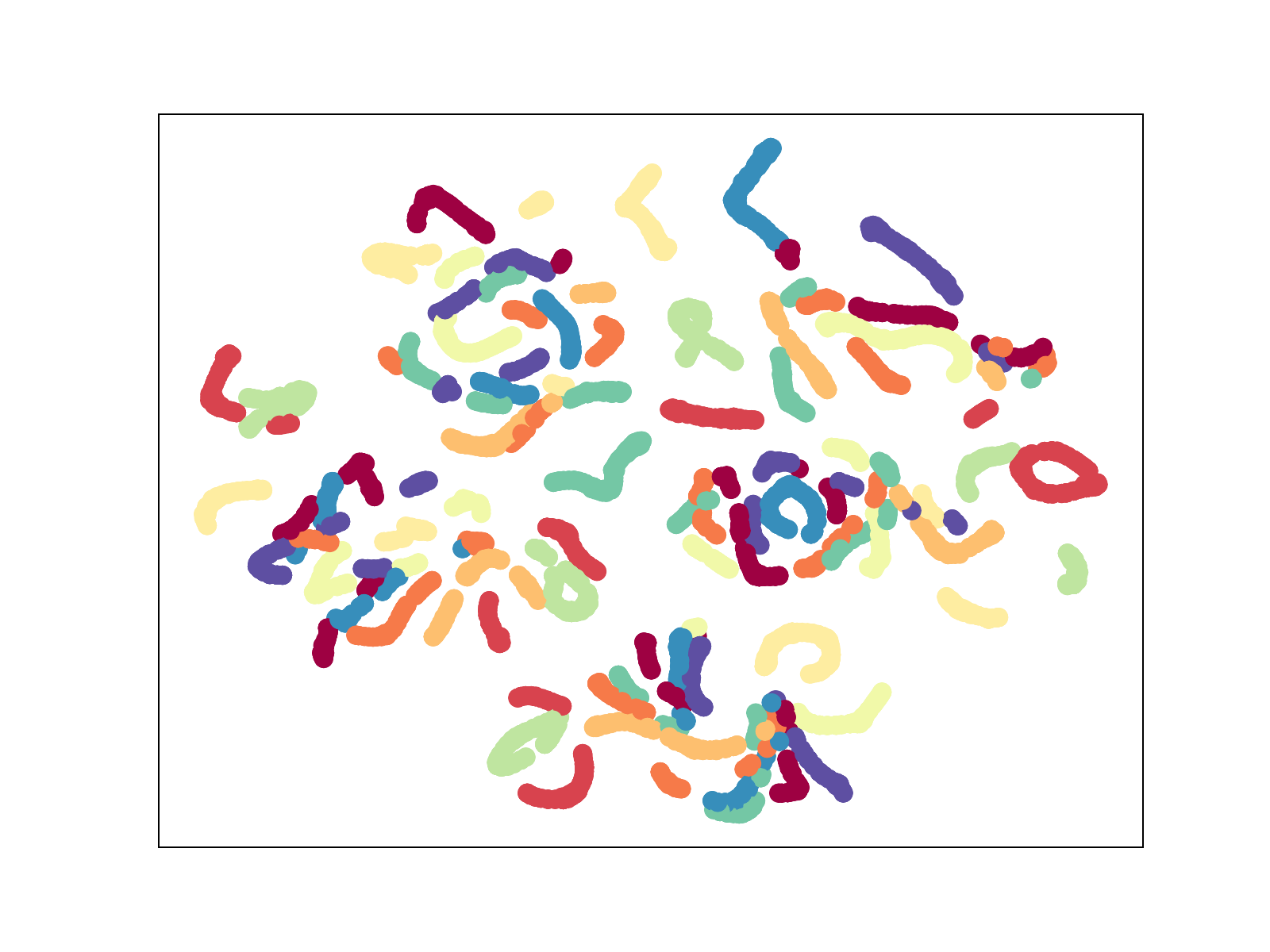}}} 
       \subfigure[SPC-MGR (KS20)]{\scalebox{0.23}{\label{SPC_1_10_KS20}\includegraphics[]{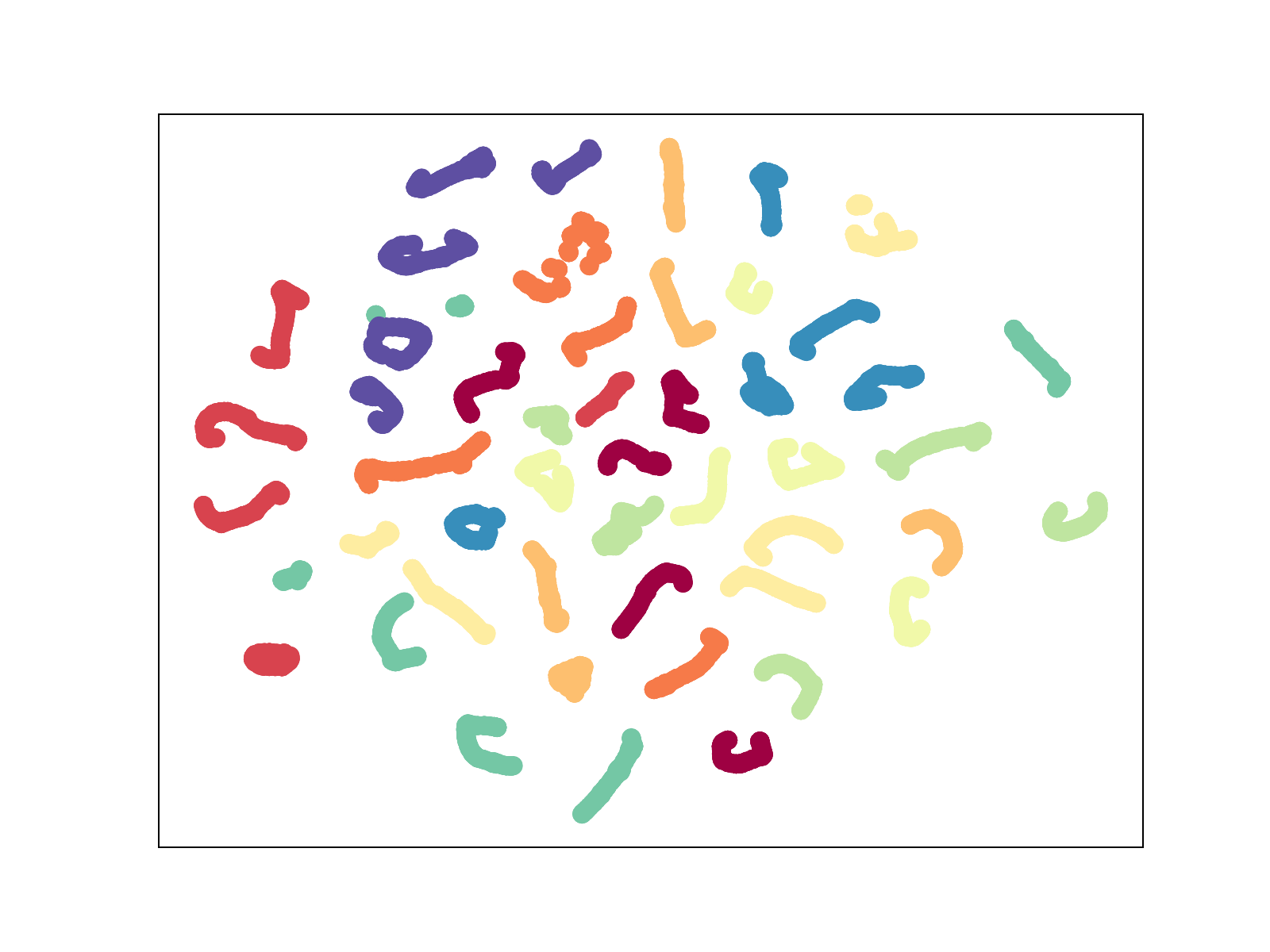}}}
    \caption{T-SNE visualization of the skeleton representations learned by AGE ((a),(d)), SM-SGE ((b), (e)), and the proposed SPC-MGR ((c), (f)) for the first 10 classes in BIWI and KS20 datasets. Note: Different colors indicate skeleton representations of different classes.}
    \label{TNS_comp}
\end{figure*}

\subsection{Analysis of Body Relation Learning}
\label{body_relation_analysis}
 To validate the ability of our approach on inferring body-component relations between different graphs, we visualize collaborative relations and corresponding relation value matrices of different graphs in Fig.  \ref{CR_relation_all}. Here we take collaborative relations between adjacent graph levels as an example for analysis. As shown in \hc{Fig. \ref{CR_relation_all} (a)-(c)}, 
 we can see that our approach can learn to assign larger relation values to significantly moving body partitions such as feet, hips, hands, and elbows (nodes 0, 1, 2, 3, 6, 7, 8 \hc{and 9} in $\mathcal{G}_{1}$), which typically perform collaborative motion during walking process and can carry abundant and unique motion features. 
 By inferring inherent relations between these body components and fusing them into multi-level graph representations, our model is able to focus on crucial patterns of skeletal motion to learn more effective skeleton representations.
Our approach can not only capture relations of major body components to their corresponding high-level components, but also allocate different attention to potentially motion-related components. For example, as shown in Fig. \ref{CR_matrice_KS20_1}, Fig. \ref{CR_matrice_BIWI_1} and Fig. \ref{CR_matrice_IAS_1}, the movement of arms (nodes 3 and 4 in $\mathcal{G}_{2}$) is highly correlated with the leg motion (nodes 0, 1, 2 and 3 in $\mathcal{G}_{1}$), and different parts of them enjoy varying degree of collaboration in walking. 
Such results demonstrate that our approach can capture body-related high-level semantics ($e.g.,$ body-component correspondence) and meanwhile facilitate global pattern learning of correlative components. \hc{More graph details and visualization results are provided in the Appendix.}

\begin{table*}[t]
\centering
\caption{Performance comparison with appearance-based and skeleton-based methods on CASIA-B.
Note: ``Cl-Nm'' denotes the probe set under ``Clothes'' condition and gallery set under ``Normal'' condition. $^{\dagger}$ refers to appearance-based methods and $^{*}$ represents requiring label information for training. ``—'' indicates no published result. } 
\label{CASIA_B_result}
\scalebox{0.825}{
\setlength{\tabcolsep}{0.58mm}{
\begin{tabular}{lcccccccccccccccccccc}
\specialrule{0.1em}{0.45pt}{0.45pt}
\textbf{Probe-Gallery} & \multicolumn{4}{c}{\textbf{Nm-Nm}}                               & \multicolumn{4}{c}{\textbf{Bg-Bg}}                                & \multicolumn{4}{c}{\textbf{Cl-Cl}}                                & \multicolumn{4}{c}{\textbf{Cl-Nm}}                               & \multicolumn{4}{c}{\textbf{Bg-Nm}}                               \\ \specialrule{0.1em}{0.45pt}{0.45pt}
\textbf{Methods}       & \textbf{top-1} & \textbf{top-5} & \textbf{top-10} & \textbf{mAP} & \textbf{top-1} & \textbf{top-5} & \textbf{top-10} & \textbf{mAP}  & \textbf{top-1} & \textbf{top-5} & \textbf{top-10} & \textbf{mAP}  & \textbf{top-1} & \textbf{top-5} & \textbf{top-10} & \textbf{mAP} & \textbf{top-1} & \textbf{top-5} & \textbf{top-10} & \textbf{mAP} \\ \specialrule{0.1em}{0.45pt}{0.45pt}
$^{\dagger}$LMNN$^{*}$ \cite{weinberger2009distance}                   & 3.9            & 22.7           & 36.1            & —            & 18.3           & 38.6           & 49.2            & —             & 17.4           & 35.7           & 45.8            & —             & 11.6           & 12.6           & 17.8            & —            & 23.1           & 37.1           & 44.4            & —            \\
$^{\dagger}$ITML$^{*}$ \cite{davis2007information}                   & 7.5            & 22.2           & 34.2            & —            & 19.5           & 26.0           & 33.7            & —             & 20.1           & 34.4           & 43.3            & —             & 10.3           & 24.5           & 36.1            & —            & 21.8           & 30.4           & 36.3            & —            \\
$^{\dagger}$ELF$^{*}$ \cite{gray2008viewpoint}                    & 12.3           & 35.6           & 50.3            & —            & 5.8            & 25.5           & 37.6            & —             & 19.9           & 43.9           & 56.7            & —             & 5.6            & 16.0           & 26.3            & —            & 17.1           & 30.0           & 37.9            & —            \\
$^{\dagger}$SDALF \cite{farenzena2010person}                  & 4.9            & 27.0           & 41.6            & —            & 10.2           & 33.5           & 47.2            & —             & 16.7           & 42.0           & 56.7            & —             & 11.6           & 19.4           & 27.6            & —            & 22.9           & 30.1           & 36.1            & —            \\
$^{\dagger}$Score-based MLR$^{*}$ \cite{liu2015enhancing}        & 13.6           & 48.7           & 63.7            & —            & 13.6           & 48.7           & 63.7            & —             & 13.5           & 48.6           & 63.9            & —             & 9.7            & 27.8           & 45.1            & —            & 14.7           & 32.6           & 50.2            & —            \\
$^{\dagger}$Feature-based MLR$^{*}$ \cite{liu2015enhancing}      & 16.3           & 43.4           & 60.8            & —            & 18.9           & 44.8           & 59.4            & —             & 25.4           & 53.3           & 68.9            & —             & 20.3           & \textbf{42.6}  & \textbf{56.9}   & —            & 31.8  & \textbf{53.6}  & \textbf{64.1}   & —            \\
AGE \cite{rao2020self}                    & 20.8           & 29.3           & 34.2            & 3.5          & 37.1           & 56.2           & 67.0            & 9.8           & 35.5           & 54.3           & 65.3            & 9.6           & 14.6           & 33.0           & 42.7            & 3.0          & \textbf{32.4}           & 51.2           & 60.1            & 3.9          \\
SM-SGE \cite{rao2021sm}                 & 50.2           & 73.5           & 81.9            & 6.6          & 26.6           & 49.0           & 59.4            & 9.3           & 27.2           & 51.4           & 63.2            & 9.7           & 10.6           & 26.3           & 35.9            & 3.0          & 16.6           & 36.8           & 47.5            & 3.5          \\
SPC-MGR (Ours)          & \textbf{71.2}  & \textbf{88.0}  & \textbf{92.8}   & \textbf{9.1} & \textbf{44.3}  & \textbf{66.4}  & \textbf{76.4}   & \textbf{11.4} & \textbf{48.3}  & \textbf{71.6}  & \textbf{81.6}   & \textbf{11.8} & \textbf{22.4}  & 40.4           & 51.0            & \textbf{4.3} & 28.9           & 49.3           & 59.1            & \textbf{4.6} \\ \specialrule{0.1em}{0.2pt}{0.2pt}
\end{tabular}
}
}
\end{table*}

\subsection{Analysis of Skeleton Representations}
\label{qq_analysis}
\subsubsection{Intra-Class Tightness and Inter-Class Looseness}
In this section, we further analyze the effectiveness of the learned skeleton representations with the proposed mACT and mRCL metrics, and compare our approach with existing self-supervised and unsupervised methods, AGE \cite{rao2020self}, SGELA \cite{rao2021self}, SM-SGE \cite{rao2021sm}), on all datasets. As shown in Fig. \ref{mACT_mICL_comp}, we can obtain the following crucial analysis. The representations learned by the proposed SPC-MGR enjoy the highest mACT and mRCL among all methods on different datasets, which indicates that our approach is able to learn higher similarity for the same-class instances while improving the distance between different classes. Such results also substantiate our intuition on SPC scheme that maximizing intra-prototype similarity and inter-prototype dissimilarity can encourage capturing both similar and unique semantics of ground-truth classes in skeleton representation learning. The model obtaining both higher mACT and mRCL can also achieve better person re-ID performance in most cases. For example, \hc{our method, SGELA, and SM-SGE achieve higher top-1 accuracy and mAP than AGE on KS20 and BIWI testing sets (see Table \ref{KS20_KGBD_IASA} and Table \ref{IASB_BIWIS_BIWIW}), which is consistent with results of mACT and mRCL shown in Fig. \ref{test_mACT} and Fig. \ref{test_mRCL}.} This further verifies that the proposed mACT and mRCL can serve as auxiliary evaluation metrics to help analyze the effectiveness of learned representations in person re-ID tasks.

\subsubsection{Visualization of Skeleton Representations}
We conduct a T-SNE \cite{van2008visualizing} visualization of skeleton representations for a qualitative analysis, and compare our approach with two state-of-the-art skeleton-based methods, $i.e.$, AGE \cite{rao2020self}, SM-SGE \cite{rao2021sm}. As presented in Fig. \ref{SPC_1_10_BIWI}, the skeleton representations learned by our approach can form different class clusters with higher separation than AGE and SM-SGE on BIWI, which suggests the lower entropy of our representations. Interestingly,
it is observed that the learned representations on KS20 are separated in small groups of the same class, as shown in Fig. \ref{SPC_1_10_KS20}, which \hc{enjoys} significantly larger looseness than other two methods. Such results imply that our model may learn skeleton representations with finer separation \hc{and enable pattern-based grouping} in a specific class.


\begin{table*}[t]
\centering
\caption{Generalized person re-ID performance of our approach with direct domain generalization (DG) from source datasets (``Source'') to target datasets (``Target''). ``UF'' represents fine-tuning the source model with the unlabeled data of target datasets. BIWI-W/S denotes the Walking/Still testing set of BIWI. Bold numbers indicates that the model using ``DG'' or ``UF'' obtains better performance than the original one trained on the same dataset.} 
\label{DG_result}
\scalebox{0.825}{
\setlength{\tabcolsep}{1.9mm}{
\begin{tabular}{cccccccccccccccccc}
\specialrule{0.1em}{0.45pt}{0.45pt}
\multicolumn{1}{l}{\textbf{}}    & \textbf{Source} & \multicolumn{4}{c}{\textbf{KS20}}                                 & \multicolumn{4}{c}{\textbf{KGBD}}                                & \multicolumn{4}{c}{\textbf{IAS-Lab}}                              & \multicolumn{4}{c}{\textbf{BIWI}}                                 \\ \specialrule{0.1em}{0.45pt}{0.45pt}
\textbf{Target}                  & \textbf{Type}   & \textbf{top-1} & \textbf{top-5} & \textbf{top-10} & \textbf{mAP}  & \textbf{top-1} & \textbf{top-5} & \textbf{top-10} & \textbf{mAP} & \textbf{top-1} & \textbf{top-5} & \textbf{top-10} & \textbf{mAP}  & \textbf{top-1} & \textbf{top-5} & \textbf{top-10} & \textbf{mAP}  \\ \specialrule{0.1em}{0.45pt}{0.45pt}
\multirow{2}{*}{\textbf{KS20}}   & \textbf{DG}     & —              & —              & —               & —             & 19.7           & 54.3           & 69.7            & 10.2         & 20.1           & 60.4           & 75.6            & 13.5          & 29.7           & 62.7           & 79.9            & 15.0          \\
                                 & \textbf{UF}     & 59.0           & 79.0           & 86.2            & 21.7          & 48.4           & 77.7           & 85.2            & 21.6         & 52.2           & 78.3           & \textbf{89.1}   & \textbf{22.8} & 50.8           & \textbf{79.7}  & \textbf{87.1}   & \textbf{21.9} \\ \specialrule{0.1em}{0.45pt}{0.45pt}
\multirow{2}{*}{\textbf{KGBD}}   & \textbf{DG}     & 18.3           & 37.3           & 46.7            & 4.4           & —              & —              & —               & —            & 15.6           & 35.6           & 46.2            & 4.0           & 20.5           & 40.8           & 50.1            & 4.9           \\
                                 & \textbf{UF}     & 28.5           & 45.2           & 52.9            & 6.4           & 40.8           & 57.5           & 65.0            & 6.9          & 31.1           & 48.1           & 55.7            & 6.5           & 29.7           & 47.4           & 55.0            & 6.4           \\ \specialrule{0.1em}{0.45pt}{0.45pt}
\multirow{2}{*}{\textbf{IAS-A}}  & \textbf{DG}     & 27.9           & 57.1           & 71.5            & 15.6          & 29.6           & 56.5           & 71.6            & 16.0         & —              & —              & —               & —             & 27.5           & 53.6           & 67.9            & 14.4          \\
                                 & \textbf{UF}     & \textbf{42.8}           & \textbf{67.7}  & \textbf{77.5}   & 23.0          & 34.1           & 59.4           & 72.0            & 18.4         & 41.9           & 66.3           & 75.6            & 24.2          & 37.2           & 61.8           & 72.2            & 23.8 \\ \specialrule{0.1em}{0.45pt}{0.45pt}
\multirow{2}{*}{\textbf{IAS-B}}  & \textbf{DG}     & 32.0           & 60.6           & 72.0            & 15.7          & 29.5           & 58.8           & 68.9            & 16.5         & —              & —              & —               & —             & 27.0           & 57.6           & 70.5            & 13.5          \\
                                 & \textbf{UF}     & \textbf{45.4}  & \textbf{68.8}  & \textbf{81.4}   & \textbf{29.0} & 35.9           & 61.6           & 72.1            & 21.9         & 43.3           & 68.4           & 79.4            & 24.1          & 39.1           & 66.9           & 74.5            & \textbf{24.2} \\ \specialrule{0.1em}{0.45pt}{0.45pt}
\multirow{2}{*}{\textbf{BIWI-W}} & \textbf{DG}     & 19.3           & 31.5           & 38.9            & \textbf{19.6} & 10.3           & 22.5           & 33.1            & 12.0         & 10.0           & 23.1           & 31.1            & 15.9          & —              & —              & —               & —             \\
                                 & \textbf{UF}     & \textbf{21.6}  & \textbf{32.3}  & \textbf{40.9}   & \textbf{20.7} & 15.0           & 29.6           & 39.1            & 14.3         & 17.7           & 29.3           & 36.2            & 16.6          & 18.9           & 31.5           & 40.5            & 19.4          \\ \specialrule{0.1em}{0.45pt}{0.45pt}
\multirow{2}{*}{\textbf{BIWI-S}} & \textbf{DG}     & 23.8           & 52.2           & \textbf{69.9}   & 14.2          & 19.0           & 49.4           & 67.4            & 8.5          & 18.8           & 46.1           & 61.1            & 12.2          & —              & —              & —               & —             \\
                                 & \textbf{UF}    & \textbf{40.4}  & \textbf{62.9}  & \textbf{74.2}   & \textbf{16.2} & 21.3           & 46.5           & 57.2            & 9.8          & 27.9           & 44.5           & 63.9            & 12.8          & 34.1           & 57.3           & 69.8            & 16.0          \\ \specialrule{0.1em}{0.2pt}{0.2pt}
\end{tabular}
}
}
\end{table*}

\subsection{Broader Applications Under General Scenarios}
\label{broader_applications}
In more general person re-ID scenarios such as large-scale RGB-based scenes, the source datasets might only contain RGB images without any skeleton data or lack sufficient data for model training. \hc{For the former case,} the proposed approach can exploit unlabeled 3D skeletons estimated from RGB videos to learn effective representations for the person re-ID task. \hc{In the later case,} our approach is able to directly transfer the \hc{pre-trained} model to perform person re-ID on a new dataset. We show the broader applications of our approach under these general settings as follows.
\subsubsection{Application to Model-Estimated Skeleton Data}
\label{estimated_skeletons}
To verify the effectiveness of our skeleton-based approach when applied to large-scale RGB-based settings (CASIA-B), we exploit pre-trained pose estimation models \cite{cao2019openpose,chen20173d} to extract 3D skeleton data from RGB videos of CASIA-B, and evaluate the performance of our approach with the estimated skeleton data. We compare our approach with representative appearance-based methods \cite{weinberger2009distance,davis2007information,gray2008viewpoint,farenzena2010person,liu2015enhancing} and skeleton-based methods \cite{rao2020self,rao2021sm}. As reported in Table \ref{CASIA_B_result}, our approach is superior to recent skeleton-based methods SM-SGE and AGE with an evident performance gain of
 $7.2$-$50.4\%$ top-1 accuracy and $1.1$-$5.6\%$ mAP in four out of five evaluation conditions of CASIA-B, which substantiates that the proposed approach is capable of learning more discriminative skeleton representations than these methods \hc{in the case of using model-estimated skeleton data.}
Compared with representative classic appearance-based methods that utilize visual features, $e.g.,$ RGB features and silhouettes, our skeleton-based approach \hc{still achieves the best performance in most conditions.} For instance, our approach not only performs better than LMNN \cite{weinberger2009distance} and ITML \cite{davis2007information} that use metric learning with different visual features (\hc{RGB and HSV colors and textures}) \cite{liu2015enhancing}, but also surpasses \hc{the score-based MLR model \cite{liu2015enhancing}} that fuses RGB appearance and GEI features by up to $57.6\%$ top-1 accuracy, $39.3\%$ top-5 accuracy, and $29.1\%$ top-10 accuracy. \hc{Despite only utilizing estimated skeleton data with noise for training, the proposed unsupervised approach can still obtain highly competitive performance compared with supervised appearance-based methods in different conditions,} which demonstrates the great potential of our approach to be applied to large-scale RGB-based datasets under more general re-ID settings.

\subsubsection{Application to Generalized Person Re-Identification}
\label{generalized_reid}
Our approach can learn a unified skeleton graph representation for different skeleton data with varying body joints or topologies, which enables the pre-trained model to be directly transferred to different datasets \hc{for the generalized person re-ID task.} To evaluate the effectiveness of our approach on generalized person re-ID, we exploit the model trained on the source dataset to perform person re-ID on the target dataset\hc{, $i.e.$, direct domain generalization (DG),} and then further fine-tune the model with the unlabeled data of target datasets\hc{, $i.e.$, unsupervised fine-tuning (UF),} to
compare the generalization performance. As shown in Table \ref{DG_result}, we can draw the following observations and conclusions. The model trained on one dataset can achieve competitive person re-ID performance on other unseen target datasets. Direct generalization is shown to be effective among different datasets, while unsupervised fine-tuning on the target dataset can further improve the person re-ID performance. Such results demonstrate that our approach possesses good generalization ability with robustness to domain shifts \cite{sun2016return} and can be promisingly applied to other open person re-ID tasks. Interestingly, we observe that training on different source datasets typically leads to different person re-ID performance on a new dataset. For example, the model trained on the KGBD fails to yield satisfactory performance on IAS-B, BIWI-W and BIWI-S, while the pre-trained model of KS20 with further fine-tuning on those testing sets can achieve superior performance to the original ones\hc{, as shown by the} bold numbers in Table \ref{DG_result}, which implies that an appropriate domain initialization \hc{or model pre-training} of our model could be potentially exploited to facilitate better generalized person re-ID performance.


\section{Conclusion and Future Work}
\label{conclusion}
In this paper, we devise unified multi-level graphs to represent 3D skeletons, and 
propose an unsupervised skeleton prototype contrastive learning paradigm with multi-level relation modeling (SPC-MGR) to learn effective skeleton representations for person re-ID. We devise a multi-head structural relation layer to capture relations of neighbor body-component nodes in graphs, so as to aggregate key correlative features into effective node representations.  To capture more discriminative patterns in skeletal motion, we propose a full-level collaborative relation layer to infer dynamic collaboration among different-level components. Meanwhile, a multi-level graph fusion is exploited to integrate collaborative node features across graphs to enhance structural semantics and global pattern learning. 
\hc{Lastly,} we propose a skeleton prototype contrastive learning scheme to cluster unlabeled skeleton graph representations and contrast their inherent similarity with representative skeleton features to learn effective skeleton representations for person re-ID. The proposed SPC-MGR outperforms several state-of-the-art skeleton-based methods, and is also highly effective in more general person re-ID scenarios.

There exist two limitations in our study. The 3D skeletons used in this work are collected with relatively high precision \hc{with} little noise, while a more general scenario for skeleton collection and person re-ID ($e.g.,$ extracted from RGB videos in outdoor scenes) has not been thoroughly studied. The datasets for evaluation are with relatively limited scale when compared with prevalent RGB-based re-ID datasets like MSMT17, since large-scale 3D skeleton based re-ID benchmarks with more pedestrians and scenarios are still unavailable. To facilitate skeleton-based research, we will build and open our own skeleton-based re-ID datasets in the future.

We believe that this work makes progress towards lightweight, general, and robust person re-ID research, where we for the first time exploit unlabeled 3D skeleton data of small scale to learn discriminative and generalizable pedestrian representations to effectively perform unsupervised, multi-view, and generalized person re-ID. There are several potential directions to be further explored. More efficient clustering schemes, \hc{such as memory-based clustering,} could be devised to improve the stability and consistency of skeleton representation learning in clustering.
Combining finer-grained contrastive learning is able to help bootstrap clustering, while employing skeleton augmentation strategies could sample more instances to enhance the skeleton prototype contrastive learning. Another important direction is to explore graph-based self-supervised pretext tasks to facilitate capturing richer high-level graph semantics from unlabeled graph representations. Our model can be potentially transferred to other skeleton-based tasks\hc{, such as 3D human action recognition,} and we can expect it to synergize other modalities for more computer vision tasks.


\section{Ethical Statements}
\label{ethical}
Person re-ID is a crucial task providing significant value for both academia and industry. Nevertheless, it could be a controversial technology like face recognition, since its improper application will probably pose threat to the public privacy and society security.
In this context, we would like to emphasize that all datasets in our experiments are either publicly available (IAS-Lab, BIWI, KGBD) or officially authorized (CASIA-B, KS20). The official agents of those datasets have confirmed and guaranteed that all data are collected, processed, released, and used with the consent of subjects. For the protection of privacy, all individuals in datasets are anonymized with simple identity numbers. Besides, our approach and models must only be used for research purpose.

\ifCLASSOPTIONcaptionsoff
  \newpage
\fi


\bibliographystyle{IEEEtran}
\bibliography{IEEEabrv,IEEEexample}

\begin{thebibliography}{10}
\providecommand{\url}[1]{#1}
\csname url@samestyle\endcsname
\providecommand{\newblock}{\relax}
\providecommand{\bibinfo}[2]{#2}
\providecommand{\BIBentrySTDinterwordspacing}{\spaceskip=0pt\relax}
\providecommand{\BIBentryALTinterwordstretchfactor}{4}
\providecommand{\BIBentryALTinterwordspacing}{\spaceskip=\fontdimen2\font plus
\BIBentryALTinterwordstretchfactor\fontdimen3\font minus
  \fontdimen4\font\relax}
\providecommand{\BIBforeignlanguage}[2]{{%
\expandafter\ifx\csname l@#1\endcsname\relax
\typeout{** WARNING: IEEEtran.bst: No hyphenation pattern has been}%
\typeout{** loaded for the language `#1'. Using the pattern for}%
\typeout{** the default language instead.}%
\else
\language=\csname l@#1\endcsname
\fi
#2}}
\providecommand{\BIBdecl}{\relax}
\BIBdecl

\bibitem{nambiar2019gait}
A.~Nambiar, A.~Bernardino, and J.~C. Nascimento, ``Gait-based person
  re-identification: A survey,'' \emph{ACM Computing Surveys}, vol.~52, no.~2,
  p.~33, 2019.

\bibitem{zheng2015towards}
W.-S. Zheng, S.~Gong, and T.~Xiang, ``Towards open-world person
  re-identification by one-shot group-based verification,'' \emph{IEEE
  transactions on pattern analysis and machine intelligence}, vol.~38, no.~3,
  pp. 591--606, 2015.

\bibitem{baltieri2011sarc3d}
D.~Baltieri, R.~Vezzani, and R.~Cucchiara, ``Sarc3d: a new 3d body model for
  people tracking and re-identification,'' in \emph{International Conference on
  Image Analysis and Processing}.\hskip 1em plus 0.5em minus 0.4em\relax
  Springer, 2011, pp. 197--206.

\bibitem{vezzani2013people}
R.~Vezzani, D.~Baltieri, and R.~Cucchiara, ``People reidentification in
  surveillance and forensics: A survey,'' \emph{ACM Computing Surveys},
  vol.~46, no.~2, p.~29, 2013.

\bibitem{su2018multi}
C.~{Su}, F.~{Yang}, S.~{Zhang}, Q.~{Tian}, L.~S. {Davis}, and W.~{Gao},
  ``Multi-task learning with low rank attribute embedding for multi-camera
  person re-identification,'' \emph{IEEE Transactions on Pattern Analysis and
  Machine Intelligence}, vol.~40, no.~5, pp. 1167--1181, 2018.

\bibitem{chen2018person}
Y.-C. {Chen}, X.~{Zhu}, W.-S. {Zheng}, and J.-H. {Lai}, ``Person
  re-identification by camera correlation aware feature augmentation,''
  \emph{IEEE Transactions on Pattern Analysis and Machine Intelligence},
  vol.~40, no.~2, pp. 392--408, 2018.

\bibitem{li2019unsupervised}
M.~{Li}, X.~{Zhu}, and S.~{Gong}, ``Unsupervised tracklet person
  re-identification.'' \emph{IEEE Transactions on Pattern Analysis and Machine
  Intelligence}, pp. 1--1, 2019.

\bibitem{qian2019leader}
X.~Qian, Y.~Fu, T.~Xiang, Y.-G. Jiang, and X.~Xue, ``Leader-based multi-scale
  attention deep architecture for person re-identification,'' \emph{IEEE
  Transaction on Pattern Analysis Machine Intelligence}, 2019.

\bibitem{yu2020unsupervised}
H.-X. {Yu}, A.~{Wu}, and W.-S. {Zheng}, ``Unsupervised person re-identification
  by deep asymmetric metric embedding,'' \emph{IEEE Transactions on Pattern
  Analysis and Machine Intelligence}, vol.~42, no.~4, pp. 956--973, 2020.

\bibitem{wang2003silhouette}
L.~Wang, T.~Tan, H.~Ning, and W.~Hu, ``Silhouette analysis-based gait
  recognition for human identification,'' \emph{IEEE Transactions on Pattern
  Analysis and Machine Intelligence}, vol.~25, no.~12, pp. 1505--1518, 2003.

\bibitem{wang2011human}
C.~Wang, J.~Zhang, L.~Wang, J.~Pu, and X.~Yuan, ``Human identification using
  temporal information preserving gait template,'' \emph{IEEE Transactions on
  Pattern Analysis and Machine Intelligence}, vol.~34, no.~11, pp. 2164--2176,
  2011.

\bibitem{wang2016person}
T.~Wang, S.~Gong, X.~Zhu, and S.~Wang, ``Person re-identification by
  discriminative selection in video ranking,'' \emph{IEEE Transactions on
  Pattern Analysis and Machine Intelligence}, vol.~38, no.~12, pp. 2501--2514,
  2016.

\bibitem{zhao2017person}
R.~{Zhao}, W.~{Oyang}, and X.~{Wang}, ``Person re-identification by saliency
  learning,'' \emph{IEEE Transactions on Pattern Analysis and Machine
  Intelligence}, vol.~39, no.~2, pp. 356--370, 2017.

\bibitem{zhang2019densely}
Z.~{Zhang}, C.~{Lan}, W.~{Zeng}, and Z.~{Chen}, ``Densely semantically aligned
  person re-identification,'' in \emph{CVPR}, 2019, pp. 667--676.

\bibitem{karianakis2018reinforced}
N.~Karianakis, Z.~Liu, Y.~Chen, and S.~Soatto, ``Reinforced temporal attention
  and split-rate transfer for depth-based person re-identification,'' in
  \emph{ECCV}, 2018, pp. 715--733.

\bibitem{ge2020selfpaced}
Y.~Ge, F.~Zhu, D.~Chen, R.~Zhao, and H.~Li, ``Self-paced contrastive learning
  with hybrid memory for domain adaptive object re-id,'' in \emph{NeurIPS},
  2020.

\bibitem{barbosa2012re}
I.~B. Barbosa, M.~Cristani, A.~Del~Bue, L.~Bazzani, and V.~Murino,
  ``Re-identification with rgb-d sensors,'' in \emph{ECCV}.\hskip 1em plus
  0.5em minus 0.4em\relax Springer, 2012, pp. 433--442.

\bibitem{andersson2015person}
V.~O. Andersson and R.~M. Araujo, ``Person identification using anthropometric
  and gait data from kinect sensor,'' in \emph{AAAI}, 2015.

\bibitem{rao2020self}
H.~{Rao}, S.~{Wang}, X.~{Hu}, M.~{Tan}, H.~{Da}, J.~{Cheng}, and B.~{Hu},
  ``Self-supervised gait encoding with locality-aware attention for person
  re-identification,'' in \emph{IJCAI}, vol.~1, 2020, pp. 898--905.

\bibitem{rao2021self}
H.~Rao, S.~Wang, X.~Hu, M.~Tan, Y.~Guo, J.~Cheng, X.~Liu, and B.~Hu, ``A
  self-supervised gait encoding approach with locality-awareness for 3d
  skeleton based person re-identification,'' \emph{IEEE Transactions on Pattern
  Analysis and Machine Intelligence}, 2021.

\bibitem{rao2021multi}
H.~Rao, S.~Xu, X.~Hu, J.~Cheng, and B.~Hu, ``Multi-level graph encoding with
  structural-collaborative relation learning for skeleton-based person
  re-identification,'' in \emph{IJCAI}, 2021.

\bibitem{rao2021sm}
H.~Rao, X.~Hu, J.~Cheng, and B.~Hu, ``Sm-sge: A self-supervised multi-scale
  skeleton graph encoding framework for person re-identification,'' in
  \emph{Proceedings of the 29th ACM international conference on Multimedia},
  2021.

\bibitem{han2017space}
F.~Han, B.~Reily, W.~Hoff, and H.~Zhang, ``Space-time representation of people
  based on 3d skeletal data: A review,'' \emph{Computer Vision and Image
  Understanding}, vol. 158, pp. 85--105, 2017.

\bibitem{991036}
R.~{Tanawongsuwan} and A.~{Bobick}, ``Gait recognition from time-normalized
  joint-angle trajectories in the walking plane,'' in \emph{CVPR}, vol.~2, Dec
  2001, pp. II--II.

\bibitem{liao2020model}
R.~Liao, S.~Yu, W.~An, and Y.~Huang, ``A model-based gait recognition method
  with body pose and human prior knowledge,'' \emph{Pattern Recognition},
  vol.~98, p. 107069, 2020.

\bibitem{munaro2014one}
M.~Munaro, A.~Fossati, A.~Basso, E.~Menegatti, and L.~Van~Gool, ``One-shot
  person re-identification with a consumer depth camera,'' in \emph{Person
  Re-Identification}.\hskip 1em plus 0.5em minus 0.4em\relax Springer, 2014,
  pp. 161--181.

\bibitem{yoo2002extracting}
J.-H. Yoo, M.~S. Nixon, and C.~J. Harris, ``Extracting gait signatures based on
  anatomical knowledge,'' in \emph{Proceedings of BMVA Symposium on Advancing
  Biometric Technologies}.\hskip 1em plus 0.5em minus 0.4em\relax Citeseer,
  2002, pp. 596--606.

\bibitem{murray1964walking}
M.~P. Murray, A.~B. Drought, and R.~C. Kory, ``Walking patterns of normal
  men,'' \emph{Journal of Bone and Joint Surgery}, vol.~46, no.~2, pp.
  335--360, 1964.

\bibitem{pala2019enhanced}
P.~Pala, L.~Seidenari, S.~Berretti, and A.~Del~Bimbo, ``Enhanced skeleton and
  face 3d data for person re-identification from depth cameras,''
  \emph{Computers \& Graphics}, 2019.

\bibitem{munaro20143d}
M.~Munaro, A.~Basso, A.~Fossati, L.~Van~Gool, and E.~Menegatti, ``3d
  reconstruction of freely moving persons for re-identification with a depth
  sensor,'' in \emph{ICRA}.\hskip 1em plus 0.5em minus 0.4em\relax IEEE, 2014,
  pp. 4512--4519.

\bibitem{hadsell2006dimensionality}
R.~Hadsell, S.~Chopra, and Y.~LeCun, ``Dimensionality reduction by learning an
  invariant mapping,'' in \emph{CVPR}, vol.~2, 2006, pp. 1735--1742.

\bibitem{wu2018unsupervised}
Z.~Wu, Y.~Xiong, S.~X. Yu, and D.~Lin, ``Unsupervised feature learning via
  non-parametric instance discrimination,'' in \emph{CVPR}, 2018, pp.
  3733--3742.

\bibitem{oord2018representation}
A.~v.~d. Oord, Y.~Li, and O.~Vinyals, ``Representation learning with
  contrastive predictive coding,'' \emph{arXiv preprint arXiv:1807.03748},
  2018.

\bibitem{chen2020big}
T.~Chen, S.~Kornblith, K.~Swersky, M.~Norouzi, and G.~Hinton, ``Big
  self-supervised models are strong semi-supervised learners,'' in
  \emph{NeurIPS}, 2020.

\bibitem{li2021prototypical}
J.~Li, P.~Zhou, C.~Xiong, and S.~Hoi, ``Prototypical contrastive learning of
  unsupervised representations,'' in \emph{ICLR}, 2021.

\bibitem{dosovitskiy2014discriminative}
A.~Dosovitskiy, J.~T. Springenberg, M.~Riedmiller, and T.~Brox,
  ``Discriminative unsupervised feature learning with convolutional neural
  networks,'' in \emph{NeurIPS}, 2014, pp. 766--774.

\bibitem{gutmann2010noise}
M.~Gutmann and A.~Hyv{\"a}rinen, ``Noise-contrastive estimation: A new
  estimation principle for unnormalized statistical models,'' in
  \emph{International Conference on Artificial Intelligence and Statistics},
  2010, pp. 297--304.

\bibitem{zhuang2019local}
C.~Zhuang, A.~L. Zhai, and D.~Yamins, ``Local aggregation for unsupervised
  learning of visual embeddings,'' in \emph{ICCV}, 2019, pp. 6002--6012.

\bibitem{tian2019contrastive}
Y.~Tian, D.~Krishnan, and P.~Isola, ``Contrastive multiview coding,''
  \emph{arXiv preprint arXiv:1906.05849}, 2019.

\bibitem{misra2020self}
I.~{Misra} and L.~van~der {Maaten}, ``Self-supervised learning of
  pretext-invariant representations,'' in \emph{CVPR}, 2020, pp. 6707--6717.

\bibitem{ye2019unsupervised}
M.~Ye, X.~Zhang, P.~C. Yuen, and S.-F. Chang, ``Unsupervised embedding learning
  via invariant and spreading instance feature,'' in \emph{CVPR}, 2019, pp.
  6210--6219.

\bibitem{ji2019invariant}
X.~Ji, J.~F. Henriques, and A.~Vedaldi, ``Invariant information clustering for
  unsupervised image classification and segmentation,'' in \emph{ICCV}, 2019,
  pp. 9865--9874.

\bibitem{chen2020a}
T.~{Chen}, S.~{Kornblith}, M.~{Norouzi}, and G.~{Hinton}, ``A simple framework
  for contrastive learning of visual representations,'' in \emph{ICML}, 2020.

\bibitem{he2020momentum}
K.~{He}, H.~{Fan}, Y.~{Wu}, S.~{Xie}, and R.~{Girshick}, ``Momentum contrast
  for unsupervised visual representation learning,'' in \emph{CVPR}, 2020, pp.
  9729--9738.

\bibitem{winter2009biomechanics}
D.~A. Winter, \emph{Biomechanics and motor control of human movement}.\hskip
  1em plus 0.5em minus 0.4em\relax John Wiley \& Sons, 2009.

\bibitem{aggarwal1998nonrigid}
J.~K. Aggarwal, Q.~Cai, W.~Liao, and B.~Sabata, ``Nonrigid motion analysis:
  Articulated and elastic motion,'' \emph{Computer Vision and Image
  Understanding}, vol.~70, no.~2, pp. 142--156, 1998.

\bibitem{velickovic2018graph}
P.~{Velickovic}, G.~{Cucurull}, A.~{Casanova}, A.~{Romero}, P.~{Liò}, and
  Y.~{Bengio}, ``Graph attention networks.'' in \emph{ICLR}, 2018.

\bibitem{ester1996density}
M.~Ester, H.-P. Kriegel, J.~Sander, X.~Xu \emph{et~al.}, ``A density-based
  algorithm for discovering clusters in large spatial databases with noise.''
  in \emph{kdd}, vol.~96, no.~34, 1996, pp. 226--231.

\bibitem{hinton2015distilling}
G.~Hinton, O.~Vinyals, and J.~Dean, ``Distilling the knowledge in a neural
  network,'' in \emph{NeurIPS Workshop}, 2014.

\bibitem{nambiar2017context}
A.~Nambiar, A.~Bernardino, J.~C. Nascimento, and A.~Fred, ``Context-aware
  person re-identification in the wild via fusion of gait and anthropometric
  features,'' in \emph{International Conference on Automatic Face \& Gesture
  Recognition}.\hskip 1em plus 0.5em minus 0.4em\relax IEEE, 2017, pp.
  973--980.

\bibitem{munaro2014feature}
M.~Munaro, S.~Ghidoni, D.~T. Dizmen, and E.~Menegatti, ``A feature-based
  approach to people re-identification using skeleton keypoints,'' in
  \emph{ICRA}.\hskip 1em plus 0.5em minus 0.4em\relax IEEE, 2014, pp.
  5644--5651.

\bibitem{yu2006framework}
S.~Yu, D.~Tan, and T.~Tan, ``A framework for evaluating the effect of view
  angle, clothing and carrying condition on gait recognition,'' in \emph{ICPR},
  vol.~4.\hskip 1em plus 0.5em minus 0.4em\relax IEEE, 2006, pp. 441--444.

\bibitem{zheng2015scalable}
L.~Zheng, L.~Shen, L.~Tian, S.~Wang, J.~Wang, and Q.~Tian, ``Scalable person
  re-identification: A benchmark,'' in \emph{ICCV}, 2015, pp. 1116--1124.

\bibitem{gray2008viewpoint}
D.~Gray and H.~Tao, ``Viewpoint invariant pedestrian recognition with an
  ensemble of localized features,'' in \emph{ECCV}.\hskip 1em plus 0.5em minus
  0.4em\relax Springer, 2008, pp. 262--275.

\bibitem{weinberger2009distance}
K.~Q. Weinberger and L.~K. Saul, ``Distance metric learning for large margin
  nearest neighbor classification.'' \emph{Journal of machine learning
  research}, vol.~10, no.~2, 2009.

\bibitem{davis2007information}
J.~V. Davis, B.~Kulis, P.~Jain, S.~Sra, and I.~S. Dhillon,
  ``Information-theoretic metric learning,'' in \emph{ICML}, 2007, pp.
  209--216.

\bibitem{farenzena2010person}
M.~Farenzena, L.~Bazzani, A.~Perina, V.~Murino, and M.~Cristani, ``Person
  re-identification by symmetry-driven accumulation of local features,'' in
  \emph{CVPR}.\hskip 1em plus 0.5em minus 0.4em\relax IEEE, 2010, pp.
  2360--2367.

\bibitem{liu2015enhancing}
Z.~Liu, Z.~Zhang, Q.~Wu, and Y.~Wang, ``Enhancing person re-identification by
  integrating gait biometric,'' \emph{Neurocomputing}, vol. 168, pp.
  1144--1156, 2015.

\bibitem{van2008visualizing}
L.~Van~der Maaten and G.~Hinton, ``Visualizing data using t-sne.''
  \emph{Journal of machine learning research}, vol.~9, no.~11, 2008.

\bibitem{cao2019openpose}
Z.~Cao, G.~Hidalgo, T.~Simon, S.-E. Wei, and Y.~Sheikh, ``Openpose: realtime
  multi-person 2d pose estimation using part affinity fields,'' \emph{IEEE
  transactions on pattern analysis and machine intelligence}, vol.~43, no.~1,
  pp. 172--186, 2019.

\bibitem{chen20173d}
C.-H. Chen and D.~Ramanan, ``3d human pose estimation= 2d pose estimation+
  matching,'' in \emph{CVPR}, 2017, pp. 7035--7043.

\bibitem{sun2016return}
B.~Sun, J.~Feng, and K.~Saenko, ``Return of frustratingly easy domain
  adaptation,'' in \emph{AAAI}, vol.~30, no.~1, 2016.

\end{thebibliography}
\end{document}